%% file: paper.tex
\documentclass[accepted]{uai2025} %

\usepackage[american]{babel}

\usepackage{natbib} %
\usepackage{mathtools} %
\usepackage{booktabs} %
\usepackage[dvipsnames, table]{xcolor}
\usepackage{tikz} %

\usepackage{xspace} %
\usepackage{placeins}

\usepackage{amsmath}
\usepackage{amssymb}
\usepackage{mathtools}
\usepackage{amsthm}

\usepackage{palette}
\usepackage{researchpack}
\usepackage{pgfplots}
\pgfplotsset{compat=newest}

\usepackage{algorithm}
\usepackage{algorithmic}
\usepackage[capitalise]{cleveref}

\hypersetup{
colorlinks=true,linkcolor=ForestGreen,citecolor=Bittersweet,urlcolor=purple
}

\newcommand{\av}[1]{}
\newcommand{\lk}[1]{}
\newcommand{\pam}[1]{}
\newcommand{\ap}[1]{}
\newcommand{\rs}[1]{}

\newcommand{\ours}{\texttt{PAL}\xspace}
\newcommand{\gasp}{\texttt{GASP!}\xspace}
\newcommand{\latte}{\texttt{LattE}\xspace}

\newcommand{\lra}{\mathcal{LRA}}
\newcommand{\smtlra}{SMT($\lra$)\xspace}

\title{A Probabilistic Neuro-symbolic Layer for Algebraic Constraint Satisfaction}

\author[1]{Leander Kurscheidt}
\author[2]{Paolo Morettin}
\author[2]{Roberto Sebastiani}
\author[2]{Andrea Passerini}
\author[1]{Antonio Vergari}
\affil[1]{%
   School of Informatics, University of Edinburgh, UK
}
\affil[2]{%
   DiSI, University of Trento, Italy
}
  
\begin{document}
\maketitle

\begin{abstract}
In safety-critical applications, guaranteeing the satisfaction of constraints over continuous environments is crucial,
e.g., an autonomous agent should never crash into obstacles or go off-road.  
Neural models struggle in the presence of these constraints, especially when they involve intricate algebraic relationships. 
To address this, we introduce a differentiable probabilistic layer that guarantees the satisfaction of non-convex algebraic constraints over 
continuous variables. 
This probabilistic algebraic layer (\ours) can be seamlessly plugged into any neural architecture and trained via maximum likelihood without requiring approximations. 
\ours defines a distribution over conjunctions and disjunctions of linear inequalities, parameterized by polynomials. This formulation enables efficient and exact renormalization via symbolic integration, which can be amortized across different data points and easily parallelized on a GPU. %
We showcase \ours and our integration scheme on  benchmarks for algebraic constraint integration and on real-world trajectory data.
\end{abstract}

\section{Introduction}
\label{sec:introduction}

\begin{figure}[t]
    \setlength{\tabcolsep}{2pt}
    \begin{tabular}
    {cccc}
        \small Ground Truth & \small GMM & \small Flow & \small\ours (\textit{ours})\\
        \includegraphics[width=1.9cm]{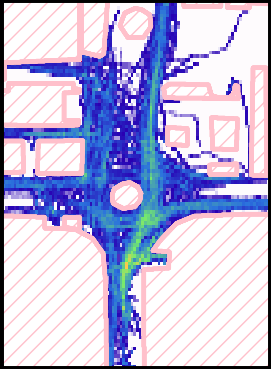}&
        \includegraphics[width=1.9cm]{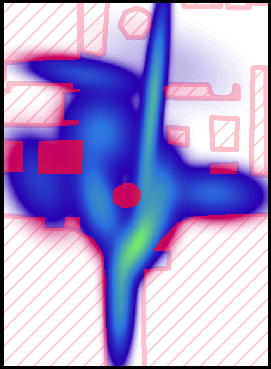}&
        \includegraphics[width=1.9cm]{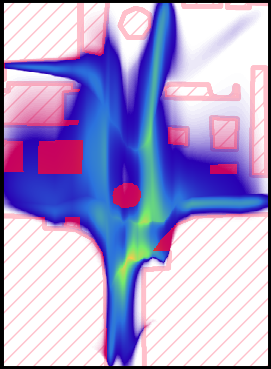}&
        \includegraphics[width=1.9cm]{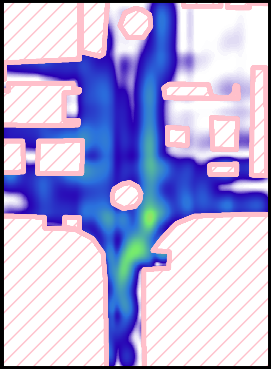}
        
    \end{tabular}
\hfill
    \caption{
    \textbf{\ours is guaranteed to place probability mass only within a given constraint} here represented as the (non deleted) walkable area of a map from the Stanford Drone Dataset, while unconstrained distribution estimators 
    violate the constraint (area shown in red).
    See \cref{sec:stanford-exp}.
    }
    \label{fig:traj-density}
\end{figure}

In safety-critical applications, a \textit{reliable} AI system should be uncertainty-aware and deliver calibrated \textit{probabilities} over its predictions. 
At the same time, it should confidently and consistently assign zero probability to certain states of the world if they are invalid, i.e., if they are violating \textit{constraints}.
These constraints can encode prior knowledge such as explicit rules \citep{marconato2023not,marconato2024bears,bortolotti2024benchmark} that can be crucial for the safety of the system and its users.  
For instance, they can encode that self-driving cars should avoid crashing into obstacles or go off-road \citep{xu2020boia,giunchiglia2023road}.

The promise of probabilistic neuro-symbolic (NeSy) methods \citep{garcez2019neural,garcez2022neural,de2021statistical} 
for trustworthy AI is to integrate symbolic reasoning over these high-level constraints into 
neural predictors.
The simplest way to achieve this is to encourage constraint satisfaction via a loss penalty at training time \citep{semanticloss1,dl2,deepseaproblog}. 
However, such an approach might have catastrophic consequences at test time, as it does not \textit{guarantee} that invalid configurations will be associated exactly probability zero: even a probability of 0.001\% to violate a constraint can be considered harmful in safety-critical applications such as autonomous driving.

A number of recent works overcome this issue by proposing architectures that \textit{certify the satisfaction of given constraints by design} \citep{manhaeve2018deepproblog, giunchiglia2020coherent, ahmed2022semantic, hoernle2022multiplexnet}.
Unfortunately, they mostly consider constraints over Boolean or discrete variables only \citep{defast} and extending them to continuous variables is highly non trivial.
This is because tackling constraints over \textit{continuous} variables 
poses unique challenges.
First, ensuring modeling a proper distribution over the constraint, i.e., renormalizing a density such that it exactly integrates to 1, is a \#P-hard problem \citep{Baldoni2008HowTI} even when the distribution and constraints have a relatively simple structure \citep{zeng2020scaling,zeng2020probabilistic}.   
Second, real-world constraints over continuous variables come in the form of intricate \textit{algebraic} relationships \citep{barrettsst21,wmipa}. 
Even considering simple linear inequalities among variables will entail that the support of the induced densities can take the form of (disjunctions of) non-convex polytopes \citep{DBLP:journals/corr/abs-2502-18237}.
While focusing on single convex polytope constraints can be easier
\citep{DBLP:journals/corr/abs-2402-04823}, it fails to capture real-world scenarios, such as modeling multiple obstacles that need to be avoided simultaneously on a map (\cref{fig:traj-density}). 
Scalability during learning is another major concern, as one would ideally want to compute fast and exact gradients for each data point.
In practice, this is sometimes achieved by approximating or relaxing the constraint \citep{deepseaproblog,de2023differentiable} thus giving up learning by exact maximum likelihood.

In this work, we narrow these gaps by introducing a \textit{probabilistic algebraic layer} (\ours) that can be seamlessly plugged as the last layer into any neural architecture while guaranteeing probabilistic predictions to satisfy complex algebraic constraints.
Our formulation for \ours uses a symbolic integration scheme that scales gracefully as it can be amortized, i.e., computed once for all datapoints, while allowing for exact gradients and hence exact maximum likelihood training, that is 
being retro-compatible with out-of-the-box optimizers using automatic differentiation.

\textbf{Contributions.}
After formalizing the problem of probabilistic prediction over algebraic constraint in \cref{sec:background}, we \textbf{C1)} introduce \ours and its ingredients while discussing its properties in \cref{sec:poly};
then we \textbf{C2)} introduce the GPU-accelerated symbolic polynomial integrator that powers \ours in \cref{sec:gasp}.
Finally, we
\textbf{C3)} run an extensive set of experiments, comprising both standard benchmarks for algebraic constraint integration and real-world trajectory data from the Stanford drone dataset \citep{DBLP:conf/eccv/RobicquetSAS16}, which we augment by manually segmenting constraints representing obstacles and buildings.

\section{Probabilistic Predictions under Algebraic Constraints}
\label{sec:background}

\textbf{Notation}.
Uppercase letters denote random variables ($X,Y$) and lowercase letters denote their assignments ($x, y$). 
We use bold for sets of variables ($\vX,\vY$), and their joint assignments ($\vx,\vy$).
We use lowercase Greek letters for denoting algebraic constraints ($\phi$) and uppercase Greek letters for denoting (vectors of) learnable parameters.
We say that $\vy$ satisfies $\phi$ (written $\vy\models\phi$) if and only if substituting $\vY$ with $\vy$ in $\phi$ makes $\phi$ true.
We denote the indicator function as $\Ind{.}$, therefore, $\Ind{\vY \models \phi}$ evaluates to 1 for all the values of $\vY$ satisfying $\phi$ and 0 otherwise. 

\begin{figure}[!t]
    \centering
    \begin{minipage}{0.32\columnwidth}
    \centering
    \includegraphics[height=1.0\textwidth]{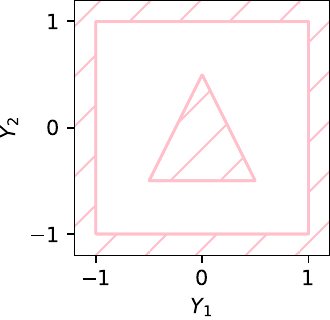}
    \end{minipage}\hspace{5pt}\begin{minipage}{0.32\columnwidth}
    \centering
    \includegraphics[height=1.0\textwidth]{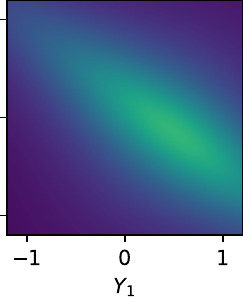}
    \end{minipage}\begin{minipage}{0.32\columnwidth}
    \centering
    \includegraphics[height=1.0\textwidth]{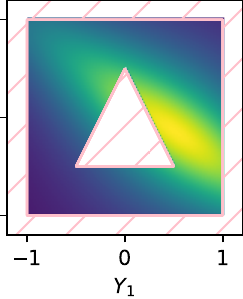}
    \end{minipage}
    \caption{\textbf{A renormalized constrained density} (right) obtained by applying \textbf{non-convex algebraic constraints} (left) over an unconstrained distribution (middle).}
    \label{fig:cut_triangle_constraints_viz}
\end{figure}

\textbf{Setting}.  
We aim at learning a parameterized conditional distribution of $\vY$ given $\vX$, i.e. $p_{\boldsymbol{\Theta}}(\vY \mid \vX)$, from a dataset $\mathcal{D}=(\vx^{(i)},\vy^{(i)})$ of realizations of mixed continuous and discrete variables $\vX$ and $\vY$.
As discussed later, the challenge will be when some $\vY$ are continuous, and we can always assume all of them to be continuous without loss of generality.
Differently from a classical supervised learning scenario, 
the support of $p_{\boldsymbol{\Theta}}$ is not the whole $\bbR^{|\vY|}$, but a restriction encoded by a constraint $\phi$ defined over $\vY$.
This constraint encodes which regions of the label space are invalid, i.e., should have exactly zero-probability and therefore should not be sampled nor predicted \citep{grivas2024taming}.
We give an example next.

\textbf{Which constraints?}
We focus on algebraic constraints as Boolean combinations of linear inequalities, which are flexible enough to represent complex supports in the form of disjunctions of non-convex polytopes.
We do so in the language of (quantifier-free) \textit{satisfiability modulo linear real arithmetic} (\smtlra) formulas~\citep{barrettsst21}. %
A constraint in \smtlra, from here on simply SMT formula or constraint, is a logical formula with arbitrary combinations of the usual Boolean connectives ($\land, \lor, \neg$) over atoms that are restricted to linear inequalities over
$\vY$:
$$\left(\sum\nolimits_i a_i Y_i \bowtie b\right) \quad a_i, b \in \mathbb{Q}, \:\bowtie \: \in \{\le, <, \ge, >, =\}.$$

\noindent \emph{Example}.
Consider the problem of learning a conditional distribution $p(Y_1, Y_2 \mid X_1, X_2)$  subject to the following constraint $\phi$:
$\vY \in [-1,1]^2 \land \vY \notin \triangle$, where $\triangle$ denotes the triangle with vertices $(-0.5, -0.5)$, $(0, 0.5)$ and $(0.5, -0.5)$, as illustrated in \cref{fig:cut_triangle_constraints_viz} (left).
We can encode such a constraint into \smtlra as:
\begin{align*}
    &\phi = (-1 \le Y_1) \land (Y_1 \le 1) \land (-1 \le Y_2) \land (Y_2 \le 1)\land \\
    & \left[ (Y_2 < -0.5) \lor (2Y_1 + 0.5 < Y_2) \lor (-2Y_1 + 0.5 < Y_2)\right].
\end{align*}
In the more realistic obstacle-avoidance example in the introduction (\cref{fig:traj-density}), 
$\phi$ can model the surface where cars and pedestrians are allowed to move.
One principled solution to design a conditional distribution $p_{\boldsymbol{\Theta}}$  that is constraint-aware is to realize a \textit{product of experts} \citep{hinton2002training}
\begin{equation}
    \label{eq:prod-experts-constraint}
    p_{\boldsymbol{\Theta}}(\vY \mid \vX = \vx) \propto q_{\boldsymbol{\Theta}}(\vY\mid\vx) \Ind{\vY \models \phi}
\end{equation}
where $q_{\boldsymbol{\Theta}}$ is an unconstrained density whose support is the full $\bbR^{|\vY|}$ (\cref{fig:cut_triangle_constraints_viz}, middle) and $\Ind{\vY \models \phi}$ encodes the satisfaction of the constraint $\phi$.
Unfortunately \cref{eq:prod-experts-constraint} alone does not encode a proper density, as it does not integrate to 1.
Alternative ways to train such a product of experts are possible \citep{hinton2002training}, but they are not retro-compatible with the usual gradient-based optimization recipe to train neural networks: maximum likelihood estimation (MLE). 
To learn the parameters $\boldsymbol{\Theta}$ by \textit{exact} MLE, and hence to design a layer can be seamlessly plugged-in any network, we would need to compute its renormalization constant, i.e., 
\begin{equation*}
\tag{WMI}
\label{eq:wmi}
    \int
    q_{\boldsymbol{\Theta}}(\vy\mid\vx) \Ind{\vy \models \phi} d\vY
\end{equation*}
which is also known as a\textit{ weighted model integration} (WMI) problem \citep{wmi}, i.e., the probability that the constraint is satisfied, $\mathsf{Pr}(\phi)$. 
\cref{app:wmi} discusses the background and literature behind \ref{eq:wmi} in depth.

Here we first note that we treat all variables $\vY$ in \cref{eq:wmi} to be continuous, something that we can do without loss of generality as we can always reduce a WMI problem over mixed discrete-continuous variables to one over continuous variables only without increasing the problem dimensionality \citep{wmireduction}.
However, the complexity of solving a WMI problem exactly is \#P-hard in general \citep{zeng2020probabilistic} and tractable algorithms are available only when the constraints $\phi$ come with certain structures \citep{zeng2020scaling}.
In the need to scale the computation of  \cref{eq:wmi}, many NeSy approaches opted to \textit{approximate} it.
For example, DeepSeaProblog (DSP) \citep{deepseaproblog} (see \cref{sec:related-works} for a discussion) designs a general probabilistic logic programming framework for WMI problems parameterized by neural networks.
However, to practically compute the \ref{eq:wmi} integral, DSP employs rejection sampling. 
Not only does this require to relax the constraints to perform backpropagation, yielding high-variance gradients, but it also hinders scalability, as the \ref{eq:wmi} integral \textit{needs to be approximated for each datapoint} $\vx$.

To be able to scale neural networks with complex real-life constraints such as the ones in \cref{fig:traj-density}, while retaining constraint satisfaction guarantees, we have to push the current boundaries of the WMI literature.
Next, we discuss how to do so while selecting a parametric form for $q_{\boldsymbol{\Theta}}$ (\cref{sec:poly}) that allows us to efficiently amortize the computation of the \ref{eq:wmi} integral into a symbolic computational graph that, once built, can leverage GPU parallelism (\cref{sec:gasp}).

\section{A PAL for guaranteed Constraint Satisfaction}
\label{sec:poly}

We devise a differentiable \textit{probabilistic algebraic layer} (PAL) realizing the following conditional distribution: 
\begin{equation}
\tag{PAL}
    \label{eq:pal}
    p_{\boldsymbol{\Theta}}(\vY \mid \vx) = \frac{q(\vY; \boldsymbol{\lambda}=f_{\boldsymbol{\psi}}(\vx)) \Ind{\vY \models \phi}}{\int q(\vy'; \boldsymbol{\lambda}=f_{\boldsymbol{\psi}}(\vx)) \Ind{\vy' \models \phi} d\vY'}
\end{equation}
where $\boldsymbol{\Theta}=\{\boldsymbol{\lambda},\boldsymbol{\psi}\}$ and $q(\vY; \boldsymbol{\lambda}=f_{\boldsymbol{\psi}}(\vx))$ is a flexible unconstrained distribution whose parameters $\boldsymbol{\lambda}$ are the output of a neural backbone $f_{\boldsymbol{\psi}}$ that takes as input $\vx$ and $\phi$ encodes an SMT constraint as discussed above.
As discussed in \cref{sec:background},
this construction guarantees that the density $p_{\boldsymbol{\Theta}}$ is non-zero only inside the region defined by $\phi$ (\cref{fig:cut_triangle_constraints_viz}, right).
Note that our layer can be used as a standalone distribution estimator when there are no input variables to condition on, i.e., to model $p_{\boldsymbol{\lambda}}(\vY)$.

We can design a neural backbone $f_{\boldsymbol{\psi}}$ by easily reusing any  existing architecture.
{Given any (possibly pretrained) neural network $\vz=h(\vx)$ that outputs an embedding $\vz$ for each datapoint $\vx$, in fact we can realize $f$ as $\boldsymbol{\lambda}=g(h(\vx))$ by adding a simple \textit{gating function} $g$ that maps $\vz$ to $\boldsymbol{\lambda}$.
Less trivial is to select a suitable model family for $q_{\boldsymbol{\lambda}}$ that allows us to efficiently amortize the computation of the denominator of \ref{eq:pal}.
If we had to deal \textit{only} with Boolean variables $\vY$ and propositional constraints $\phi$, we could leverage
probabilistic circuits (PCs) \citep{darwiche2002knowledge,choi2020probabilistic}, 
compact multilinear polynomials over tractable functions
as in \citet{ahmed2022semantic}.
Unfortunately, there is no equivalent circuit representation for SMT constraints with the required structural properties to guarantee tractability \citep{vergari2021compositional}.

\textbf{General polynomials to the rescue.} Whereas we cannot leverage the properties of structured polynomials such as PCs, we can still employ general (piecewise) polynomials to model the unconstrained density $q_{\boldsymbol{\lambda}}$ in \ref{eq:pal}. 
They will still provide \textit{expressiveness} as they can approximate any density up to arbitrary precision \citep{wmisurvey,chengmultilinear} and, more crucially, they are closed under integration over a polytope \citep{zeng2020scaling}.
The general form of polynomials we consider is therefore:
\begin{equation}
    \label{eq:poly-def}
    q(\vy;\boldsymbol{\lambda}) = \sum\nolimits_{i=1}^{M}\lambda_i\prod\nolimits_j y_j^{\alpha_{ij}}
\end{equation}
where each coefficient $\lambda_i\in\bbR$ is one of the outputs of the backbone $f_{\boldsymbol\psi}$ and the exponents $\alpha_{ij}\in\bbN$ are constant parameters.
As we will discuss in the next section, the complexity of integration will depend, among other things, on the degree of the polynomial and on the number of monomials $M$.
At the same time, this will offer the opportunity to design a fixed-parameter tractable integration scheme, as we can bound the polynomial degree by construction.

\textbf{How to construct polynomials?}
We can generate polynomial structures in an exhaustive way given a max degree $d$, i.e., by generating all possible monomials with degree $\leq d$, or by randomly subsampling them in order to have a compact polynomial of high degree.
Alternatively, we can use piecewise polynomials such as \textit{splines}, which have been demonstrated to be very expressive in ML, even with low degree \citep{DBLP:conf/nips/DurkanB0P19}.
In particular, in our experiments, we use cubic Hermite splines \citep{smith1980practical}, see \cref{app:splines} for details.

Whereas all the aforementioned polynomials can always be rewritten in the canonical form of \cref{eq:poly-def}, they might not guarantee to model a valid density as they might yield negative values.
To this end, we propose to use \textit{squared polynomials} in \ref{eq:pal}, which have been recently investigated in the PC literature for their expressiveness properties
\citep{loconte2024subtractive,loconte2024sos,loconte2025faster}.
Specifically, we consider sum of squared polynomials, i.e., polynomials of the form:
\begin{equation}
    \label{eq:sum-squares-poly}
    \sum\nolimits_{k}w_k\left(\sum\nolimits_i u_i \prod\nolimits_j y_j^{\alpha_{ijk}}\right)^{2}
\end{equation}
where $w_k>0$ and $u_i\in\bbR$ and therefore where $\boldsymbol{\lambda}=\{w_k\}_{k}\cup\{u_i\}_{i}$ when we rewrite \cref{eq:sum-squares-poly} into \cref{eq:poly-def}. 
Now we have all the ingredients to discuss how to parallelize the computation of the denominator in \ref{eq:pal}.

\begin{figure*}
  \centering
  \includegraphics[width=0.8\textwidth]{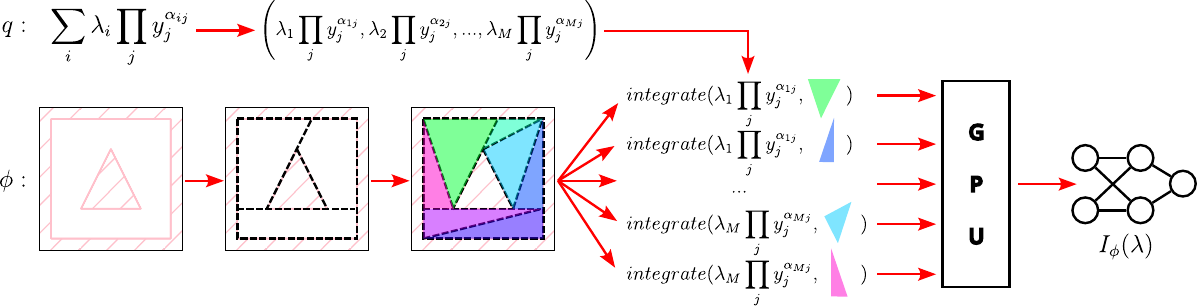}
\caption{\textbf{An overview of our pipeline with  \gasp}. The integrand $q$ is decomposed into parallelizable monomial integrations (\emph{top}). The non-convex constraint is first decomposed into several convex regions (\cref{app:wmi}) that are  further decomposed into (colored) simplices (\emph{bottom}). The resulting computational graph is highly parallelizable on a GPU.} 
\label{fig:gasp_pipeline}
\end{figure*}

\section{Extreme Parallelization of Polynomial Integrals on a GPU}
\label{sec:gasp}

State-of-the-art (SoTA) WMI solvers propose various ways to break the \ref{eq:wmi} integral over $\phi$ into smaller integrals over disjoint  convex polytopes $\mu_1, ... \mu_K$ such that $\bigvee_i \mu_i \equiv \phi$.
Hence, from here on, we will focus on the problem of integrating a polynomial in canonical form (\cref{eq:poly-def}) over a single convex polytope.
\textit{Our solution to scale the computation of this simpler problem will therefore speed up any SoTA WMI solver}.
In practice, for \ref{eq:pal} we will adopt SAE4WMI~\citep{sae4wmi} to break $\phi$ into  $\mu_1, ... \mu_K$, as it  is the most advanced WMI solver that deals with arbitrary non-convex algebraic constraints $\phi$ at the time of writing. 
See \cref{app:wmi} for a discussion. 

Our solution, named GPU-Accelerated Simplicial !ntegrator (\gasp), builds upon the idea that we can \textit{compile once}  the \ref{eq:wmi} integral into a highly-parallelizable computational graph $I_{\phi}(\boldsymbol{\lambda})$ that is a function over the polynomial parameters $\boldsymbol{\lambda}$, and reuse this compiled function to amortize the evaluation of the denominator of \ref{eq:pal} for every datapoint $\vx$.
In fact, we can rewrite the WMI integral over a convex polytope $\mu$ as a sum of integrals over monomials:
\begin{align*}
    I_{\phi}(\boldsymbol{\lambda}) = &\int q(\vy ; \vlambda) \Ind{\vy \models \mu} d\vY \\
    = &\sum\nolimits_i \lambda_i \int \prod_j y_j^{\alpha_{i,j}} \Ind{\vy \models \mu} d\vY = \sum\nolimits_i \lambda_i \eta_i.
\end{align*}
By treating $\boldsymbol{\lambda}$ as symbolic variables, we have no dependence on $\vX$ anymore, and we recover $I_{\phi}(\boldsymbol{\lambda})$ as a symbolic polynomial whose coefficients are the results of the monomial integrals $\eta_i=\int \prod_j y_j^{\alpha_{i,j}} \Ind{\vy \models \mu} d\vY$.
Note that solving all these integrals is an embarrassingly parallelizable problem (see \cref{app:gasp_symb_polynomial}).
While in principle one could use a symbolic solver such as sympy \citep{sympy}, or another exact WMI solver,
these are extremely slow in practice and won't allow \ours to scale, 
as we empirically confirmed in \cref{sec:exp_gasp}. 

To fully harness GPUs acceleration, we look at ways of further parallelizing the integration of each monomial.
An approximate solver such as Sampo~\citep{sampo} could leverage the GPU to perform rejection sampling.
However, the quality of this approximation scheme will degrade quickly with the number of dimensions and complexity of $\phi$, as observed in \cref{sec:exp_gasp}.

\begin{algorithm}[!t]
    \caption{$\textsf{GASP!}(q, \vH)$}\label{alg:gasp_overview}
    \label{alg:gasp}
    \textbf{Input} polynomial $q$ and a convex polytope $\vH $\newline
    \textbf{Output} The Integral $\int_{\vH} \hat{q}(\vy)d\vY$
    \begin{algorithmic}[1]

    \STATE $d \gets \mathsf{totalDegree}(q)$ \;
    \STATE  $(\vR, \vw) \gets \mathsf{prepareGrundmannM\mathrm{\ddot{o}}ller}(d, \mathsf{dim}(\vH))$
    \STATE \COMMENT{points and weights of the cubature, see algorithm \ref{alg:prepare_gm_cubature}}
    \STATE $\vV \gets \mathsf{HtoVDescription}(\vH)$
    \STATE \COMMENT{turns inequalities into vertices spanning $\vH$}
    \STATE $\vS \gets \mathsf{Triangulate}(\vV)$ \COMMENT{obtain simplices}
    \RETURN $\mathsf{GASPCubature}(\hat{q}, (\vR,\vw), \vS)$
    \end{algorithmic}
\end{algorithm}

\begin{algorithm}[!t]
   \caption{$\mathsf{GASPCubature}(q, (\vR,\vw), \vS)$}
   \label{alg:gasp_integrate}
   \textbf{Input} a polynomial $q$, cubature points $\vR \in \mathbb{R}^{l_{gm} \times n}$ and weights $\vw \in \mathbb{R}^{l_{gm}}$, simplices $\vS \in \mathbb{R}^{l_s \times (n + 1)}$\newline
   \textbf{Output} the integral $\sum_i \int_{s_i} \hat{q}(\vy)d\vY$

    \begin{algorithmic}[1]
    \STATE $r_{gasp} \gets \mathsf{Array}(\mathsf{Length}(\vS))$
    \FOR[this loop runs in parallel]{$i \gets 1$ \textbf{to} $\mathsf{Length}(\vS)$}
    \STATE $\vr_{gm} \gets \mathsf{array}(\mathsf{length}(x))$
    \STATE $vol \gets \mathsf{volume}(\vS[i])$
    \FOR{$j \gets 1$ \textbf{to} $\mathsf{length}(\vR)$}
    \STATE \COMMENT{this loop runs batched (e.g., $512$ points)}
    \STATE $\vx \gets \mathsf{coordinateChange}(\vR[j], \vS[i])$
    \STATE \COMMENT{transform from unit simplex to $s_i$}
    \STATE $\vr_{gm}[j] \gets w[j] \cdot \mathsf{polyEval}(\hat{q}, \vx)$
    \STATE \COMMENT{monomials are evaluated in parallel in $\mathsf{polyEval}$}
    \ENDFOR
    \STATE $\vr_{gasp}[i] \gets vol \cdot \mathsf{stableSum}(\vr_{gm})$
    \ENDFOR
   \RETURN{$\mathsf{stableSum}(\vr_{gasp})$}
   \end{algorithmic}
\end{algorithm}

Our \gasp  pushes parallelism to the limit and 
consists of a further decomposition in sub-problems (\cref{alg:gasp}) which in turn are run as small numerical quadrature tasks that can fully leverage the tensor operations of a GPU (\cref{alg:gasp_integrate}).
In a nutshell, we aim to integrate monomials via the Grundmann and M{\"o}llers integration formula~\citep{doi:10.1137/0715019}, which is a simple cubature rule that however operates on the unit-simplex.

Therefore, we first decompose each convex polytope $\mu$ into simplices using Delaunay triangulation\footnote{We use the QHull library~\citep{10.1145/235815.235821}.} (L$5{-}7$ in \cref{alg:gasp}).
Then we transform all these simplices in unit-simplices (L$5$,$8$ and $12$ in \cref{alg:gasp_integrate}).
We finally apply the Grundmann and M{\"o}llers cubature rule which reduces to the evaluation of the (transformed) monomials at specific points on the unit-simplex, summing them according to some precomputed weights. 
This cubature rule guarantees exact integration by generating a number of points that is a function of the polynomial degree. 
\cref{fig:gasp_pipeline} provides an overview of the full integration pipeline in \gasp.
Note that \gasp can be used also as a \textit{standalone} numerical integrator for polynomials, e.g., when there is no neural network $f_{\boldsymbol{\psi
}}$ and the coefficients $\boldsymbol{\lambda}$ are constants.

\textbf{Complexity.}
Integrating an arbitrary polynomial on a convex polytope is an NP-hard task, which however becomes tractable if the dimensionality is bounded while the polynomial degree and the dimension of the simplex are allowed to vary \citep{baldoni2011integrate}.
Our scalability will therefore depend on the complexity of the constraint $\phi$ through the number of simplices, on the polynomial degree through the number of monomials and cubature points, and finally on the number of dimensions we integrate over.
\cref{app:gasp_analysis_overall} provides an in-depth discussion of how these parameters impact \gasp. Additionally, we expand on the overall complexity of \ours in relationship to the constraint $\phi$ and w.r.t. the number of bins for the piecewise spline.
We remark that we have to run \gasp \textit{only once}, before training, and we can reuse the computational graph $I_{\phi}(\boldsymbol{\lambda})$ for all datapoints $\vx$, greatly amortizing computation (see \cref{fig:numerical_integration}).

\section{Probabilistic Reasoning with PAL }

As \gasp allows us to feasibly marginalize out all variables $\vY$ in order to compute the \ref{eq:wmi} integral, it allows us also to compute arbitrary marginals as we integrate out a subset of $\vY$.
This enables \ours to query for the probability of different events defined over the labels.
Specifically, it crucially allows us to answer other probabilistic reasoning tasks at \textit{test time}, such as exactly computing the probability of satisfying (or violating) a \textit{new} constraint, e.g., 
the area around a new obstacle that just appeared on a map like \cref{fig:traj-density}.
This can be easily done by computing $\mathsf{Pr}(\phi\wedge\gamma)/\mathsf{Pr}(\phi)$ where $\mathsf{Pr}(\phi)$ is the usual \ref{eq:wmi} integral, $\phi\wedge\gamma$ the conjunction of the old SMT constraint $\phi$ with a new one $\gamma$ (e.g., representing the obstacle) over which one can compute the updated WMI probability $\mathsf{Pr}(\phi\wedge\gamma)$.

\textbf{Sampling with \gasp} As \gasp allows us to marginalize out all variables $\vY$ in order to compute the \ref{eq:wmi} integral, it allows us also to compute arbitrary marginals as we integrate out a subset of $\vY$.
This in turn enables us to draw samples from \ref{eq:pal} that naturally satisfy the constraint $\phi$ without rejection but using inverse transform sampling.
Given a variable ordering $Y_1, Y_2,\ldots,Y_n$, we can \emph{iteratively}
sample $y_1 \sim p(Y_1|\vx)$ and $y_i \sim p(Y_i|y_{<i},\vx)$ for $i \in \{2,...,n\}$, where each sampling step requires numerically inverting the corresponding conditional cumulative distribution function (another WMI integral computable with \gasp), which can be efficiently done via binary search.
This procedure is detailed in \cref{app:sampling}.

\section{Related Work}
\label{sec:related-works}

Many probabilistic NeSy~\citep{nesysurvey} approaches try to satisfy constraints only \emph{in expectation}.
These include incorporating the constraints as a loss term, applied to various formalisms such as propositional logic~\citep{semanticloss1,semanticloss2}, fuzzy logic~\citep{sbr}, physical laws~\citep{DBLP:conf/aaai/StewartE17} or other algebraic constraints on the outputs of the NNs~\citep{dl2}. 
Compared to \ours, these approaches do not guarantee the satisfaction of the constraints.

A series of other works, instead, guarantees the satisfaction of constraints by \textit{embedding} them in the network architecture. 
For example, \citet{cdgm,pishield} take a similar approach to \ours, by adding a constraint layer on top of NNs, which however is limited to constraints in the form of conjunctions of inequalities.
\citet{DBLP:journals/corr/abs-2502-18237} goes beyond this limitation and devise a layer that projects samples into non-convex constraints.
Differently from us, these layers do not yield densities or probabilities associated to the sampled points or to given constraints.
However, their sampling schemes scale much better than autoregressive sampling with \ours.
Multiplexnet~\citep{hoernle2022multiplexnet} introduces a layer restricted to constraints in disjunctive normal form, which must be relaxed during learning. As a result, it struggles to scale to the intricate constraints in our setting.
DeepSade~\citep{deepsade} takes a different route, pairing gradient descent with constrained optimization to guarantee SMT constraints for classification and regression tasks. 
This approach supports SMT formulas with quantifiers, however, its extension to the probabilistic setting is highly non-trivial.
Our work is inspired by semantic probabilistic layers (SPL)~\citep{spl}, which combines neural networks with probabilistic circuits \citep{loconte2025relationshiptensorfactorizationscircuits} in the propositional logic case.
We generalize SPL to constraints involving both logical and continuous variables.
Similar to SPL, DeepProbLog~\citep{deepproblog} uses probabilistic logic programming to create a layer, but considers only Boolean logic and fully-factorized distributions \citep{van2024Independence}. 

As discussed in \cref{sec:background}, closer to our work is DeepSeaProbLog (DSP)~\citep{deepseaproblog}, which extends DeepProbLog to continuous domains. 
In contrast to \ours, in their implementation, densities are not guaranteeing constraint satisfaction.
In fact, DSP is trained to maximize the probability of the constraint being satisfied, i.e., the \ref{eq:wmi} integral, via sampling. 
We detail the \ref{eq:dsp_loss} in the appendix.
As a side effect, one cannot avoid the rejection layer in DSP as samples directly drawn from the unconstrained probability of a DSP program can violate the constraints.
\cref{app:related-works} discusses additional related works.

\section{\scalebox{1.1}{\ours} in Action}
\label{sec:experiments}

In this section, we aim to answer the following research questions:\footnote{The source code and the instructions to reproduce our results
can be found at: \href{https://github.com/april-tools/pal}{github.com/april-tools/pal}}
\textbf{RQ1} How does \gasp compare with other integration solvers?
\textbf{RQ2} Can \ours learn and scale with an increasing number of constraints?
\textbf{RQ3} Can \ours handle real-world data and its constraints?

\subsection{RQ1) Benchmarking \scalebox{1.1}{\gasp}}
\label{sec:exp_gasp}

\begin{figure}
    \begingroup
    \setlength{\tabcolsep}{1pt} %
    \begin{tabular}{c c c}
        \small  \hspace{20pt}$d=8$ & \small $d=12$ & \small amortized \\
        \includegraphics[height=3.42cm]{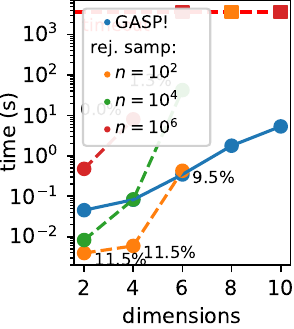}
        &  \includegraphics[height=3.42cm]{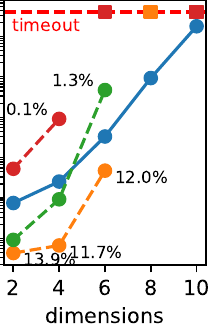}
        &  \includegraphics[height=3.42cm]{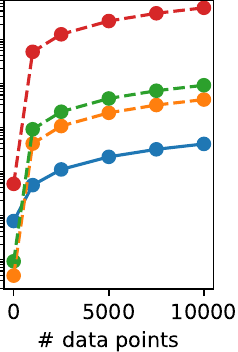}
    \end{tabular}
    \endgroup
    \caption{\emph{Left} and \emph{center}: runtime of \gasp and rejection sampling (using $10^{\{2,4,6\}}$ samples) on integrals over $4$ random simplices for increasing dimensions. We report the relative error ($\%$) for rejection sampling. \emph{Right}: Our amortization speed-up compared to rejection-sampling in $2$-dimensions for the $12$-degree polynomial.
    \cref{fig:numerical_integration_all} shows more results for different degrees and constraints.
    }
    \label{fig:numerical_integration}
\end{figure}

We compare \gasp first against approximate numerical schemes such as rejection sampling, which is commonly used to scale probabilistic NeSy approaches such as DSP (\cref{sec:related-works}), and then against SoTA exact polynomial integrators such as \latte \citep{latte}.

\textbf{\gasp vs numerical approximations for \ours.}
We evaluate \gasp w.r.t. an implementation of rejection sampling that runs on the GPU,  on random non-convex integration problems of increasing dimensionality and degree. 
\cref{app:rejection_vs_gasp} details this experimental setting and \cref{fig:numerical_integration_all} reports all results.
Here we show in \cref{fig:numerical_integration} (\emph{left} and \emph{center}) how rejection sampling struggles to obtain accurate results in higher dimensions for a fixed time budget (1 hr) and polynomials of degree 8 and 12, as the number of rejected samples grow exponentially.
Notably, the complexity of the constraints is fixed in this experiment to 4 random simplices. We expect rejection sampling to perform much worse with increasingly complex regions.
\gasp instead provides exact computation and scales better (notice that the y-axis is log-scale).
More crucially, \gasp can amortize computation as the compiled polynomial $I_{\phi}(\boldsymbol{\lambda})$ can be reused throughout the training of \ours.
\cref{fig:numerical_integration} (\emph{right}) illustrates this for a degree $12$ polynomial in $10$ dimensions: integrating over a single convex polytope for $10^{4}$ predictions, i.e. $10^{4}$ different values for $\boldsymbol{\lambda}$, takes less then 10 seconds with \gasp thanks to the amortization and up to 1 hour with rejection sampling.
This is fundamental to use \gasp in \ours, as the number of evaluations of \ref{eq:wmi} integrals can easily be orders of magnitude larger than that.

\begin{figure}[!t]
\begin{minipage}{.25\textwidth}
\centering
\includegraphics[width=0.9\textwidth]{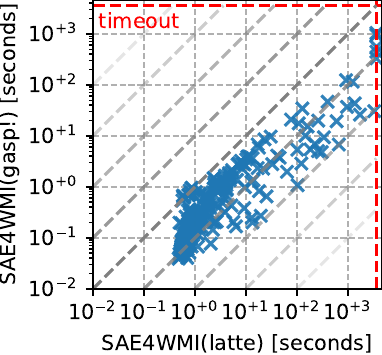}    
\end{minipage}\hspace{5pt}\begin{minipage}{.2\textwidth}
\caption{
\label{fig:gasp_vs_latte}
\textbf{\gasp can be 1 to 2 orders of magnitude faster than a SoTA polynomial integrator} such as \latte when evaluated on standard WMI benchmarks from \citet{sae4wmi}
}    
\end{minipage}
\end{figure}

\textbf{\gasp as a stand-alone integrator.}
We compare \gasp against \latte, a SoTA exact polynomial integrator  based on cone decomposition~\citep{latte}, in the context of WMI integration (details in \cref{app:latte_vs_gasp}).
We adopt SAE4WMI (also used in \ours, see \cref{sec:gasp}), and compare it while using either \gasp or \latte on 270 \ref{eq:wmi} instances of varying complexity taken from ~\citet{sae4wmi}.
Compared to the previous experiment, these WMI problems are highly non-convex, requiring the computation of up to
hundreds of integrals each. 
Notably, here \gasp cannot harness its amortization capabilities and needs to instantiate a different computational graph every time.
Despite this, \gasp  proves faster than \latte in 263 instances over 270, speeding computation up to one or two orders of magnitude, as shown in \cref{fig:gasp_vs_latte}.
Lastly, and as a sanity check, we compare \gasp against classical symbolic integrators such as sympy in \cref{sec:appendix_gasp_vs_xadd}.
While we retrieve the same computation, we achieve a speed up of $1200$ times. 
Overall, our results confirm that \gasp is the best available solver for \ours, positively answering \textbf{RQ1}.

\subsection{RQ2) Scaling constraints with \scalebox{1.1}{\ours}}
\label{sec:nstar}

\begin{figure}
\centering
    \begin{tabular}{ccc}
        \multicolumn{1}{c}{\small ground truth} & \multicolumn{1}{c}{\small NN+\ours} & \multicolumn{1}{c}{\small NN + GMM}\\
        \includegraphics[width=0.25\columnwidth]{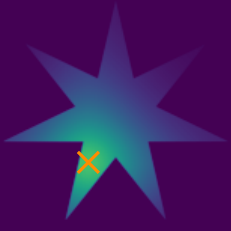}
        & \includegraphics[width=0.25\columnwidth]{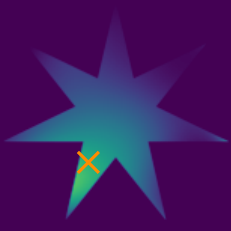}
        & \includegraphics[width=0.25\columnwidth]{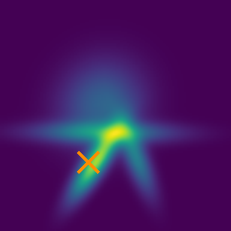}
        \\
    \end{tabular}
\hfill
    \caption{While \ours naturally handles constraints on the $N$-Star, the neural network + GMM has trouble fitting both constraints and data. 
    More quantitative results in \cref{tab:result_nstar} and qualitative ones in
    \cref{sec:appendix_experiments_nstar_details}. %
    }
    \label{fig:nstar_comparison}
\end{figure}

In this experiment, we evaluate \ours in a controlled setting to explore how accurately learning in \ours scales when increasing the number of constraints, while keeping the dimensionality fixed.
The task is learning the distribution $p(Y_1, Y_2 \mid X_1, X_2)$ constrained by a $N$-pointed star, as shown in \cref{fig:nstar_comparison}, for $N=7$ whose unconstrained distribution is a Cauchy with constant scale and mode $\vX$.
\cref{tab:result_nstar}  reports the average test log-likelihood of \ours versus an unconstrained NN, a mixture density network  \citep{bishop1994mixture} with a Gaussian mixture model (GMM) with different components ($K$) as output, and DSP.
All models use the same fully-connected NNs with ReLUs as a backbone. 
\cref{sec:appendix_experiments_nstar_details} further details our setting.

As $N$ increases, the gap between \ours and our the competitors widens as the mode tends to be closer to the extremes of the feasible region. 
Being agnostic to the constraint $\phi$, the NN+GMM has an hard time fitting both the data and respecting the constraints.
The same goes for DSP which is hindered by having to fit a single multivariate Gaussian while maximizing the mass inside $\phi$, i.e., $\mathsf{Pr}(\phi)$, see \ref{eq:dsp_loss}.
This has often the unwanted effect of pushing the learned mode far from the target mode. 
This aspect prompted us to report the performance of DSP selecting the model according to best log-likelihood on holdout data as opposed to the standard best loss criterion.
\ours in comparison has an easier time, it naturally handles constraints and therefore just has to move the probability mass around, prompting us to answer \textbf{RQ2} positively.

\begin{table}
\begingroup
\setlength{\tabcolsep}{3pt}  %
\centering
\small
\caption{\textbf{\ours  can be more accurate than unconstrained networks and other NeSy baselines} in terms of 
average test log-likelihood on the NStar-dataset. %
We report the mean over $10$ repetitions, more results in \cref{sec:appendix_experiments_nstar_details}.
\label{tab:result_nstar}
}
\begin{tabular}{lrrrrrr}
\toprule
 & \multicolumn{2}{c}{NN + \ours} & \multicolumn{2}{c}{NN + GMM} & \multicolumn{2}{c}{DSP} \\
\cmidrule(lr){2-3} \cmidrule(lr){4-5}\cmidrule(lr){6-7}
$N$ & $d=10$ & $d=14$ & $K{=}8$ & $K{=}32$ & by logLike & by Loss \\
 \midrule
3 & -4.749 & \bfseries -4.674 & -4.740 & -4.723  & -5.027 & -42.821 \\
7 & -4.529 & \bfseries -4.527 & -4.708 & -4.612 & -5.019 & -206.411 \\
11 & \bfseries -4.570 & -4.584 & -4.791 & -4.620 & -83.042 & -43.151 \\
19 & -4.506 & \bfseries -4.492 & -4.925 & -4.652 & -5.001 & -155.139 \\
\bottomrule
\end{tabular}
\endgroup
\end{table}

\subsection{\textbf{RQ3)} Stanford Drone Dataset}
\label{sec:stanford-exp}

After showing \ours scalability to many constraints, we now evaluate it on real-world data where the ground truth is unavailable.
As the majority of existing approaches are unable to process non-convex constraints (\cref{sec:related-works}), established  benchmarks are lacking. 
We fill this gap by introducing a new, challenging task based on the Stanford drone dataset (SDD) \citep{DBLP:conf/eccv/RobicquetSAS16}, which shoes aerial views of pedestrians, cars and bikes traversing different scenes in the Stanford campus. \footnote{The dataset can be found at \href{https://github.com/april-tools/constrained-sdd}{https://github.com/april-tools/constrained-sdd}}
We extract trajectories as in \citet{DBLP:journals/pami/WuLZC23}
and we manually annotate two scenes, number 12, which contain the most trajectories, and 2, which contains highly non-convex constraints.

Specifically, we create constraints in SMT by segmenting the images over the areas that are non-walkable.
\cref{fig:sdd_constraints} shows an example of an intersection from scene 12 and the extracted constraints.
We evaluate two settings: where we model all trajectories together, and when the task is to probabilistically predict the future position given an observed partial trajectory.
In contrast to RQ1 and RQ2, the probability mass in the SDD is less spread out and more localized at regions of high density. This setting is less suited to be modeled by a single polynomial and more applicable to a piecewise construction. We therefore choose Hermite splines because they are easy to train and scale well with our dataset. The exact construction of the splines is described in \cref{app:splines}.

\begin{figure}[!t]
\hfill
\begin{minipage}{0.32\columnwidth}
    \centering
    scene
    \includegraphics[width=0.9\textwidth]{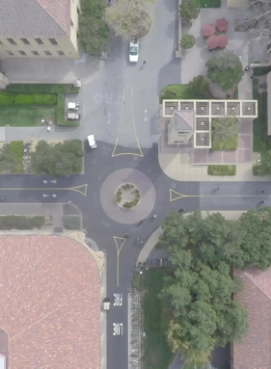} %
\end{minipage}
\hfill
\begin{minipage}{0.32\columnwidth}
    \centering
    trajectories
    \includegraphics[width=0.9\textwidth, keepaspectratio]{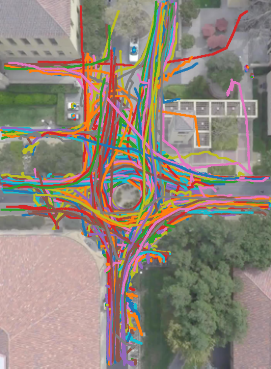}
\end{minipage}
\hfill
\begin{minipage}{0.32\columnwidth}
    \centering
    constraints
    \includegraphics[width=0.9\textwidth, keepaspectratio]{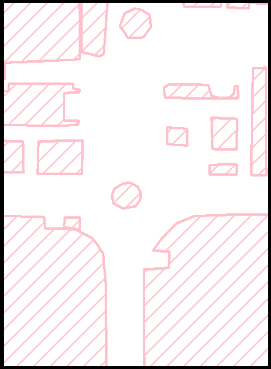}
\end{minipage}
\caption{\textbf{Our dataset combines challenging constraints with real-world data} on trajectories (\textit{middle}) and aerial maps (\textit{left}) taken from \citet{DBLP:conf/eccv/RobicquetSAS16}. We manually label the data, segmenting invalid areas out (\textit{right}).}
\label{fig:sdd_constraints}
\end{figure}

\textbf{Modeling joint trajectories ($p(\vY)$).}
We tackle the problem of estimating the joint distribution of all trajectories, i.e., $p(\vY)$. 
This means we optimize  \ours as a standalone distribution estimator,  without any neural backbone. 
\cref{sec:appendix_experiments_sdd_marginal} details our experimental setting, in which we compare \ours with polynomial splines of increasing complexity against a GMM with increasing number of components and finally a neural spline flow \citep{DBLP:conf/nips/DurkanB0P19} with multiple transformation layers.
\ours is able to achieve competitive test log-likelihoods w.r.t. similarly sized GMMs and much larger flows (with one order of magnitude more parameters), as reported in \cref{tab:result_sdd_marginal}.
More crucially, \ours never places probability mass outside the given constraint, while the other competitors do so as shown in \cref{fig:traj-density} and under $p(\neg\phi)$ in \cref{tab:result_sdd_marginal}, denoting the probability of bumping into an obstacle, here approximated with $10^{6}$ samples for the GMM and the flow.
Note that violating the constraint less than 2\% of the time can still be greatly harmful for safety-critical applications.
In summary, \ours is able to guarantee constraint satisfaction while not compromising accuracy, nor time. 
In fact, \gasp takes only $17$ seconds on an NVIDIA RTX A6000 to integrate the largest polynomial we consider ($d{=}12$) on the intricate constraint in \cref{fig:sdd_constraints} (right).

\begin{table}[!t]
\begingroup
\setlength{\tabcolsep}{1pt}  %
\caption{\textbf{\ours does not trade expressiveness for constraint-satisfaction} when compared against a  GMM and a neural spline flow with $t{=}1$ and $t{=}5$ transformations, as it provides competitive average test log-likelihood but never violates constraints for test-set predictions ($\mathsf{Pr}(\neg \phi)$). We report the mean over $10$ repetitions, more results in \cref{sec:appendix_experiments_sdd_marginal}. The setting is the unconditional $P(\vY)$-case.
\label{tab:result_sdd_marginal}}
\scalebox{.85}{
\begin{tabular}{ll rr rr rr}
\toprule
 & & \multicolumn{2}{c}{\ours} & \multicolumn{2}{c}{GMM} & \multicolumn{2}{c}{Flow}\\
 \cmidrule(lr){3-4} \cmidrule(lr){5-6} \cmidrule(lr){7-8} 
 & & 10 knots & 16 knots & $K{=}50$ & $K{=}100$ & $t{=}1$ & $t{=}5$\\
scene & (params) & (410) & (650) & (300) & (610) & (2670) & (13350)\\
 \midrule
\textbf{1} & ll & $-2.98$ & $-2.93$ & $-2.98$ & \boldmath $-2.91$ & $-3.10$ & $-2.94$\\
& ${\mathsf{Pr}(\neg \phi)}$ & \boldmath $0.0\%$ & \boldmath $0.0\%$ & ${\approx} 2.3\%$ & ${\approx} 1.2\%$ &  ${\approx} 5.6\%$ &  ${\approx} 1.6\%$\\
\midrule
\textbf{2} & ll & $-3.35$ & $-3.30$ & $-3.34$ & \boldmath $-3.26$ & $-3.43$ & $-3.26$\\
& ${\mathsf{Pr}(\neg \phi)}$ & \boldmath $0.0\%$ & \boldmath $0.0\%$ & ${\approx} 1.2\%$ & ${\approx} 0.6\%$ &  ${\approx} 2.2\%$ &  ${\approx} 0.7\%$\\
\bottomrule 
\end{tabular}
}
\endgroup
\end{table}

\begin{figure}[!t]
\centering
\begin{minipage}{0.40\columnwidth}
    \centering
    NN+\ours (10 knots)
    \includegraphics[width=0.9\textwidth]{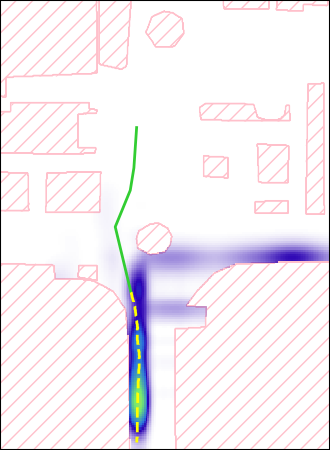} 
\end{minipage}
\hspace{20pt}
\begin{minipage}{0.40\columnwidth}
    \centering
    NN+GMM ($K=80$)
    \includegraphics[width=0.9\textwidth]{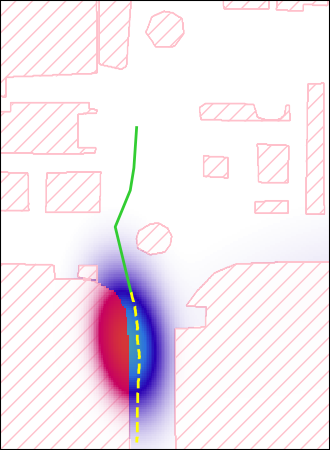}
\end{minipage}
\hfill
\caption{\textbf{\ours captures meaningful modes when modelling how a trajectory might evolve ($p(\vY \mid \vX)$) while never violating the constraint} differently from a NN with a GMM output. The observed test trajectory and ground-truth on scenario 1 completion are reported in green/solid and dashed/yellow.
Additional plots in \cref{tab:sdd_conditional_predictions_scene_1}.
}
\label{fig:sdd_example_predictions}
\end{figure}

\textbf{Modeling future positions ($p(\vY \mid \vX)$).}
After showing that our layer can effectively model the joint distribution, we consider estimating the distribution over future positions given some observations on the current trajectory.
To this end, we subsample five equidistant points from a trajectory that we consider as input variable $\vX$, and we predict the probability of the model of being in any other point $Y_1, Y_2$ in the map. 
\cref{app:sdd_conditional} completely details our stoetting.

Figure \ref{fig:sdd_example_predictions} and \ref{fig:sdd_example_predictions_image_2} depict the conditional distributions modeled by \ours and NN+GMM for a single observed trajectory.
As expected, multiple future trajectories are visible in the distribution modelled by \ours. 
In contrast, the conditional GMM is less good at capturing the multimodality of $p(\vY \mid \vX)$, while also clearly violating the constraints.
\cref{tab:result_sdd} reports a quantitative comparison of NN+\ours with NN+GMM and DSP (finer grained results are reported in \cref{tab:sdd_traj_big_table} and \ref{tab:sdd_traj_big_table_image_2}) showing 
the average log-likelihood of the points in the (unobserved) trajectory completions and the average probability of sampling future positions that violate $\phi$ (estimated using $10^6$ samples for NN+GMM and DSP).
We do not compare against flows in this setting, as parameterized them with a neural network turned out infeasible: it would have required millions of parameters just to realize a linear gating function $g$ (\cref{sec:poly}).
Results in \cref{tab:result_sdd} show that the constraint violation is on average around the 18\% for the baselines across all trajectories. This value raises up to 70\% for the GMM model on certain trajectories.
We highlight that the probability of violating the constraint is higher for conditional predictions because of less data: having direct access to the constraint $\phi$ greatly improve data efficiency for \ours.

\begin{figure}
\hfill
\begin{minipage}{0.40\columnwidth}
    \centering
    NN+\ours (14 knots)
    \includegraphics[width=0.95\textwidth]{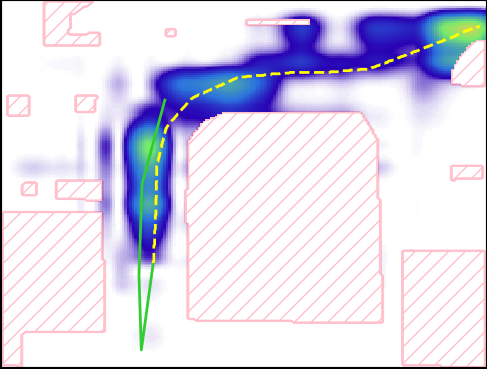} 
\end{minipage}
\hfill
\begin{minipage}{0.40\columnwidth}
    \centering
    NN+GMM ($K{=}32$)
    \includegraphics[width=0.95\textwidth]{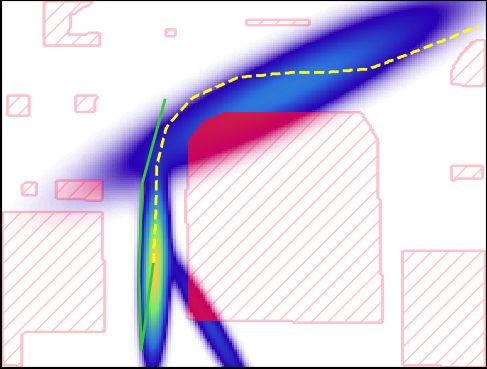}
\end{minipage}
\hfill
\caption{\textbf{\ours wraps around the boundaries of our constraints}, in contrast to the NN with an GMM, which predicts a trajectory straight trough the constraint. The observed test trajectory and ground-truth completion on scenario 2 are reported in green/solid and dashed/yellow respectively.}
\label{fig:sdd_example_predictions_image_2}
\end{figure}

All in all, these results show that \ours, both alone and combined with NNs and  in contrast to the other baselines, show promise in effectively modelling complex real world distributions,
without trading off expressiveness for constraint satisfaction.
We answer \textbf{RQ3} positively.

\begin{table}[!t]
\begingroup
\caption{\textbf{\ours shows competitive likelihoods while guaranteeing constraint-satisfaction} when compared to neural GMM and DSP, for which we provide statistics for the fitted neural distributional fact. We approximate the probability of violating the constraints ($\mathsf{Pr}(\neg \phi)$) at test time numerically per data point and average it. We report the mean over $10$ repetitions, further details in \cref{tab:sdd_traj_big_table} and \cref{tab:sdd_traj_big_table_image_2}. The setting is the conditional $P(\vY|\vX)$-case.
\label{tab:result_sdd}}
\setlength{\tabcolsep}{2pt} %
\begin{tabular}{ll rr rr r}
\toprule
 & & \multicolumn{2}{c}{NN + \ours} & \multicolumn{2}{c}{NN + GMM} & DSP\\
  \cmidrule(lr){3-4} \cmidrule(lr){5-6} \cmidrule(lr){7-7} 
scene & & 10 knots & 14 knots & $K{=}50$ & $K{=}100$ & \multicolumn{1}{c}{-}\\
\midrule
\textbf{1} & ll & \boldmath $-2.08$ & $-2.27$ & $-2.64$ & $-2.83$ & $-3.87$\\
& $\mathsf{Pr}(\neg \phi)$ & \boldmath $0\%$ & \boldmath $0\%$ & ${\approx} 21\%$ & ${\approx} 20\%$ & $\approx 49\%$ \\
\midrule
\textbf{2} & ll & $-2.23$ & \boldmath $-2.09$ & $-2.39$ & $-2.42$ & $-3.61$\\
& $\mathsf{Pr}(\neg \phi)$ & \boldmath $0\%$ & \boldmath $0\%$ & ${\approx} 15\%$ & ${\approx} 14\%$ & $\approx 36\%$ \\
\bottomrule
\end{tabular}
\endgroup
\end{table}

\section{Conclusion}

In this work, we introduced \ours, a probabilistic NeSy layer that can be plugged as the prediction layer in any neural network that has to deal with multiple continuous labels and in the presence of algebraic constraints over them.
To scale \ours to real-world data, we had to advance the field of WMI by proposing \gasp, a parallelizable polynomial integrator that challenges even established scientific software such as \latte reaching a speed-up of up to one or two orders of magnitude.
Furthermore, \ours offers flexible marginalization over new constraints at test time.
While this also allows us to use \gasp to sample from \ours, it does not allow us to easily tackle maximum a posteriori (MAP) in a similar way. How to compute MAP-style queries for \ours remains an open research question \citep{zeng2021parameter}.
In the future, we plan to investigate and scale \gasp further as a standalone software for a number of computationally intense applications using polynomials such as inference in Bayesian  \citep{zeng2023collapsed} and physics informed neural networks \citep{lu2021physics}.

At the same time, \ours offers a number of interesting future directions such as enforcing constraints in several applications ranging from fairness \citep{DBLP:conf/l4dc/PfrommerGZS22, DBLP:journals/corr/abs-2401-03799} to climate modeling \citep{PhysRevLett.126.098302, DBLP:journals/jmlr/HarderHRYSSWR23, willard2020integrating, beucler2020towards} and 
probabilistic verification \citep{morettin2024unified}.
Furthermore, we plan to extend \ours to deal with more types of constraints than \smtlra. Building on the work of \citet{chin2020efficient}, we aim to extend \gasp to handle parametric curved boundaries, such as B-splines. This would then unlock further applications, e.g. in engineering, as it allows us to apply \ours to domains defined by NURBS surfaces, a common format to describe shapes in engineering design.

\begin{contributions} %
AV, RS and AP discussed to extend SPL to algebraic constraints and
LK and AV conceptualized a way to do so, which resulted in \ours.
LK is responsible for coming up with \gasp and for the design and implementation of all experiments and plots with the exception of experiment \ref{sec:exp_gasp}, which was contributed by PM. 
AV, PM and LK wrote the paper. 
AV supervised and provided feedback for all the phases of the project  with help from RS and AP.
\end{contributions}

\begin{acknowledgements} %
We are grateful to Lennert De Smet and Pedro  Zuidberg Dos Martires for insightful discussions regarding both the experimental evaluation and the general direction of the project, particularly in comparing DeepSeaProbLog and \ours. Additionally, we thank Adri\'{a}n Javaloy, Gennaro Gala, Andreas Grivas for valuable feedback on the draft. AV was supported by the “UNREAL: Unified Reasoning Layer for Trustworthy ML” project (EP/Y023838/1) selected by the ERC and funded by UKRI EPSRC. 
PM was supported by the MSCA project “Probabilistic Formal Verification for Provably
Trustworthy AI - PFV-4-PTAI” under GA no. 101110960.
RS was supported in part by the MUR PNRR project FAIR - Future AI Research (PE00000013) funded by the NextGenerationEU. AP and RS were supported by the TANGO project funded by the EU Horizon Europe research and innovation program under GA No 101120763.
Funded by the European Union.
RS was supported in part under the NRRP, Mission 4 Component 2 Investment 1.4, by the European Union - nextGenerationEU (proj. nr. CN 00000013).
Views and opinions expressed are however those of the author(s) only and do not necessarily reflect those of the European Union, the European Health and Digital Executive Agency (HaDEA) or The European Research Council. 
Neither the European Union nor the granting authority can be held responsible for them.

\end{acknowledgements}

\bibliography{referomnia}

\newpage

\onecolumn

\title{A Probabilistic Neuro-symbolic Layer for Algebraic Constraint Satisfaction\\(Supplementary Material)}
\maketitle

\appendix

\section{Background on Weighted Model Integration}
\label{app:wmi}

The task of marginalizing over a distribution defined by a density over \smtlra constraints is known as \emph{weighted model integration}~\citep{wmi}.
WMI generalizes weighted model counting (WMC), i.e. the task of summing over the models of a propositional logic formula, to the hybrid logical/continuous domain.
In WMC, each satisfying truth assignment $\mu$ is a model, whose weight typically factorizes over the literals:
\begin{align*}
    \mathsf{WMC}(\phi, w) = \sum_{\mu \models \phi} \: \prod_{\ell \in \mu} w(\ell)
\end{align*}
In WMI, each $\mu$ induces a convex (and disjoint) subregion of $\phi$. For $\mathsf{WMI}(\phi, w)$ to be finite, $\phi$ must encode a closed region. In those cases, each $\mu$ induces a convex polytope containing infinitely many models, i.e. assignments $\vx \models \mu$. The weight function $w$ can be interpreted in probabilistic terms as an unnormalized density over $\vX$.
Then, obtaining the weight of each $\mu$ additionally requires integrating $w$ over those models:
\begin{align}
    \label{eq:wmi-full}
    \mathsf{WMI}(\phi, w) 
    &= \underbrace{\sum_{\mu \models \phi}}_{\text{(1)}} \underbrace{\int w(\vx) \Ind{\vx \models \mu} \:d\vX}_{\text{(2)}}
\end{align}
Computing WMI requires solving two subtasks:
\begin{itemize}
    \item[(1)] \emph{enumerating} all the $\mu \models \phi$;
    \item[(2)] \emph{integrating} $w$ inside each $\mu$.
\end{itemize}
For our purposes, WMI exactly corresponds to the problem of computing the normalizing constant in Eq.~\ref{eq:pal}:
\begin{align}
    \label{eq:z-as-wmi}
    \int w(\vx) \Ind{\vx \models \phi} \:d\vX= \sum_{\mu \models \phi} \int w(\vx) \Ind{\vx \models \mu} \:d\vX
\end{align}

\textbf{Partial enumeration.} In \gasp, we solve Eq.~\ref{eq:z-as-wmi} by decomposing the integral into convex polytopes first and then further decomposing each polytope into simplices. Clearly, obtaining a compact decomposition of $\phi$ into disjoint convex regions is paramount for reducing the size of the computational graph. In this work, we leverage the enumeration procedure of SAE4WMI~\citep{sae4wmi} for solving subtask (1).
SAE4WMI builds upon a line of work that leverages advanced SMT techniques~\citep{wmipa,wmisapa} for minimizing the number of integrations.
A key idea of these solver is the enumeration of \emph{partial} (as opposed to \emph{total}) satisfying truth assignments.
For the purpose of WMI, the set of satisfying truth assignments $TA(\phi)$ must be complete ($\bigvee_{\mu \in TA(\phi)} \mu \equiv \phi$) and it must contain mutually exclusive/disjoint elements ($\forall i \neq j \:.\: \mu_i \land \mu_j \models \bot$).
As opposed to total assignments, partial assignments are not required to map every inequality of $\phi$ to $\{True, False\}$.

It is easy to show that the latter can exponentially reduce the number of partitions of $\phi$ by considering a disjunction among $N$ atomic formulas: $\phi = \bigvee_{i=1}^N A_i$.
While the size of the set of total satisfying truth assignment is $2^N - 1$, we can fully characterize the formula with $N$ disjoint partial assignments $\phi = \bigvee_{i=1}^N \mu_i$, where:
\begin{align*}
    &\mu_1 = A_1 \\
    &\mu_2 = \neg A_1 \land A_2 \\
    &\mu_3 = \neg A_1 \land \neg A_2 \land A_3 \\
    & ... \\    
    &\mu_N = \neg A_1 \land \neg A_2 \land ...  \land \neg A_{N-1} \land A_N
\end{align*}
The benefits of enumerating partial truth assignments is evident even in our lower dimensional example, as depicted in Figure~\ref{fig:partial}.

\begin{figure}
\centering

\includegraphics[height=3cm]{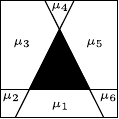} 
\phantom{of the opera} %
\includegraphics[height=3cm]{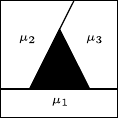}

\caption{A decomposition of the non-convex constraint in our running example with total (\emph{left}) vs. partial (\emph{right}) satisfying truth assignments.}
\label{fig:partial}
\end{figure}

\section{Training and Parametrizing Polynomials}
\label{app:polynomials}
We tested \ours with two kinds of polynomial density: single polynomials and cubic, hermite splines.

As expressiveness is tied to increasing the total degree, the former might be difficult to employ in fitting very complex distributions. 
We turned to the latter in the Stanford Drone use cases, partitioning the overall density into equally-sized bins, each encoded as the product of squared univariate splines.
Furthermore, we consider a mixture of the parametric form above.
The resulting parametric form is both stable to train and expressive, while also leveraging the amortization capabilities of \gasp for each bin.

We will first specify the exact re-parametrization and methods we used for both raw polynomials and splines, and then detail the loss.

\subsection{Polynomials}
\label{app:training_polynomials}
We experienced some difficulty during training when trying to directly predict coefficients of single polynomials, due to the different scale-sensitivity of the monomials.
We tackled this issue for the $N$-pointed star experiments in 
by adding a re-parametrization layer in front of the polynomial. 
Let $I(\vLambda) = \sum_i \sum_j \vLambda_i \vLambda_j \eta_{ij}$ be the integrated, squared polynomial. Then our re-parametrization is the following: 
$$r(z_i) = \mathtt{sign}(z_i)\frac{\sqrt{z_i + d'} - \sqrt{d'}}{\sqrt{\eta_{ii}}}$$ with $d'=0.1$.
This dampens the impact of coefficients of high-degree monomials, as they are often associated with large $\eta_{ii}$.

Additionally, before training, we initialize the magnitude of the output of our last layer by fitting a constant scalar per dimension using L-BFGS \citep{liu1989limited} over the first $1000$ training samples.

While, with these methods, we are able to train \ours for a single, high-degree polynomial in a stable manner, we want to stress that they are not needed for the spline-parametrization.

\subsection{Polynomial splines}
\label{app:splines}

We use univariate, cubic, Hermite splines \citep{smith1980practical}, which we then square to guarantee non-negativity and multiply to create multivariate splines.

Each spline is specified by the value and derivative at the knots. In order to evaluate and integrate, we create the explicit polynomial. We will now focus on a specific bin $[k_i, k_{i+1}]$. We view the spline just as a re-parametrization of a polynomial and compute the coefficients explicitly, in order to plug into the our framework to quickly compute the integral $I(\vlambda)$. In comparison to the usual way to construct splines, in which the spline is constructed on the standard-bin $[0,1]$, we have to account for scale. We therefore construct the parameters corresponding to the polynomial on the $[0, k_{i+1} - k_i]$ and shift the monomials $m$ via $m'(x) = m(x - k_{i+1})$ before handing them to \gasp. The flexibility of shifting via a coordinate transform, instead of transforming the parameters of the polynomial, which is numerically instable, shows the adaptability and versatility of \gasp. The computed parameters of \gasp belong to the polynomial on $[0, k_{i+1} - k_i]$ per bin, and are numerically significantly more stable to evaluate than explicitly constructing the parameters of the shifted polynomial on $ [k_{i}, k_{i+1}]$.

Given $k_{i}$ and $k_{i+1}$, we compute the parameters as follows, which is a straightforward adaptation of the standard construction on $[0,1]$.

First, we construct the parameters $a + b \cdot x + c\cdot x^2 + d\cdot x^3$ on $[0,1]$. With $v_i$, $v_{i+1}$ we denote the value and $v'_i$, $v'_{i+1}$ the derivative:
\begin{align*}
    a &= v_i \\
    b &= v'_i \\
    c &= 3 \cdot (v_{i+1} - v_i) - 2 \cdot v'_i - v'_{i+1} \\
    d &= 2 \cdot (v_i - v_{i+1}) +  v'_i + v'_{i+1} \\
\end{align*}
We then scale the polynomial by transforming the parameters. Let $\Delta = (k_{i+1} - k_i)$, then:
\begin{align*}
    a' &= a \\
    b' &= b \cdot (1 / \Delta) \\
    c' &= c \cdot (1 / \Delta^2) \\
    d' &= c \cdot (1 / \Delta^3) \\
\end{align*}

These are the coefficients corresponding to the unsquared polynomial. The squared polynomial (of degree $6)$ is then just the combination of these parameters with itself, just as the multiplication of two splines is a combination of the respective parameters. Additionally, we take a mixture of these splines. These form $\vlambda$.

We integrate by enumerating the bins, shifting the monomials, and then obtain the coefficients for $I(\vlambda)$ and during training, we view the piecewise spline as a re-parametrization layer that transforms the output of the neural network, so value and derivative at the knots, into the coefficients of the polynomial.

\subsection{Loss}
\label{app:loss}

For \ours, we minimize the following loss:

\begin{align}
    \label{eq:loss}
    l(\vx^{[1:b]}, \vy^{[1:b]}) &= (-1)*\underbrace{\sum_b\log p_{\boldsymbol{\Theta}}(\vy^{(b)} \mid \vx^{(b)})}_{\text{constrained log-likelihood}} + \underbrace{\sum_b \Ind{\log i_b \geq 10} (\log i_b - 10)^2}_{\text{penalty on too large values of $I(\vlambda = f_\psi(\vx^{(b)}))$}}
\end{align}

with $i_b = I(\vlambda = f_\psi(\vx^{(b)}))$. This loss biases the neural network towards more numerically-stable range of integral-values. Due to the scale-invariance, this does not influence the expressivity and only influences numerical stability. Although we use it for both raw polynomials and splines, it's main use is to stabilize training of raw polynomials as the splines are inherently easier to train.

\section{\gasp}
\label{app:appendix-gasp}

\subsection{Handling the symbolic Polynomial}
\label{app:gasp_symb_polynomial}

We start with an symbolic polynomial $q(\vy, \vlambda)$ of the form:

\begin{align*}
    q(\vy, \vlambda) &= \sum_i \lambda_i \prod_j y_j^{\alpha_{ij}} \\
    &= \sum_i \lambda_i m_i(\vy)
\end{align*}

where $m_i(\vy)$ denotes the $i$th monomial. This induces the vector-valued version of the polynomial:

\begin{align*}
    \vec{v}_q(\vy) &= \begin{pmatrix}
        m_{i_1}(\vy) \\
        m_{i_2}(\vy) \\
        \vdots \\
        m_{i_n}(\vy)
    \end{pmatrix}
\end{align*}

This function $\vec{v}_q$ is completely independent of $\vlambda$, and can be directly plugged into \gasp to obtain the integral over each monomial $m_i(\vy)$. This construction also makes it straightforward to parallelize using existing frameworks like PyTorch \citep{paszke2019pytorch}.

We therefore use the following algorithm for the symbolic integration using \gasp:

\begin{algorithm}
   \caption{$\mathsf{SymbolicIntegral}(q, \vH)$}
   \label{alg:gasp_symbolic_integral}
   \textbf{Input} Polynomial $q(\vy, \vlambda)$, Polytope $\vH$\\
   \textbf{Output} $I(\vlambda) = \int_{\vH} q(\vy, \vlambda) d\vY = \sum \lambda_i \eta_i$
   \begin{algorithmic}[1]
    \STATE $\vec{v}_q \gets \mathsf{ToVectorValued}(q)$
    \STATE $\eta \gets \mathsf{GASP!}(\vec{v}_q, \vH)$
    \COMMENT{see \cref{alg:gasp}}
    \RETURN $\vlambda \to \mathsf{I}(\vlambda ; \eta)$ \COMMENT{a function (computational graph) that maps $\vlambda$ to the integral $\mathsf{I}(\vlambda ; \eta)$ }
   \end{algorithmic}
\end{algorithm}

\subsection{Extended Complexity Analysis of \gasp}
\label{app:gasp_analysis_overall}

\subsubsection{Detailed Analysis of \gasp}
\label{app:gasp_detail}

We will now detail the algorithm used to solve the non-symbolic integration problem over the convex polytope $\vH$: $\int_{\vH} q(\vy) d\vY$, where $q$ denotes a polynomial that is only over $\vy$ (not $\vy$ and $\vlambda$).

The main entry-point for \gasp is \cref{alg:gasp}, which takes a polynomial $q$ and a convex polytope $\vH$ in $\calH$-description, so defined via $\vH = \{\vy | \vA \vy \leq \vb \}$, and returns the integral $\int_{\vH} q(\vy)d\vY$. Being based on a cubature integration-formula over the unit-simplex, the first step is first querying the total degree of $q$ and then creating the cubature points and weight (L$2$, \cref{alg:gasp}). The total number of points and weights depend on the degree, but are exact for any polynomial up to the respective degree. The cubature points and weights follow theorem $4$ of \citet{doi:10.1137/0715019}, also provided in \cref{alg:prepare_gm_cubature}. Given a polynomial of total degree $d$ and dimension $n$, enumerating the points has a complexity of $\mathcal{O}(r \cdot \binom{r + n - 1}{n - 1})$ for $r=\lceil \frac{d}{2} - 1\rceil$. In practice, this is done on the CPU and re-used for every polytope we want to integrate over. We then move our polytope $\vH=\{\vx \mid \vA\vx \leq \vb\}$ from $\mathcal{H}$ description into its $\mathcal{V}$-description and call it $\vV$ (L$4$, \cref{alg:gasp}). This operation has a complexity of $\mathcal{O}(m^{\lfloor n/2 \rfloor})$, with $m$ being the number of inequalities \citep{10.1007/BF02573985}, so polynomial for some fixed dimension $n$. Afterwards, we triangulate the vertices $\vV$ into an array of simplices $\vS$ (L$6$, \cref{alg:gasp}). While finding the minimal triangulation is NP-complete \citep{kaibel2002algorithmicproblemspolytopetheory}, finding some triangulation using the Delaunay-algorithm can be done in $\mathcal{O}(v^{\lceil n/2 \rceil})$ \citep{10.5555/1283383.1283502}, with $v$ being the number of vertices obtained previously. These computations are also executed on the CPU using QHull \citep{10.1145/235815.235821}. We are now prepared for the actual numerical integration, which will happen on the GPU (L$7$, alg. \ref{alg:gasp}).

This algorithm is detailed in \cref{alg:gasp_integrate}. It takes the polynomial $q$, points and weights $(\vR,\vw)$ from the cubature rule and simplices $\vS$. In practice, these are PyTorch \citep{DBLP:journals/corr/abs-1912-01703} tensors. We then loop over each simplex in $\vs_i$ (L$2$, \cref{alg:gasp_integrate}) and compute the cubature over all cubature points and weights $(\vR,\vw)$ (L$6$, \cref{alg:gasp_integrate}). As these cubature points are distributed over the unit-simplex, we need a coordinate change to transform the points from the unit-simplex to points on $\vs_i$ (L$8$, \cref{alg:gasp_integrate}), which is just a matrix-vector multiplication. We also need to calculate the absolute determinant of the Jacobian of this transformation for the pullback-measure, which coincides with the volume of the simplex (L$5$,\cref{alg:gasp_integrate}) and has the complexity $n^3$ due to the determinant. We then evaluate the polynomial on each point and add the sum weighted according to the cubature weight $\vw$ (L$10$ and $13$, \cref{alg:gasp_integrate}). In the end we sum up our integral over each simplex to arrive at the integral over our polytope (L$15$, \cref{alg:gasp_integrate}). In order to increase numerical accuracy, we first divide into positive and negative parts, sort each, and then sum up the elements via $\mathsf{stableSum}$. In practice, this is done per batch in the inner loop, as it runs batched. Every loop in this algorithm, including over multiple monomials arising due to the symbolic integral, is done in parallel and leverages the parallelism of the GPU. 

Finally, we want to stress that every computation in our algorithm uses established, heavily optimized numerical routines. This approach allows us to exploit the unique performance characteristics and heavy parallelism of GPUs to tackles this challenging problem. Therefore, the overall complexity of algorithm \ref{alg:gasp_integrate} is $\mathcal{O}(l_s \cdot (l_{gm} \cdot (n^2 +\log l_{gm}) + n^3) \log l_s)$, with $l_s$ being the number of simplices and $l_{gm}$ the number of cubature points.

This enables us to compute the overall complexity of gasp by noting that the number of simplices grows with $\mathcal{O}(m^{\lceil n/2 \rceil^2})$ \citep{SEIDEL1995115} with $m$ being the number of inequalities. We arrive at $\mathcal{O}(m^{\lceil n/2 \rceil^2} \cdot (l_{gm} \cdot (n^2 +\log l_{gm}) + n^3) \lceil n/2 \rceil^2 \log m)$ and $l_{gm}=r \cdot \binom{r + n - 1}{n - 1}$. While this complexity seems unwieldy, we first want to note that the problem is fundamentally hard, as integrating an arbitrary polynomial over a single simplex is already NP-Hard \citep{baldoni2011integrate}. Furthermore, in many applications, we can assume $dim(\vY) \ll dim(\vX)$ and therefore $dim(\vY)$ is actually reasonable. Finally, we can amortize this computation as it must only be done once per constraint, enabling fast and efficient training. %

\subsubsection{Grundmann-M\"{o}ller-Cubature}

\begin{algorithm}
   \caption{$\mathsf{PrepareGrundmannM\mathrm{\ddot{o}}ller}(d, n)$}
   \label{alg:prepare_gm_cubature}
   \textbf{Input} Total degree $d$, dimensions $n$\\
   \textbf{Output} Cubature points $\vR \in \mathbb{R}^{l_{gm} \times n}$ and weights $\vw \in \mathbb{R}^{l_{gm}}$
   \begin{algorithmic}[1]
   \STATE $\vR \gets []$ \;
   \STATE $\vw \gets []$ \;
   \STATE $s \gets \lceil \frac{d}{2} - 1\rceil$ \;
   \STATE \COMMENT{According to Theorem 4 of \citet{doi:10.1137/0715019}}
   \FOR{$i \gets 0$ \textbf{to} $s$}
       \STATE $w_i \gets (-1)^i 2^{-2s} \frac{(d+n-2i)^d}{i!(d+n-i)!}$ \;
       \STATE $\vGamma \gets \mathsf{combinationsSummingTo}(n, s - i)$ \;
       \STATE \COMMENT{All combinations of $n$ natural numbers summing to $s - i$}
       \FOR{$\vgamma \in \vGamma$}
           \STATE $\vr \gets \left(\frac{2\gamma_0+1}{d+n-2i}, \dots, \frac{2\gamma_n+1}{d+n-2i} \right)$ \;
           \STATE $\mathsf{Append}(\vR, \vr)$
           \STATE $\mathsf{Append}(\vw, w_i/\mathsf{Len}(\Gamma))$ \;
       \ENDFOR
   \ENDFOR
   \RETURN{$\vR, \vw$}
   \end{algorithmic}
\end{algorithm}

\subsubsection{Overall Complexity}
\label{app:complexity_pal_overall}

When analyzing the overall complexity of \ours, there are two major additional sources of computational complexity: the complexity of the SMT formula and, assuming the use of splines, the number of bins.

We will first focus on the complexity originating from the logical formula. The WMI problem in general is \#P-hard \citep{zeng2020probabilistic}. 
Coarsely speaking, the SAE4WMI procedure may require exponential time in the worst case. It begins with an AllSMT step that enumerates all partial assignments satisfying the formula---potentially exponentially many. Each of these assignments is then checked for $\lra$ consistency, which takes polynomial time per assignment. 
Afterwards, we have the integration of the polynomial on each polytope described by the $\lra$-consistent truth assignment — a task that is NP-hard in itself, as discussed in the previous section. Thus, the logical component alone can dominate the overall complexity, especially when the number of satisfying assignments grows rapidly. That said, this compilation cost is amortized: we pay it only once before training.

Another source of complexity arises from the number of bins when using a spline-based density, as our pipeline must be executed once per bin. While this might seem at first like a significant disadvantage for using a spline, in practice the complexity of the logical formula per bin, and therefore the overall time required for the compilation, decreases with an overall increasing number of bins. This is due to the space being partitioned into smaller and smaller patches as we increase the number of bins, with fewer constraints intersecting the average bin. For example, the time it takes to integrate a $169$-bin spline over scene $1$ of the Stanford drone dataset is just $3.77$x the time it takes to integrate a spline with a single bin over the scene, instead of $169$x.
We show the relationship between linear scaling and our actual measured integration times in figure \ref{fig:analysis_gas_time_vs_bin}.

\begin{figure}
    \centering
    \includegraphics[width=0.5\linewidth]{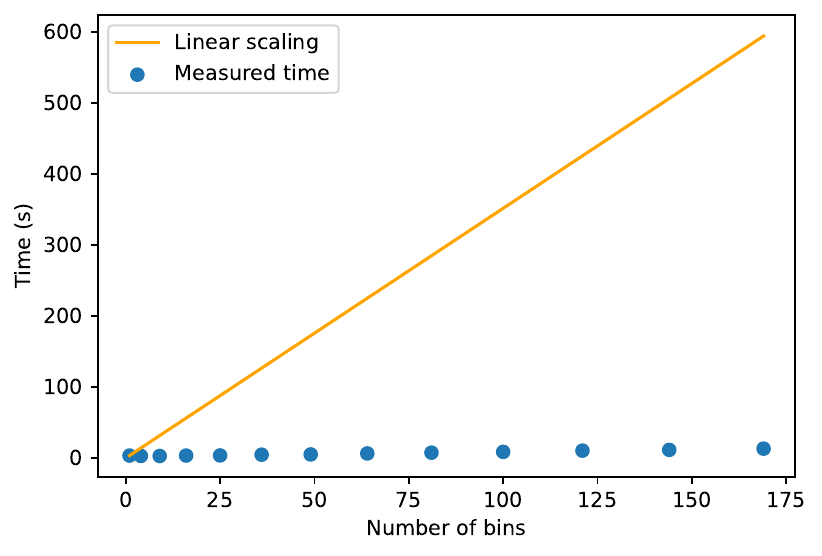}
    \caption{The integration time for splines on the Stanford drone dataset (scene $1$) does not grow linearly with the number of bins. In this figure, we compare the time it takes to integrate our squared spline for an increasing number of bins versus a simple linear scaling, which we would expect if the empirical computational complexity per bin would stay approximately constant.}
    \label{fig:analysis_gas_time_vs_bin}
\end{figure}

\FloatBarrier

\section{Additional Related Works}
\label{app:related-works}

\textbf{Other constraints in deep learning.} Many approaches have been proposed for dealing with a variety of specialized constraints in neural networks. 
An abundant body of work investigates architectures that guarantee specific properties, such as monotonicity~\citep{deeplattice} or permutation invariance of inputs~\citep{deepsets}.
Other works impose constraints over they dynamics of a neural network, e.g., to follow a physics-based constraint or a PDE \citep{raissi2019physics,li2020fourier,beltran2024galerkin}.
These approaches are orthogonal to ours and cannot be easily generalized in our framework where constraints are algebraic.

\textbf{Sampling with constraints.} It is well known that (algebraic) constraints pose significant challenges for sampling procedures.\citet{algebraicgibbs} addressed these challenges with Markov Chain Monte Carlo, which is however unsuited for training \ours via gradient descent.
\citet{abboud2022approximate} introduce an FPRAS scheme to approximate WMI problems whose constraints are represented in disjunctive normal form (DNF). While this approach provides some guarantees, it would still require many samples to get an accurate estimate of a \ref{eq:wmi} integral and DNFs are not compact representations of many real-world constraints \citep{hoernle2022multiplexnet}.
Another recent work on sampling under algebraic constraints is the Disjunctive Refinement Layer \citep{DBLP:journals/corr/abs-2502-18237}, which is an iterative projection onto non-convex sets defined by quantifier-free conjunctions, disjunctions and negations of linear inequalities. It allows for sampling with guaranteed constraint satisfaction, but the iterative projection of the invalid probability mass leads to a clustering of projected samples at the boundaries and obstructs gradient flow.

\textbf{Learning constraints.} The problem of learning \smtlra constraints from positive/negative examples was first addressed by INCAL~\citep{incal}, an incremental approach built upon SMT solvers.
LARIAT~\citep{lariat} addressed the problem of jointly learning the \smtlra constraints and a piecewise polynomial density from unlabelled data. These two components are learned separately and then combined into a probabilistic model that can be queried using WMI solvers. 
In this paper, we assume the constraints are given and they are accounted for when learning the parameters of the density function. 
Combining \ours with the above approaches is an interesting future direction.

\section{Experimental details}

\subsection{Rejection sampling vs. \gasp}
\label{app:rejection_vs_gasp}

The polynomials with dimension $n$ and total degree $d$ we want to integrate are all of the form 
$$\left(\sum_{\substack{\valpha_i \in \mathbb{N}^n \\ 1^T\valpha_i \leq d}} c_i \prod_j y_j^{\alpha_{ij}}\right)^2 .$$
The coefficients $c_i$ are distributed as follows: $c_i \sum \pm\text{Poisson}(2)$, where $\pm$ denotes a random, equal chance, sign.

We generate the random simplices by following the same procedure as many times as needed:

\begin{enumerate}
    \item draw a random unit simplex;
    \item scale it between $0.5$ and $1.5$ (uniformly);
    \item transform the vertices using a random orthonormal matrix;
    \item translate it between $0$ and $6$;
    \item keep the simplex if it does not overlap any previous simplex.
\end{enumerate}

\begin{figure}[h!]
    \centering
    \begin{tabular}{l l l l}
        & \multicolumn{1}{c}{degree: $4$} & \multicolumn{1}{c}{degree $8$} & \multicolumn{1}{c}{degree $12$} \\
        \rotatebox{90}{\parbox{4cm}{\centering num. simplices: 1}}
        & \includegraphics[height=4cm]{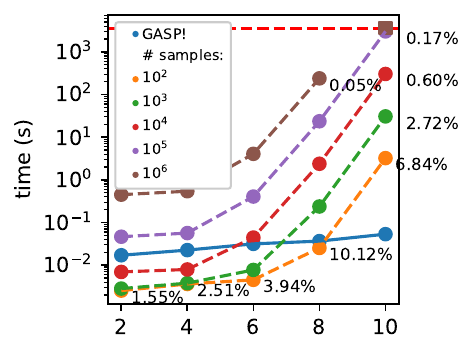}
        & \includegraphics[height=4cm]{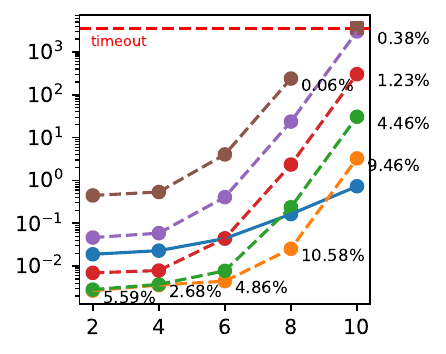}
        & \includegraphics[height=4cm]{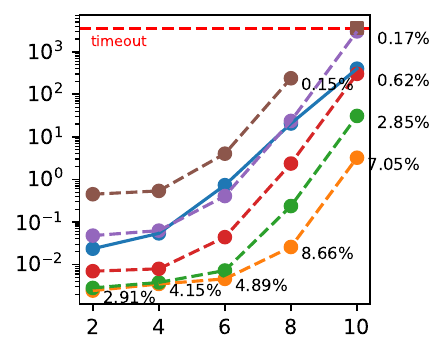} \\
        \rotatebox{90}{\parbox{4cm}{\centering num. simplices: 2}}
        & \includegraphics[height=4cm]{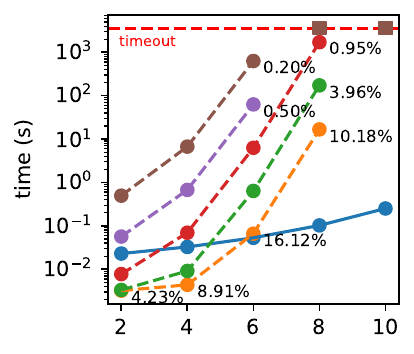}
        & \includegraphics[height=4cm]{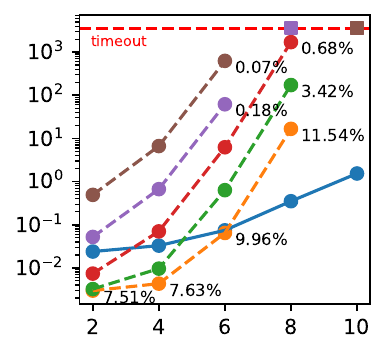}
        & \includegraphics[height=4cm]{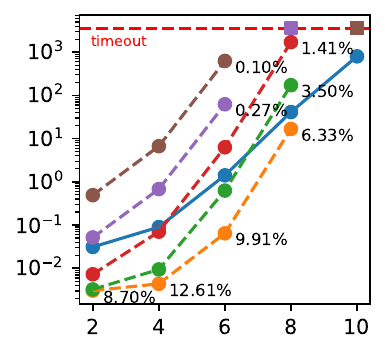} \\
        \rotatebox{90}{\parbox{4cm}{\centering num. simplices: 4}}
        & \includegraphics[height=4.3cm]{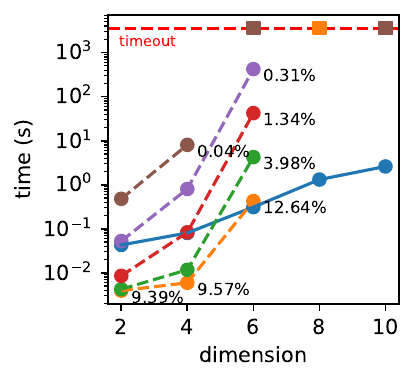}
        & \includegraphics[height=4.3cm]{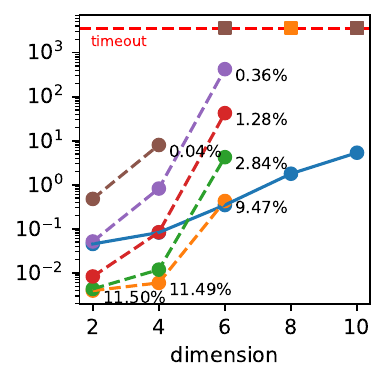}
        & \includegraphics[height=4.3cm]{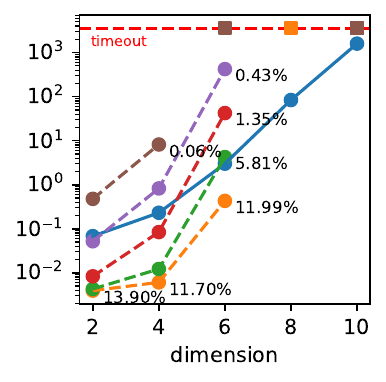} \\
    \end{tabular}
    \caption{Runtime (in seconds) of rejection sampling vs a single \gasp-run for integrating random polynomials of varying degree over $4$ random simplices. For rejection sampling, we additionally report its relative error. This benchmark was run using an NVIDIA RTX A6000, an AMD EPYC 7452 32-Core Processor and 528 Gigabyte RAM.}
    \label{fig:numerical_integration_all}
\end{figure}

\FloatBarrier

\subsection{SAE4WMI(\latte) vs. SAE4WMI(\gasp)}
\label{app:latte_vs_gasp}

For this experiment, we used the benchmarking suite originally
released with SAE4WMI~\citep{wmisapa}
(\url{https://github.com/unitn-sml/wmi-benchmarks}) for generating
random WMI problems.
These instances are composed of a \smtlra formula encoding the support
of the distribution and a piecewise polynomial weight function. The
weight functions have arbitrary \smtlra conditions as internal nodes
and non-negative polynomial leaves. In contrast with the benchmarks
employed by Spallitta et al., our weight functions do not have sums or
product as internal nodes. This minor modification was made to enforce
a tighter control over the maximum overall degree of the weight
function.
The problems have $n \in \{3, 4, 5\}$ real variables and are generated
in a recursive manner, with formulas and weight functions having depth
$r \in \{2, 3, 4\}$, the latter having polynomial leaves with maximum
degree $d \in \{0, 2, 4\}$.
For each configuration of $\langle n,d,r \rangle$, we generated $10$
instances, for a total of $270$ WMI problems.
This benchmark was run using an NVIDIA RTX A6000, an AMD EPYC 7452 32-Core Processor and 528 Gigabyte RAM.

\subsubsection{\gasp vs SymPy}
\label{sec:appendix_gasp_vs_xadd}

Another option is to compare the whole \gasp pipeline, including PA \citep{wmipa} to enumerate the convex Polytopes, to a completely orthogonal approach. XADD \citep{DBLP:conf/ijcai/KolbMSBK18} tackles the weighted model integral via symbolic manipulations. In practice, this means integrating via an explicit anti-derivative and then replacing the variable with the symbolic lower and upper bounds, analogously on how solving an integral by hand is done. As the symbolic polynomial is represented using SymPy \citep{10.7717/peerj-cs.103}, our polynomial over both variables and coefficients can be naturally expressed and running XADD \citep{DBLP:conf/ijcai/KolbMSBK18} on this polynomial directly results in $I_\phi(\vLambda)$. 
We benchmark \gasp vs. XADD equipped with SymPy on the NStar-constraints, which is just an n-pointed star where we always connect with opposite points. An example of the constraints can be seen in figure \ref{tab:7_star_1_cauchy_density}. Using our algorithm \gasp, we were able to reduce the runtime for the integral a polynomial of total degree $12$ for a star with $17$ corners from $6$ hours and $20$ minutes to $19$ seconds, a significant speedup of $3$ magnitudes or approximately $1200$ times faster. Detailed results for the benchmark are in the Appendix in table \ref{tab:integral_symbolic_benchmark_xadd} and \ref{tab:integral_symbolic_benchmark_gasp}.

\begin{table}[h!]
\centering
\input{tables/benchmark_sympy}
\caption{We show the results for integrating the NSTar-Benchmark using the XADD-Algorithm equipped with SymPy \citep{DBLP:conf/ijcai/KolbMSBK18}. Results in $hh:mm:ss.$. This benchmark was run on the same machine as in \ref{tab:integral_symbolic_benchmark_gasp}. This benchmark was run using an NVIDIA RTX A6000, an AMD EPYC 7452 32-Core Processor and 528 Gigabyte RAM. The results for \gasp are provided in table \ref{tab:integral_symbolic_benchmark_gasp}.}
\label{tab:integral_symbolic_benchmark_xadd}
\end{table}

\begin{table}[h!]
\centering
\input{tables/benchmark_pa}
\caption{\gasp is significantly faster compared to XADD \citep{DBLP:conf/ijcai/KolbMSBK18}. We show results for integrating the NSTar-Benchmark using \gasp. Results in $hh:mm:ss.$. This benchmark was run using an NVIDIA RTX A6000, an AMD EPYC 7452 32-Core Processor and 528 Gigabyte RAM. The results for XADD are provided in table \ref{tab:integral_symbolic_benchmark_xadd}.}
\label{tab:integral_symbolic_benchmark_gasp}
\end{table}

\FloatBarrier

\subsection{NStar}
\label{sec:appendix_experiments_nstar}

For the NStar dataset, we generate $800.000$ $(\vx, \vy)$-points on the NStar via rejection sampling. The NStar-constraints are formed by taking the n-pointed star on the circle with radius $10$, where the the constraints are constructed by connecting each corner-point with the two most-opposite points. The distribution of $\vX$ is uniform over the star, the distribution of $\vY$ is a cauchy-distribution with location $\vX$ and scale $1.5\sqrt{10}$. The star is located in $[-10,10]^2$.

We form train, test and validation datasets by dividing the points with a share of $70\%$, $15\%$ and $15\%$.

\FloatBarrier

\subsubsection{Models}
\label{sec:appendix_experiments_nstar_details}

All models are trained with a batch-size of $512$.

\textbf{\ours} We will now describe our settings for the \ours-models on the NStar-Dataset.
For every variant of the star, we perform a grid-search over the following configurations, picking the best-performing according to the log-likelihood on the held-out dataset:
\begin{itemize}
    \item epochs: $1500$
    \item learning rate:$1e-06$, $1e-05$
    \item network hidden layers: $[1024, 1024]$, $[1024]$
\end{itemize}

The network is a fully-connected neural network using ReLU \citep{DBLP:journals/jmlr/GlorotBB11} as activations. We use the schedule-free version of Adam \citep{Defazio2024TheRL}. We use a single polynomials, with the re-parametrization as detailed in \cref{app:training_polynomials} and train using a loss composed of log-likelihood and a penalty on very large integral-values detailed in \cref{app:loss}.

In this experiment, our density is modeled by a single, squared polynomial over $\vY$ with the maximum number of terms given the required total degree. So, for example, the polynomial with a total degree of $10$ squared has a total degree of $5$ unsquared. Therefore we collect all multivariate monomials up to the total degree of $5$ to build our parametrized polynomial over $\vY$.

\textbf{GMM} For the GMM-Models, we perform the following grid-search:
\begin{itemize}
    \item covariance: full and independent (although full covariance leads to better performing models for our dataset)
    \item epochs: $1500$
    \item learning rate: $1e-04$, $1e-05$
    \item network hidden layers: $[1024, 1024]$, $[1024]$
\end{itemize}
We use the Adam optimizer \citep{DBLP:journals/corr/KingmaB14}.

\textbf{DSP} For DeepSeaProbLog, when performing the grid-search, we select the model according to the loss with the constant, final multiplier for the continous approximation for the inequality. The grid is over the following parameters:
\begin{itemize}
    \item epoch: $1500$
    \item learning rate: $0.001$, $0.0001$, $0.00001$
    \item minimum-multiplier for the inequality-relaxation: $0.1$, $1.0$
    \item maximum-multiplier: $5$
    \item network hidden layers: $[1024, 1024]$, $[1024]$
    \item optimiser: AdaMax and Adam \citep{DBLP:journals/corr/KingmaB14}
\end{itemize}
Due to the slower training speed, we train DSP with a patience of $200$-epochs (the other models are picked as the model with the best validation-score over all $1500$ epochs). Finally, we train with sampling $50$ times per input $\vx^{(b)}$ in order to approximate $P(valid)$.

DSP is trained by optimizing both the fit on the data, as well as constraint satisfaction:

\begin{align}
    \tag{DSP-Loss}
    \label{eq:dsp_loss}
    l_{dsp}(\vx^{[1:b]}, \vy^{[1:b]}) &= \sum_i (-1)\cdot\log p(\vy^{i} | \vx^{i}) + \mathsf{CE}(p(\vY \models \phi \mid \vX = \vx^i))
\end{align}
where $p(\vY \models \phi \mid \vX = \vx^i)$ is approximated numerically in DSP in comparison to \ours. $\mathsf{CE}$ denotes the binary cross-entropy loss against the constant $1$ label.

\subsubsection{Results}

\begin{table}[h!]
\centering
\begin{adjustbox}{center}
\resizebox{1.0\linewidth}{!}{%
\input{tables/nstar_overview_all_models_replicated}
}
\end{adjustbox}
\caption{Average log-likelihood on the test-set for the NStar-dataset. We test on a $3$, $7$, $11$ and $19$-Star and compare our approach (NN+\ours) to a conditional GMM and the neural distributional fact fitted by DeepSeaProblog. For DeepSeaProblog, we report the performance of two model, one selected by best log-likelihood (LL) and one by best loss (Loss), which takes into consideration both the fit and the probability of violating the constraints. After choosing the hyper-parameters, all runs were repeated $10$-times and we report mean and standard deviation.}
\label{tab:nstar_big-table}
\end{table}

\begin{table}[h!]
    \centering
    \begin{adjustbox}{center}
\resizebox{1.0\linewidth}{!}{%
    \begin{tabular}{p{2cm}p{2cm}p{2cm}p{2cm}p{2cm}p{2cm}p{2cm}p{2cm}p{2cm}p{2cm}}
\toprule    
Ground Truth & \multicolumn{3}{c}{NN + \ours } & \multicolumn{4}{c}{NN + GMM} & \multicolumn{2}{c}{DeepSeaProblog (by LL)}\\
\cmidrule(lr){2-4} \cmidrule(lr){5-8} \cmidrule(lr){9-10}
& deg $5^2$ & deg $7^2$ & deg $9^2$ & $K{=}1$ & $K{=}4$ & $K{=}8$ & $K{=}32$ & density & query \\
    
\includegraphics[width=2cm]{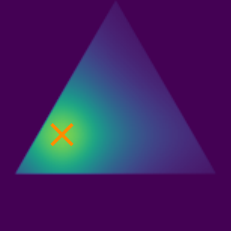}
& \includegraphics[width=2cm]{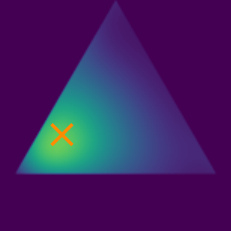}
& \includegraphics[width=2cm]{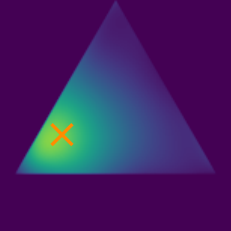}
& \includegraphics[width=2cm]{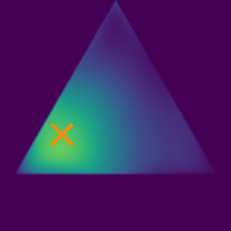}
& \includegraphics[width=2cm]{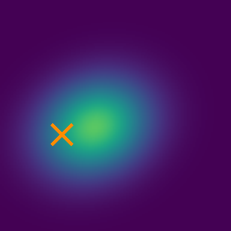}
& \includegraphics[width=2cm]{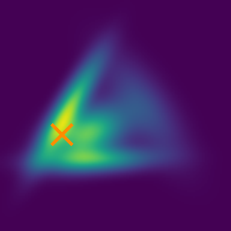}
& \includegraphics[width=2cm]{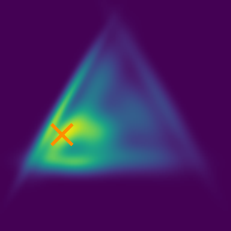}
& \includegraphics[width=2cm]{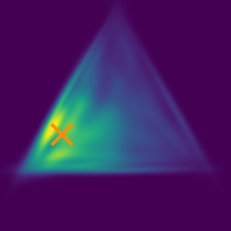}
& \includegraphics[width=2cm]{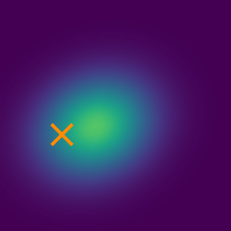}
& \includegraphics[width=2cm]{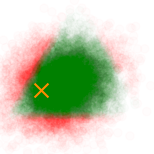}\\
\includegraphics[width=2cm]{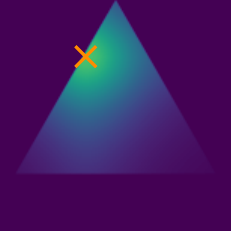}
& \includegraphics[width=2cm]{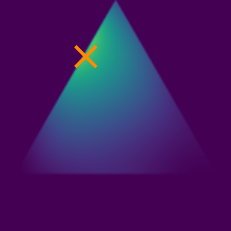}
& \includegraphics[width=2cm]{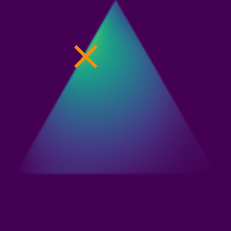}
& \includegraphics[width=2cm]{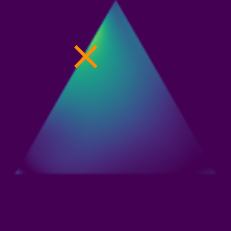}
& \includegraphics[width=2cm]{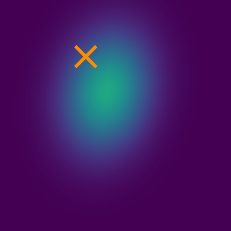}
& \includegraphics[width=2cm]{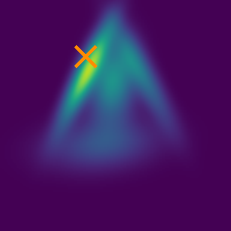}
& \includegraphics[width=2cm]{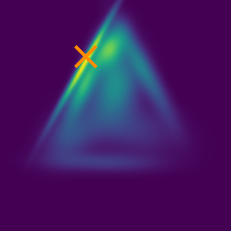}
& \includegraphics[width=2cm]{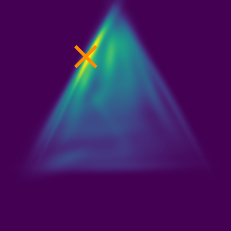}
& \includegraphics[width=2cm]{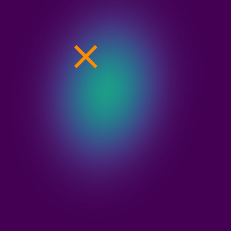}
& \includegraphics[width=2cm]{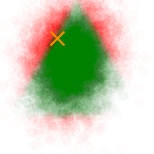}\\
\includegraphics[width=2cm]{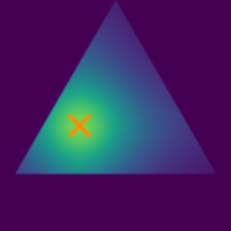}
& \includegraphics[width=2cm]{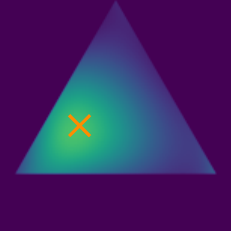}
& \includegraphics[width=2cm]{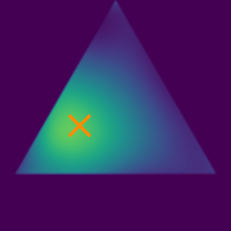}
& \includegraphics[width=2cm]{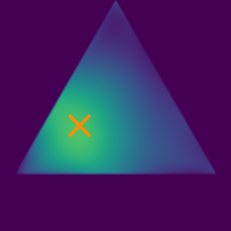}
& \includegraphics[width=2cm]{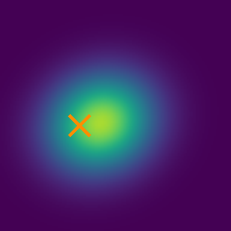}
& \includegraphics[width=2cm]{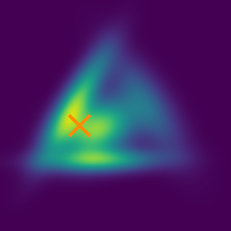}
& \includegraphics[width=2cm]{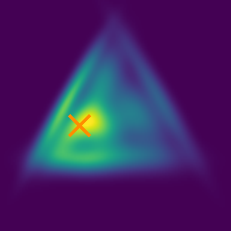}
& \includegraphics[width=2cm]{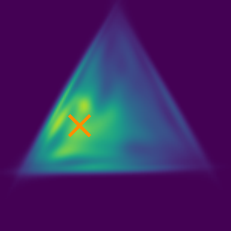}
& \includegraphics[width=2cm]{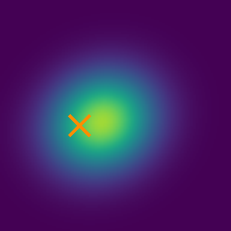}
& \includegraphics[width=2cm]{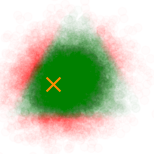}\\
\bottomrule
\end{tabular}
}
\end{adjustbox}
    \caption{Densities of the Ground-Truth compared to the polynomial, GMM and DeepSeaProbLog for the $3$-Star problem with a cauchy-density. For the DSP model selected by log-likelihood, we show the density of the neural distributional fact, and we also show the result of querying the ProbLog program representing our constraints $10000$ times. The samples associated with an true-label are shown in green, the samples associated with a false-label are shown in red.}
    \label{tab:3_star_1_cauchy_density}
\end{table}

\begin{table}[h!]
    \centering
    \begin{tabular}{p{2cm}p{2cm}p{2cm}}
\toprule    
Ground Truth & \multicolumn{2}{c}{DeepSeaProblog (by loss)}\\
\cmidrule(lr){2-3}
& density & query \\
\includegraphics[width=2cm]{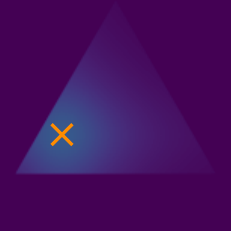}
& \includegraphics[width=2cm]{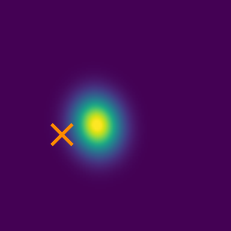}
& \includegraphics[width=2cm]{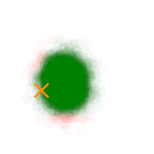}\\
\includegraphics[width=2cm]{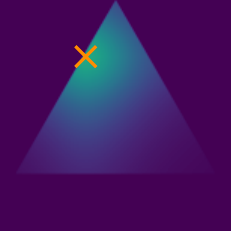}
& \includegraphics[width=2cm]{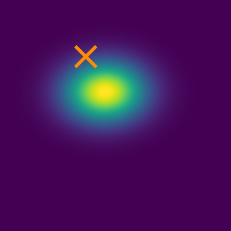}
& \includegraphics[width=2cm]{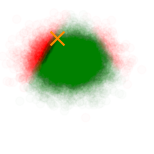}\\
\includegraphics[width=2cm]{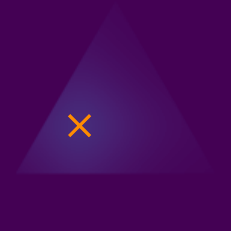}
& \includegraphics[width=2cm]{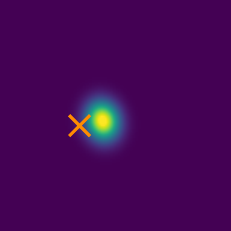}
& \includegraphics[width=2cm]{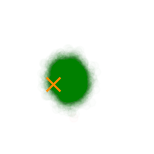}\\
\bottomrule
    \end{tabular}
    \caption{Densities of the Ground-Truth compared to the DeepSeaProbLog for the $3$Star problem with a cauchy-density. DeepSeaProbLog is selected by loss, and due to avoiding the constraints, more concentrated and therefore visualized separately. We show the density for the neural distributional fact and the samples obtained by the query. The samples associated with an true-label are shown in green, the samples associated with a false-label are shown in red. We choose to visualize this separately in order to keep the color-scheme in figure \ref{tab:3_star_1_cauchy_density} reasonable.}
    \label{tab:3_star_1_cauchy_density_deepsea_loss}
\end{table}

\begin{table}[h!]
    \centering
    \begin{adjustbox}{center}
\resizebox{1.0\linewidth}{!}{%
    \begin{tabular}{p{2cm}p{2cm}p{2cm}p{2cm}p{2cm}p{2cm}p{2cm}p{2cm}p{2cm}p{2cm}}
\toprule
Ground Truth & \multicolumn{3}{c}{NN + \ours } & \multicolumn{4}{c}{NN + GMM} & \multicolumn{2}{c}{DeepSeaProblog}\\
\cmidrule(lr){2-4} \cmidrule(lr){5-8} \cmidrule(lr){9-10}
& deg $5^2$ & deg $7^2$ & deg $9^2$ & $K{=}1$ & $K{=}4$ & $K{=}8$ & $K{=}32$ & density & query \\
\includegraphics[width=2cm]{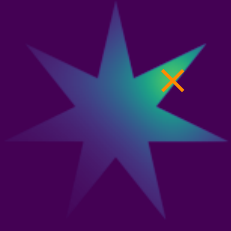}
& \includegraphics[width=2cm]{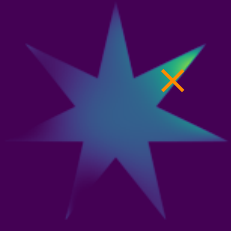}
& \includegraphics[width=2cm]{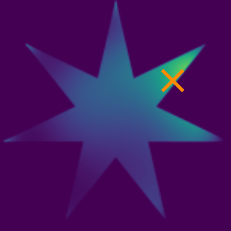}
& \includegraphics[width=2cm]{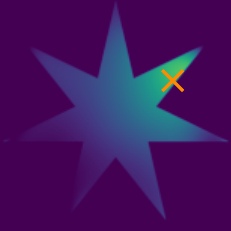}
& \includegraphics[width=2cm]{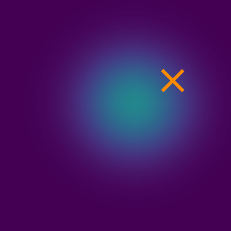}
& \includegraphics[width=2cm]{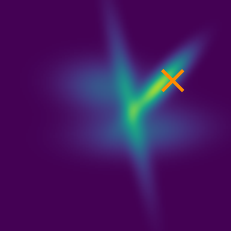}
& \includegraphics[width=2cm]{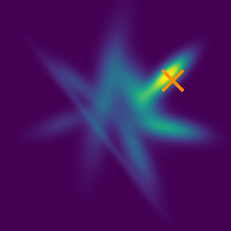}
& \includegraphics[width=2cm]{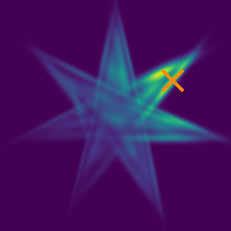}
& \includegraphics[width=2cm]{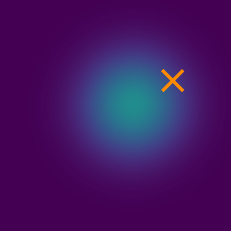}
& \includegraphics[width=2cm]{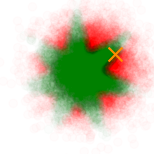}\\
\includegraphics[width=2cm]{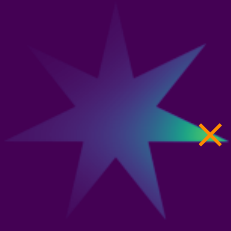}
& \includegraphics[width=2cm]{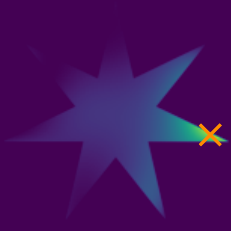}
& \includegraphics[width=2cm]{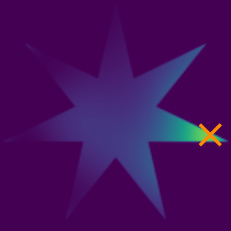}
& \includegraphics[width=2cm]{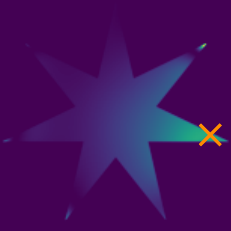}
& \includegraphics[width=2cm]{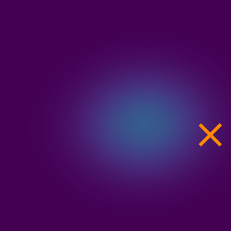}
& \includegraphics[width=2cm]{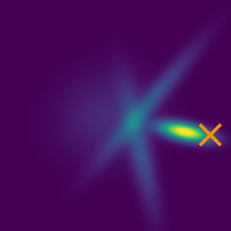}
& \includegraphics[width=2cm]{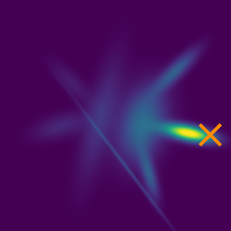}
& \includegraphics[width=2cm]{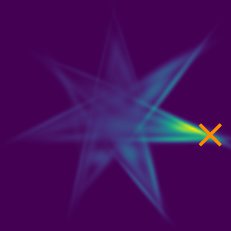}
& \includegraphics[width=2cm]{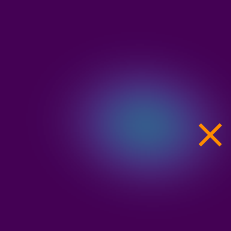}
& \includegraphics[width=2cm]{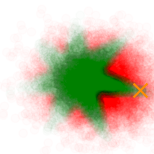}\\
\includegraphics[width=2cm]{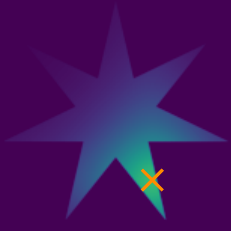}
& \includegraphics[width=2cm]{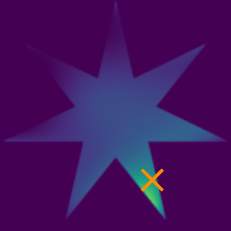}
& \includegraphics[width=2cm]{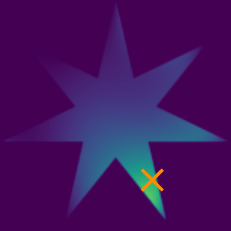}
& \includegraphics[width=2cm]{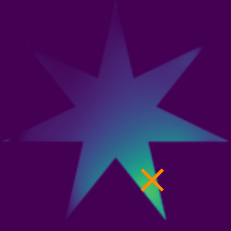}
& \includegraphics[width=2cm]{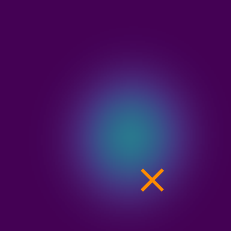}
& \includegraphics[width=2cm]{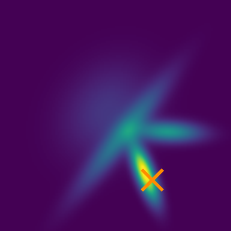}
& \includegraphics[width=2cm]{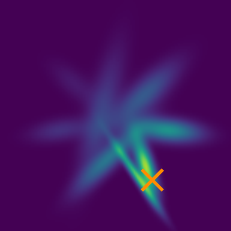}
& \includegraphics[width=2cm]{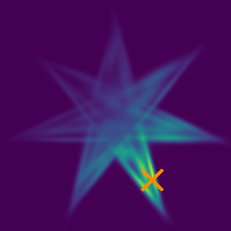}
& \includegraphics[width=2cm]{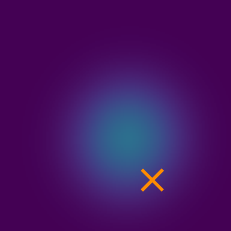}
& \includegraphics[width=2cm]{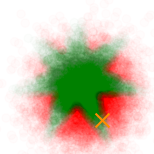}\\
\bottomrule
\end{tabular}
}
\end{adjustbox}
    \caption{Densities of the Ground-Truth compared to the polynomial, GMM and DeepSeaProbLog for the $7$-Star problem with a cauchy-density. For the DSP model selected by log-likelihood, we show the density of the neural distributional fact, and we also show the result of querying the ProbLog program representing our constraints $10000$ times. The samples associated with an true-label are shown in green, the samples associated with a false-label are shown in red.}
    \label{tab:7_star_1_cauchy_density}
\end{table}

\begin{table}[h!]
    \centering
    \begin{tabular}{p{2cm}p{2cm}p{2cm}}
\toprule
Ground Truth & \multicolumn{2}{c}{DeepSeaProblog (by loss)}\\
\cmidrule(lr){2-3}
& density & query \\
\includegraphics[width=2cm]{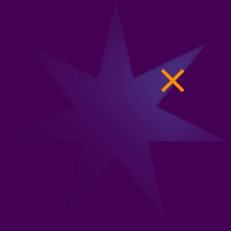}
& \includegraphics[width=2cm]{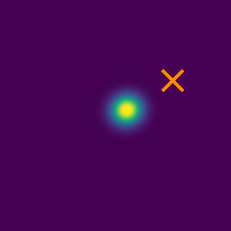}
& \includegraphics[width=2cm]{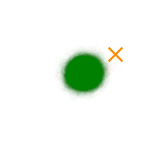}\\
\includegraphics[width=2cm]{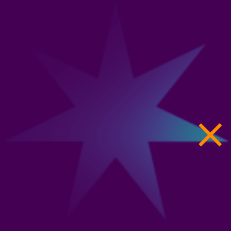}
& \includegraphics[width=2cm]{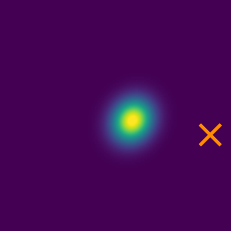}
& \includegraphics[width=2cm]{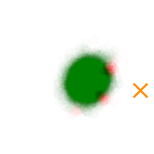}\\
\includegraphics[width=2cm]{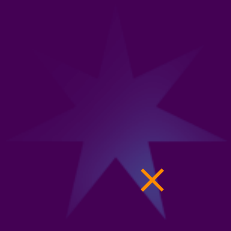}
& \includegraphics[width=2cm]{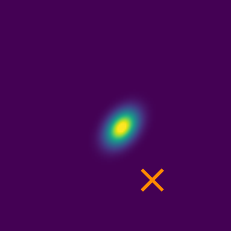}
& \includegraphics[width=2cm]{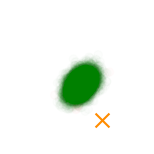}\\
\bottomrule
    \end{tabular}
    \caption{Densities of the Ground-Truth compared to the DeepSeaProbLog for the $7$Star problem with a cauchy-density. DeepSeaProbLog is selected by loss, and due to avoiding the constraints, more concentrated and therefore visualized separately. We show the density for the neural distributional fact and the samples obtained by the query. The samples associated with an true-label are shown in green, the samples associated with a false-label are shown in red. We choose to visualize this separately in order to keep the color-scheme in figure \ref{tab:7_star_1_cauchy_density} reasonable.}
    \label{tab:7_star_1_cauchy_density_deepsea_loss}
\end{table}

\begin{table}[h!]
    \centering
    \begin{adjustbox}{center}
\resizebox{1.0\linewidth}{!}{%
    \begin{tabular}{p{2cm}p{2cm}p{2cm}p{2cm}p{2cm}p{2cm}p{2cm}p{2cm}p{2cm}p{2cm}}
\toprule
Ground Truth & \multicolumn{3}{c}{NN + \ours } & \multicolumn{4}{c}{NN + GMM} & \multicolumn{2}{c}{DeepSeaProblog}\\
\cmidrule(lr){2-4} \cmidrule(lr){5-8} \cmidrule(lr){9-10}
& deg $5^2$ & deg $7^2$ & deg $9^2$ & $K{=}1$ & $K{=}4$ & $K{=}8$ & $K{=}32$ & density & query \\
\includegraphics[width=2cm]{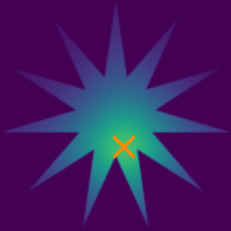}
& \includegraphics[width=2cm]{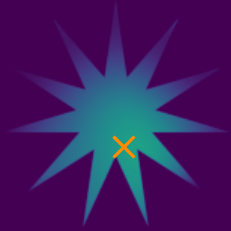}
& \includegraphics[width=2cm]{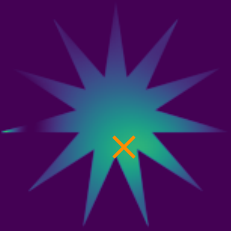}
& \includegraphics[width=2cm]{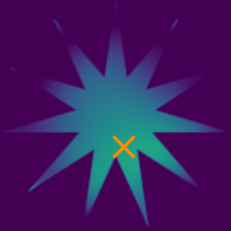}
& \includegraphics[width=2cm]{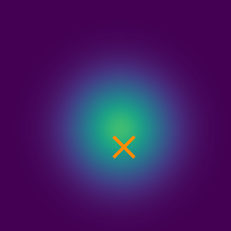}
& \includegraphics[width=2cm]{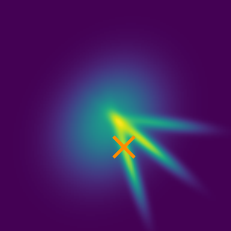}
& \includegraphics[width=2cm]{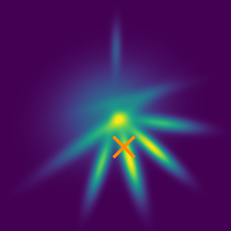}
& \includegraphics[width=2cm]{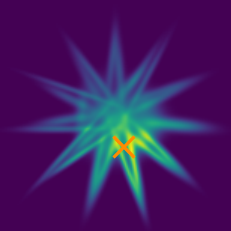}
& \includegraphics[width=2cm]{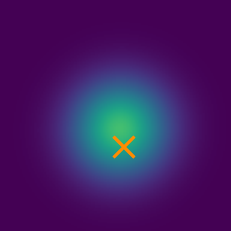}
& \includegraphics[width=2cm]{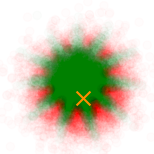}\\
\includegraphics[width=2cm]{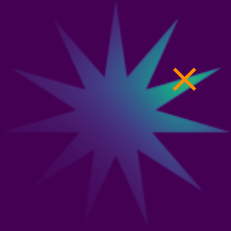}
& \includegraphics[width=2cm]{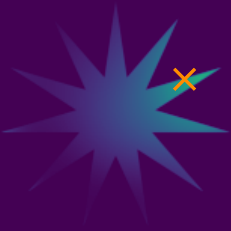}
& \includegraphics[width=2cm]{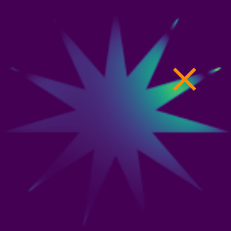}
& \includegraphics[width=2cm]{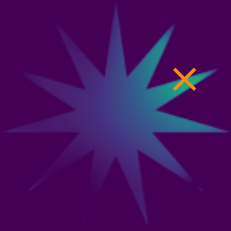}
& \includegraphics[width=2cm]{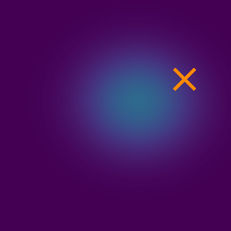}
& \includegraphics[width=2cm]{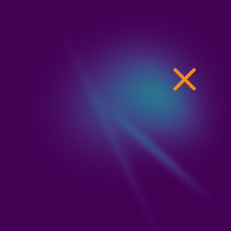}
& \includegraphics[width=2cm]{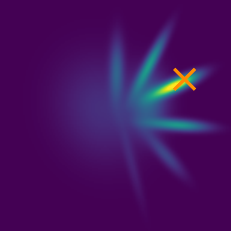}
& \includegraphics[width=2cm]{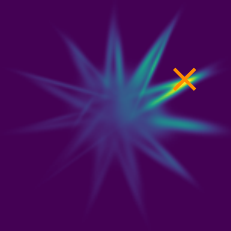}
& \includegraphics[width=2cm]{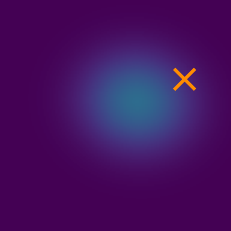}
& \includegraphics[width=2cm]{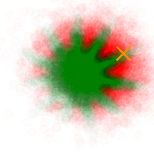}\\
\includegraphics[width=2cm]{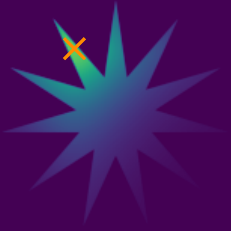}
& \includegraphics[width=2cm]{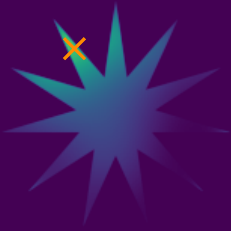}
& \includegraphics[width=2cm]{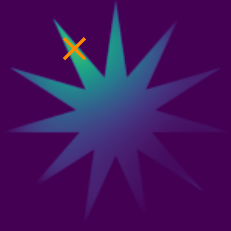}
& \includegraphics[width=2cm]{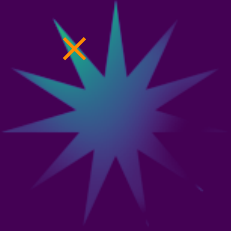}
& \includegraphics[width=2cm]{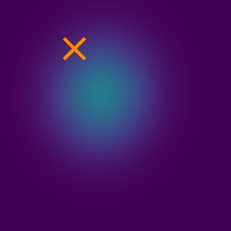}
& \includegraphics[width=2cm]{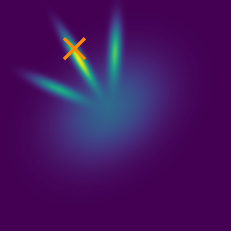}
& \includegraphics[width=2cm]{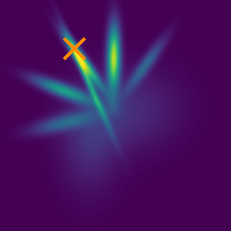}
& \includegraphics[width=2cm]{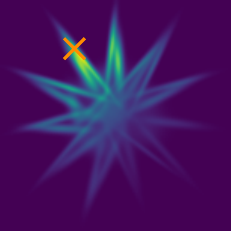}
& \includegraphics[width=2cm]{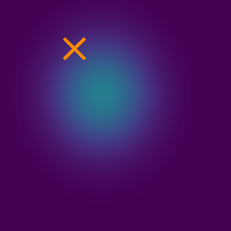}
& \includegraphics[width=2cm]{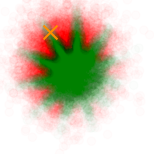}\\
\bottomrule
\end{tabular}
}
\end{adjustbox}
    \caption{Densities of the Ground-Truth compared to the polynomial, GMM and DeepSeaProbLog for the $11$-Star problem with a cauchy-density. For the DSP model selected by log-likelihood, we show the density of the neural distributional fact, and we also show the result of querying the ProbLog program representing our constraints $10000$ times. The samples associated with an true-label are shown in green, the samples associated with a false-label are shown in red.}
    \label{tab:11_star_1_cauchy_density}
\end{table}

\begin{table}[h!]
    \centering
    \begin{tabular}{p{2cm}p{2cm}p{2cm}}
\toprule
Ground Truth & \multicolumn{2}{c}{DeepSeaProblog (by loss)}\\
\cmidrule(lr){2-3}
& density & query \\
\includegraphics[width=2cm]{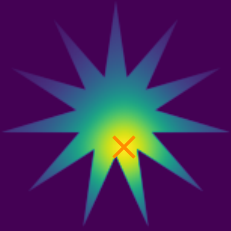}
& \includegraphics[width=2cm]{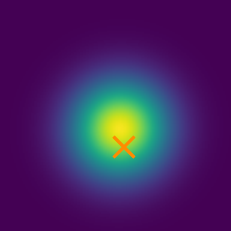}
& \includegraphics[width=2cm]{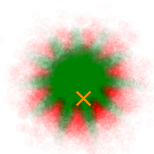}\\
\includegraphics[width=2cm]{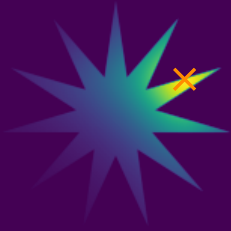}
& \includegraphics[width=2cm]{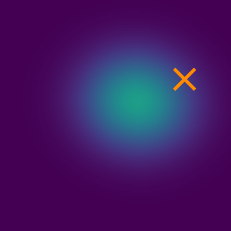}
& \includegraphics[width=2cm]{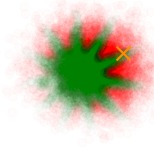}\\
\includegraphics[width=2cm]{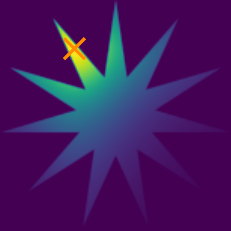}
& \includegraphics[width=2cm]{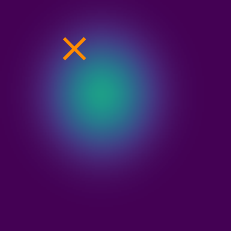}
& \includegraphics[width=2cm]{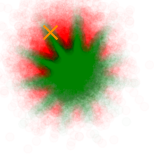}\\
\bottomrule
    \end{tabular}
    \caption{Densities of the Ground-Truth compared to the DeepSeaProbLog for the $11$Star problem with a cauchy-density. DeepSeaProbLog is selected by loss, and due to avoiding the constraints, more concentrated and therefore visualized separately. We show the density for the neural distributional fact and the samples obtained by the query. The samples associated with an true-label are shown in green, the samples associated with a false-label are shown in red. We choose to visualize this separately in order to keep the color-scheme in figure \ref{tab:11_star_1_cauchy_density} reasonable.}
    \label{tab:11_star_1_cauchy_density_deepsea_loss}
\end{table}

\begin{table}[h!]
    \centering
    \begin{adjustbox}{center}
\resizebox{1.0\linewidth}{!}{%
    \begin{tabular}{p{2cm}p{2cm}p{2cm}p{2cm}p{2cm}p{2cm}p{2cm}p{2cm}p{2cm}p{2cm}}
\toprule
Ground Truth & \multicolumn{3}{c}{NN + \ours } & \multicolumn{4}{c}{NN + GMM} & \multicolumn{2}{c}{DeepSeaProblog}\\
\cmidrule(lr){2-4} \cmidrule(lr){5-8} \cmidrule(lr){9-10}
& deg $5^2$ & deg $7^2$ & deg $9^2$ & $K{=}1$ & $K{=}4$ & $K{=}8$ & $K{=}32$ & density & query \\
\includegraphics[width=2cm]{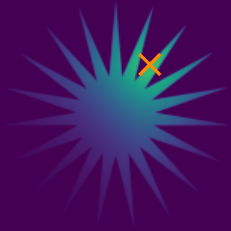}
& \includegraphics[width=2cm]{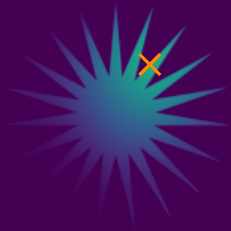}
& \includegraphics[width=2cm]{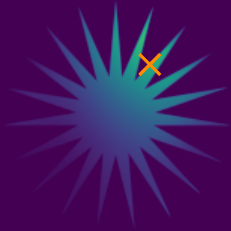}
& \includegraphics[width=2cm]{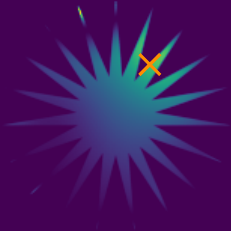}
& \includegraphics[width=2cm]{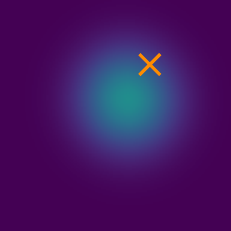}
& \includegraphics[width=2cm]{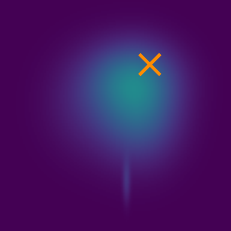}
& \includegraphics[width=2cm]{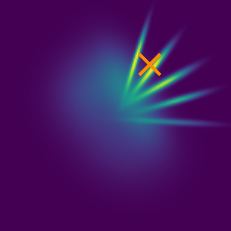}
& \includegraphics[width=2cm]{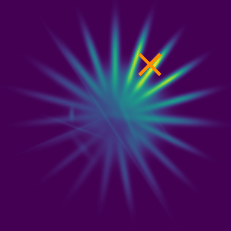}
& \includegraphics[width=2cm]{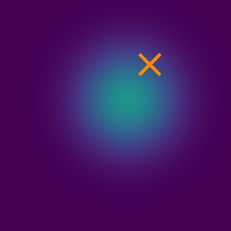}
& \includegraphics[width=2cm]{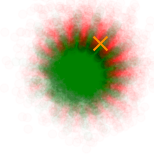}\\
\includegraphics[width=2cm]{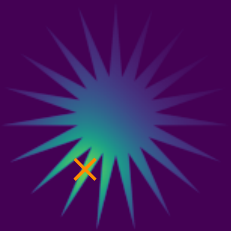}
& \includegraphics[width=2cm]{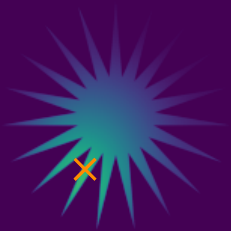}
& \includegraphics[width=2cm]{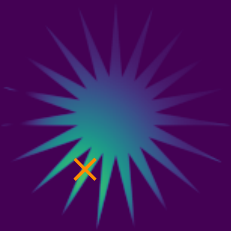}
& \includegraphics[width=2cm]{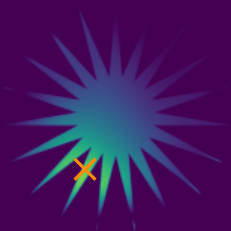}
& \includegraphics[width=2cm]{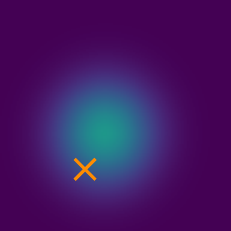}
& \includegraphics[width=2cm]{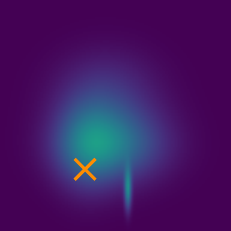}
& \includegraphics[width=2cm]{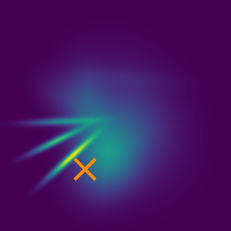}
& \includegraphics[width=2cm]{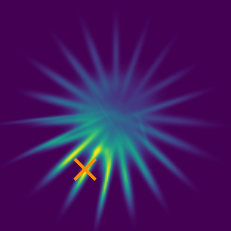}
& \includegraphics[width=2cm]{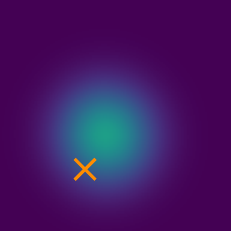}
& \includegraphics[width=2cm]{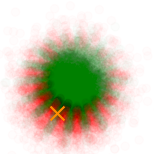}\\
\includegraphics[width=2cm]{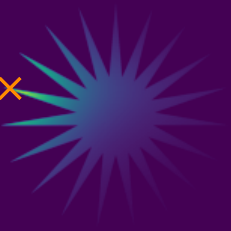}
& \includegraphics[width=2cm]{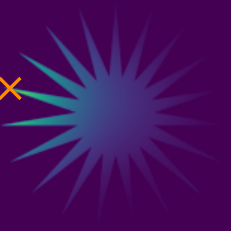}
& \includegraphics[width=2cm]{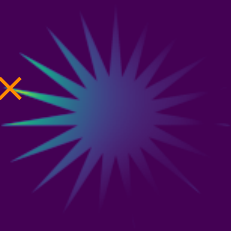}
& \includegraphics[width=2cm]{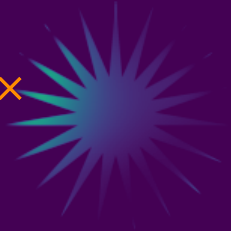}
& \includegraphics[width=2cm]{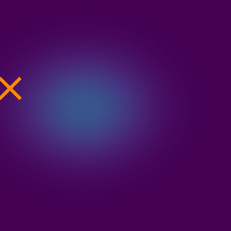}
& \includegraphics[width=2cm]{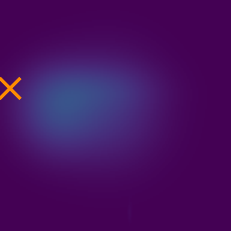}
& \includegraphics[width=2cm]{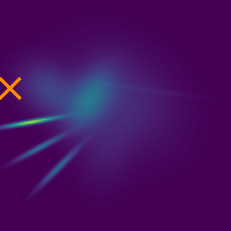}
& \includegraphics[width=2cm]{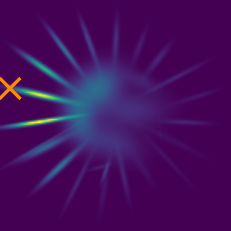}
& \includegraphics[width=2cm]{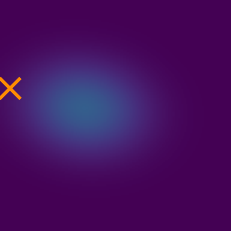}
& \includegraphics[width=2cm]{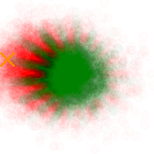}\\
\bottomrule
\end{tabular}
}
\end{adjustbox}
    \caption{Densities of the Ground-Truth compared to the polynomial, GMM and DeepSeaProbLog for the $19$-Star problem with a cauchy-density. For the DSP model selected by log-likelihood, we show the density of the neural distributional fact, and we also show the result of querying the ProbLog program representing our constraints $10000$ times. The samples associated with a true-label are shown in green, the samples associated with a false-label are shown in red.}
    \label{tab:19_star_1_cauchy_density}
\end{table}

\begin{table}[h!]
    \centering
    \begin{tabular}{p{2cm}p{2cm}p{2cm}}
\toprule
Ground Truth & \multicolumn{2}{c}{DeepSeaProblog (by loss)}\\
\cmidrule(lr){2-3}
& density & query \\
\includegraphics[width=2cm]{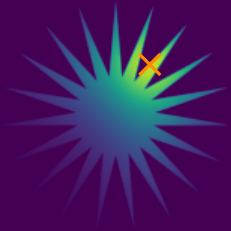}
& \includegraphics[width=2cm]{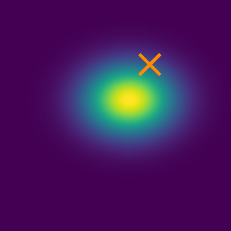}
& \includegraphics[width=2cm]{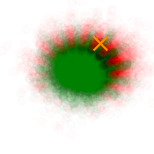}\\
\includegraphics[width=2cm]{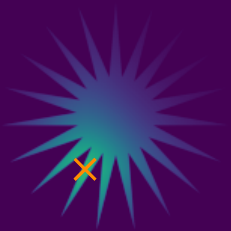}
& \includegraphics[width=2cm]{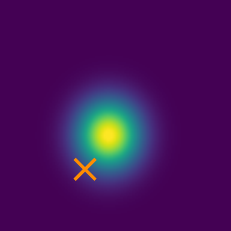}
& \includegraphics[width=2cm]{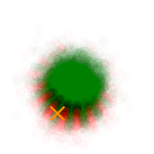}\\
\includegraphics[width=2cm]{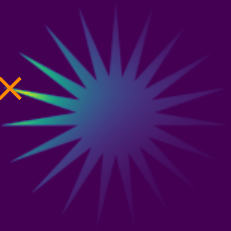}
& \includegraphics[width=2cm]{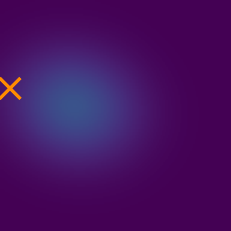}
& \includegraphics[width=2cm]{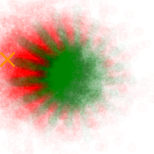}\\
\bottomrule
    \end{tabular}
    \caption{Densities of the Ground-Truth compared to the DeepSeaProbLog for the $19$Star problem with a cauchy-density. DeepSeaProbLog is selected by loss, and due to avoiding the constraints, more concentrated and therefore visualized separately. We show the density for the neural distributional fact and the samples obtained by the query. The samples associated with an true-label are shown in green, the samples associated with a false-label are shown in red. We choose to visualize this separately in order to keep the color-scheme in figure \ref{tab:19_star_1_cauchy_density} reasonable.}
    \label{tab:19_star_1_cauchy_density_deepsea_loss}
\end{table}

\FloatBarrier

\subsection{Stanford Drone Dataset - joint distribution}
\label{sec:appendix_experiments_sdd_marginal}

\subsubsection{Details of the Dataset - Scenario 1}
\label{sec:appendic_experiments_sdd_marginal_details}

We focus on image $12$, the image with the most trajectories. We first want to note that the time-resolution, due to it being extracted from a $30$-fps video, is quite high. In order to eliminate outliers, we first delete all points without movement (trajectories with a total variance of less than $20$ and points with a distance of less than $0.1$ compared to the previous). Then, we clean the data by discarding all short trajectories ($=$ length of less than $50$). We arrive at $415$ moving trajectories out of the $499$ we started with. We split this dataset, by trajectory, into train, test and validation ($70\%$, $15\%$ and $15\%$) and then concatenate all the points. We arrive at $124998$ train points, $26786$ validation points and $26786$ test points. While this appears like a significant of points, we want to stress that many are heavily autocorrelated due to the high resolution, and the actual, total, amount of trajectories is only $415$.

\FloatBarrier

\begin{figure}
    \centering
    \begin{minipage}{0.35\columnwidth}
    \centering
    Before Filtering:
    \includegraphics[height=1.0\textwidth]{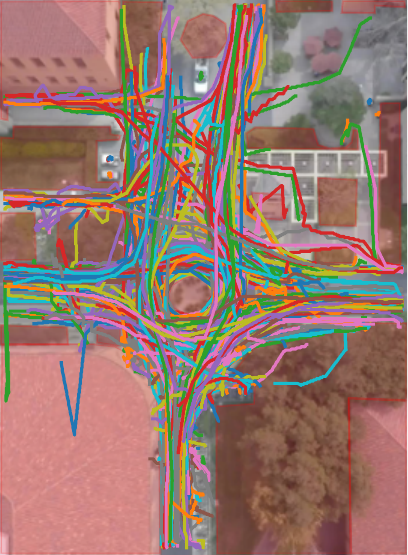}
    \end{minipage}
    \begin{minipage}{0.35\columnwidth}
    \centering
    After Filtering:
    \includegraphics[height=1.0\textwidth]{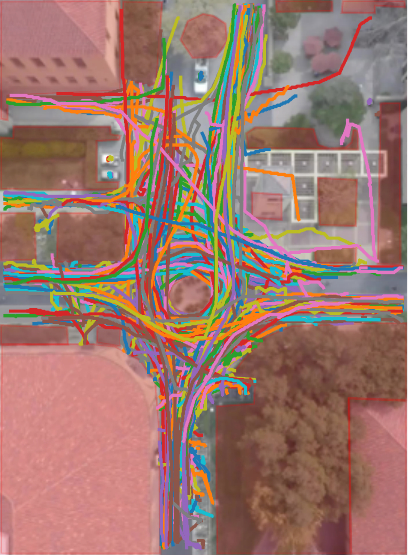}
    \end{minipage}
    \caption{The trajectories before/after filtering for image $12$ in the Stanford Drone Dataset. The constraints are shown in red.}
    \label{fig:sdd_trajectories_before_after}
\end{figure}

\begin{table}[h!]
\centering
\input{tables/sdd_marginal_big_replicated}

\caption{Results for the $p(\vY)$-case of the Stanford-Drone dataset for scenario 1. All Spline models have equal number of knots in $y_1$ and $y_2$. The average percent of probability mass covering invalid space ($\mathsf{Pr}(\neg \phi)$) over our test-set is given in percent. After choosing the hyper-parameters, all runs were repeated $10$-times and we report mean and standard deviation.
}
\label{tab:sdd_results_big_table}
\end{table}

\begin{table}[h!]
    \centering
    \begin{adjustbox}{center}
\resizebox{1.0\linewidth}{!}{%
    \begin{tabular}{p{2cm}p{2cm}p{2cm}p{2cm}p{2.2cm}p{2.2cm}p{2.2cm}p{2.2cm}p{2.2cm}}
Training Data & \ours ($8$ knots) & \ours ($10$ knots) & \ours ($12$ knots) & \ours ($14$ knots) & \ours ($16$ knots) \\
\includegraphics[width=2cm]{imgs/sdd_marginal/training_data_nbins_100_with_borders_hatched.png}
& \includegraphics[width=2cm]{imgs/sdd_marginal/hspl/spline_8_10_density_with_borders_below.png}
& \includegraphics[width=2cm]{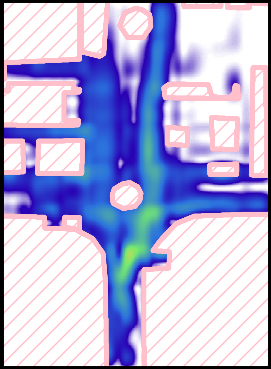}
& \includegraphics[width=2cm]{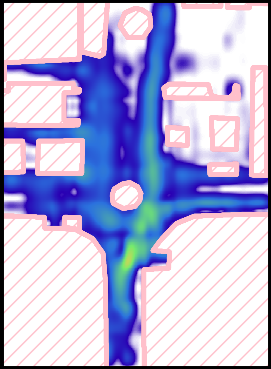}
& \includegraphics[width=2cm]{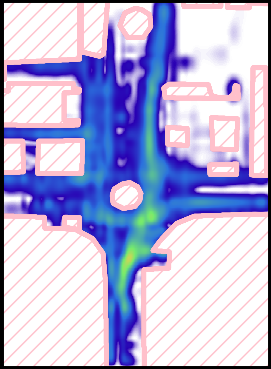}
& \includegraphics[width=2cm]{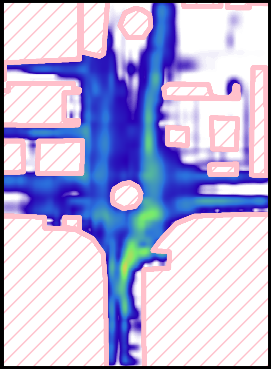}\\
& GMM 5-comp. & GMM 10-comp. & GMM 20-comp. & GMM 50-comp. & GMM 100-comp. \\
& \includegraphics[width=2cm]{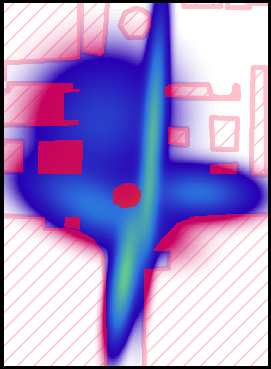}
& \includegraphics[width=2cm]{imgs/sdd_marginal/gmm/gmm_10_with_border_below.png}
& \includegraphics[width=2cm]{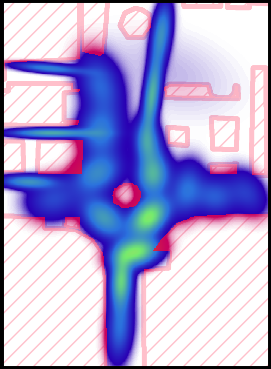}
& \includegraphics[width=2cm]{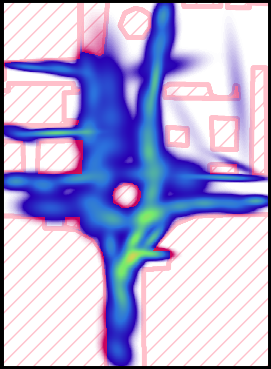}
& \includegraphics[width=2cm]{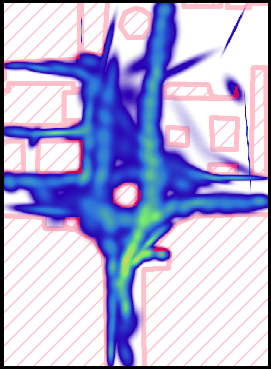}\\
& Flow $t=1$ & Flow $t=2$ & Flow $t=5$ & Flow $t=10$\\
& \includegraphics[width=2cm]{imgs/sdd_marginal/flow/flow_1_128_density_with_borders_below.png}
& \includegraphics[width=2cm]{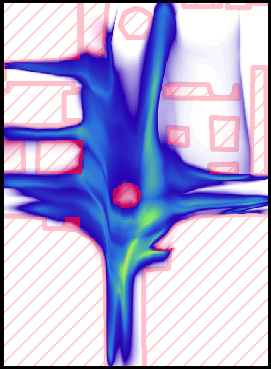}
& \includegraphics[width=2cm]{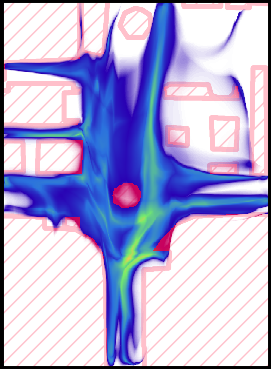}
& \includegraphics[width=2cm]{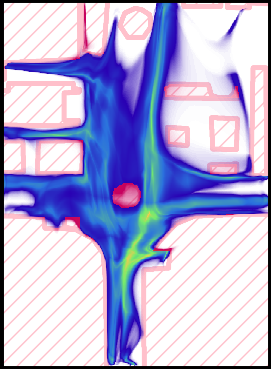}
\end{tabular}
}
\end{adjustbox}
    \caption{Densities on the $p(\vY)$-case of the Stanford-Drone dataset for scenario 1. All Spline models have $10$-mixture components and equal number of knots in $Y_1$ and $Y_2$. The flow-models have two hidden layers of size $128$ per transformation.}
    \label{tab:sdd_marginal_densities}
\end{table}

\FloatBarrier

\subsubsection{Details of the Dataset - Scenario 2}
\label{sec:appendic_experiments_sdd_marginal_details_2}

We focus on image $2$, which consists only of 119 trajectories. We perform the same filtering etc. as in scenario 1 and arrive at  $53504$ train points, $10197$ validation points and $13860$ test points. 

\begin{table}[h!]
\centering
\input{tables/sdd_marginal_big_2_replicated}    
\caption{Results for the $p(\vY)$-case of the Stanford-Drone dataset for scenario 2. All Spline models have equal number of knots in $y_1$ and $y_2$. The average percent of probability mass covering invalid space ($\mathsf{Pr}(\neg \phi)$) over our test-set is given in percent. After choosing the hyper-parameters, all runs were repeated $10$-times and we report mean and standard deviation.
}
\label{tab:sdd_2_results_big_table}
\end{table}

\begin{table}[h!]
    \centering
    \begin{adjustbox}{center}
\resizebox{1.0\linewidth}{!}{%
    \begin{tabular}{p{2cm}p{2cm}p{2cm}p{2cm}p{2.2cm}p{2.2cm}p{2.2cm}p{2.2cm}p{2.2cm}}
Training Data & \ours ($8$ knots) & \ours ($10$ knots) & \ours ($12$ knots) & \ours ($14$ knots) & \ours ($16$ knots) \\
\includegraphics[width=2cm]{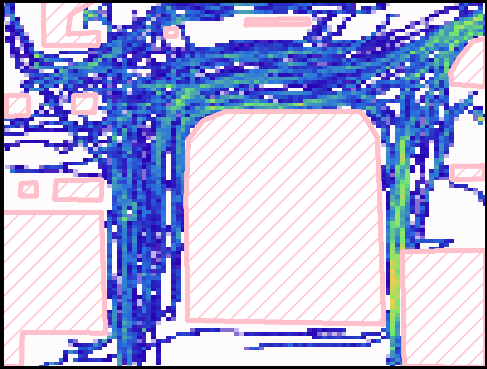}
& \includegraphics[width=2cm]{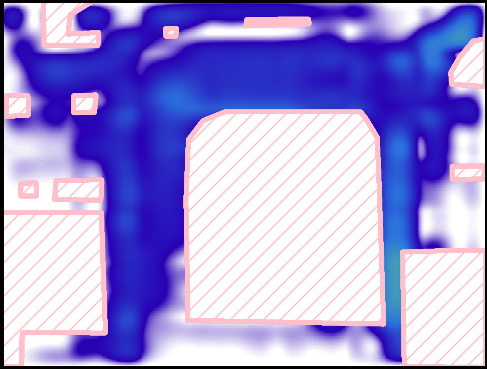}
& \includegraphics[width=2cm]{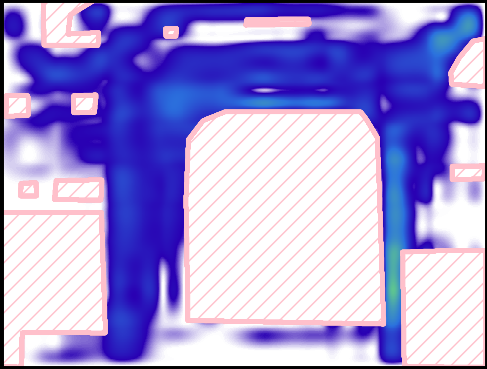}
& \includegraphics[width=2cm]{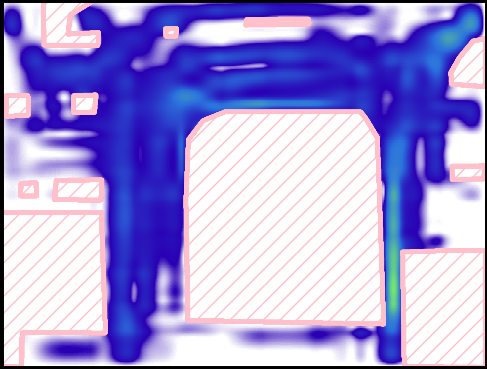}
& \includegraphics[width=2cm]{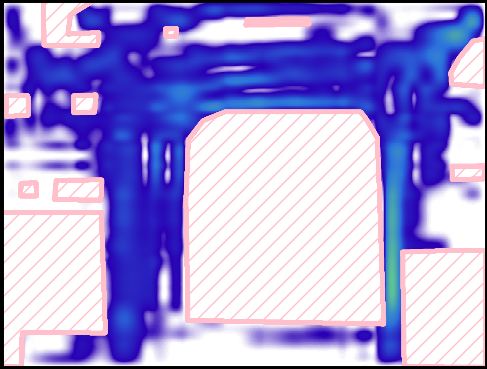}
& \includegraphics[width=2cm]{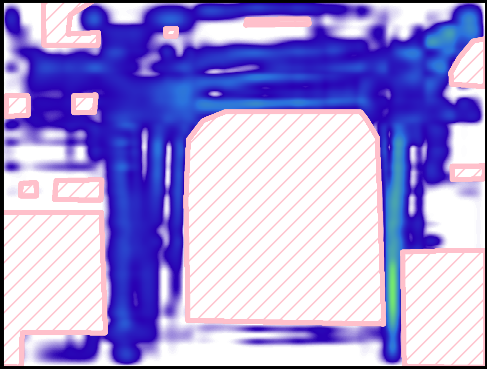}\\
& GMM 5-comp. & GMM 10-comp. & GMM 20-comp. & GMM 50-comp. & GMM 100-comp. \\
& \includegraphics[width=2cm]{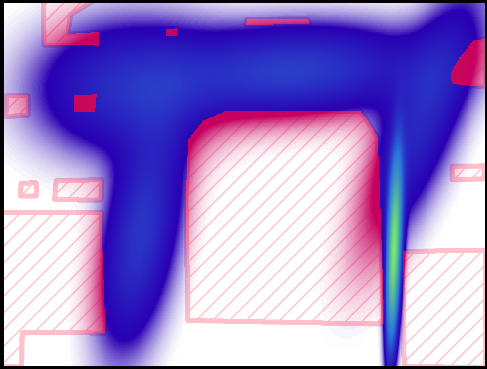}
& \includegraphics[width=2cm]{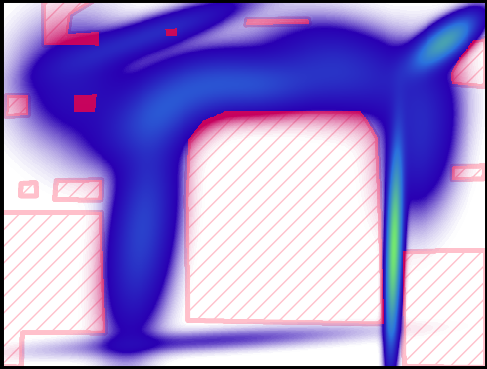}
& \includegraphics[width=2cm]{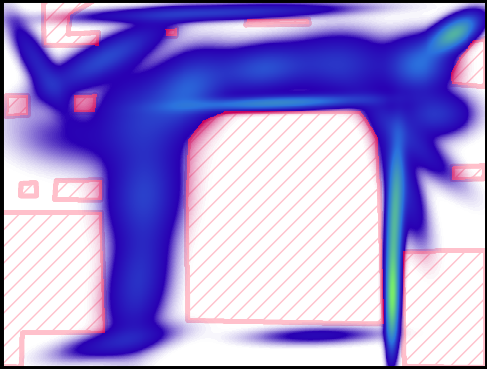}
& \includegraphics[width=2cm]{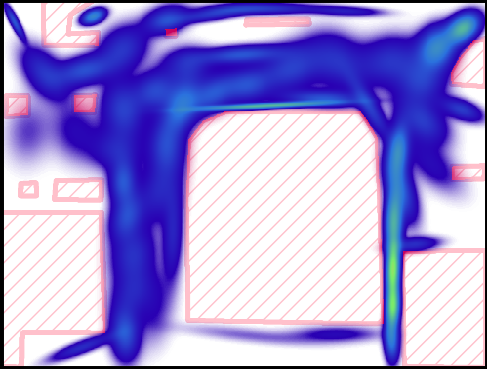}
& \includegraphics[width=2cm]{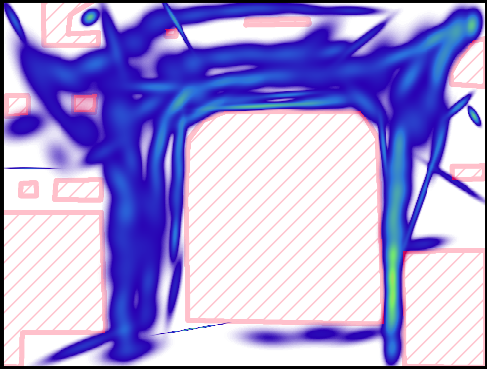}\\
& Flow $t=1$ & Flow $t=2$ & Flow $t=5$ & Flow $t=10$\\
& \includegraphics[width=2cm]{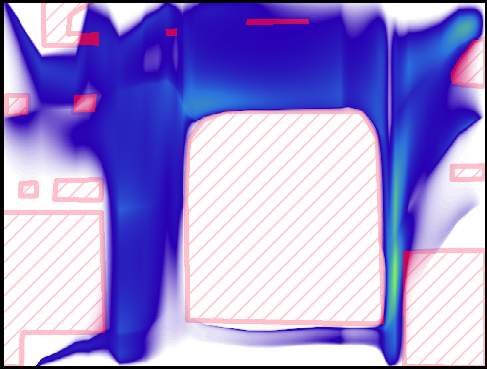}
& \includegraphics[width=2cm]{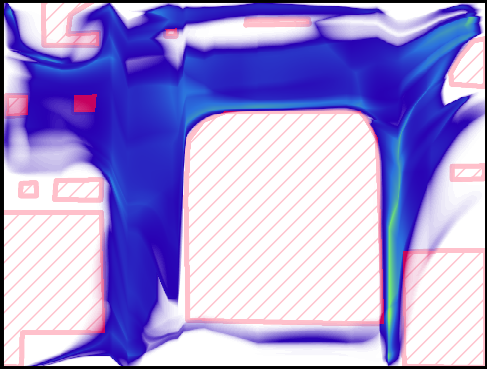}
& \includegraphics[width=2cm]{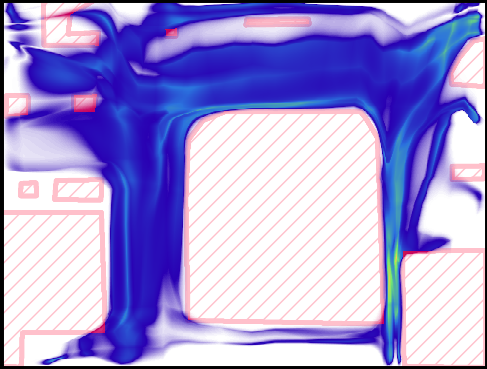}
& \includegraphics[width=2cm]{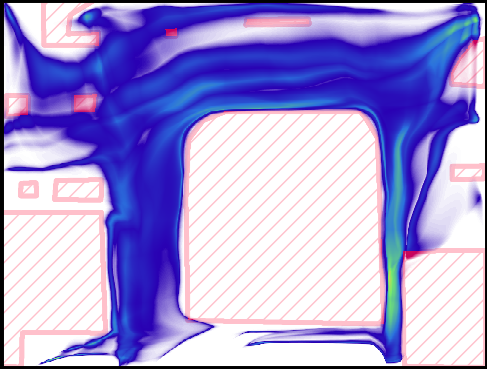}
\end{tabular}
}
\end{adjustbox}
    \caption{Densities on the $p(\vY)$-case of the Stanford-Drone dataset for scenario 2. All Spline models have $10$-mixture components and equal number of knots in $Y_1$ and $Y_2$. The flow-models have two hidden layers of size $32x2$ per transformation.}
    \label{tab:sdd_2_marginal_densities}
\end{table}

\FloatBarrier

\subsection{Stanford Drone Dataset - conditional distributions}
\label{app:sdd_conditional}

\subsubsection{Details of the Dataset}

The goal in this task is to fit the distribution of possible future trajectories, so if our random variable for the coordinates at timestep $t$ is $\vC_t$, then our goal is to predict $\vY=\vC_{t\geq t'}$ given the current a window of 5 equidistant points $\vX = (\vC_{t = t' - 0\cdot \Delta}, \vC_{t = t' - 1\cdot \Delta}, \dots, , \vC_{t = t' - 5\cdot \Delta})$, with $\Delta$ being some step-size. This challenging construction induces multi-modality and uncertainty in the predictive distribution, because if there is a chance of visiting a certain area in the future given the $5$ steps, it must have some probability mass assigned to it. 

In order to bias our models towards more well-connected paths, we bias the network towards the closer in time datapoints. The distribution is therefore a mixture of both a uniform distribution over the whole future trajectory and a uniform distribution over the future trajectory of length step-size $s$, so $70$ in our case. Both choices have equal chance.

\subsubsection{Scenario 1}

For this task, we focus on image $12$, the image with the most trajectories. We want to note that the time-resolution, due to it being extracted from a $30$-fps video, is quite high. As we want to focus on the trajectory, we eliminate all points without movement, so points with a distance of less than $0.1$ compared to the previous. We take a step-size of $70$ for the window of $5$ points that form our $\vX$, which we will slide through the trajectory. We also discard all trajectories that are too short to fill our window, so where the length is less than $5 \cdot 70$, as we want to focus on long trajectories. We split the trajectories into $70\%$ train, $15\%$ validation and $15\%$ test. We then create a static validation and test-dataset by sampling $10$ future $\vY$ points per window $\vX=\vx$ at creation statically. For the train-dataset, we sample the during training, so for the same $\vX=\vx$ it will see different future points. We arrive at $22619$-train datapoints, with $\vY$ dynamically sampled per $\vX=\vx$, $42740$-validation datapoints and $49660$-test datapoints.\\

\FloatBarrier

\textbf{Model Details}\newline
\label{app:sdd_scenario1_model_details}

All our models are simple, fully connected neural networks with ReLu as an activation function \citep{DBLP:journals/jmlr/GlorotBB11}.

We denote the following sizes:
\begin{itemize}
    \item \textbf{large}: $2$ hidden layers of size $2048$ each
    \item \textbf{medium}: $2$ hidden layers of size $1024$ each
    \item \textbf{small}: $2$ hidden layers of size $512$ each
\end{itemize}

We train all models for a maximum of $500$ epochs with a patience of $20$ epochs and run the hyper-parameter search for each network-size per model-type.

\textbf{\ours} For the \ours models, we do a grid-search over the following parameters:
\begin{itemize}
    \item \textbf{number of mixtures}: $8$ and $10$
    \item \textbf{number of knots}: $10$ and $14$ (equal over $y_1$ and $y_2)$
    \item \textbf{optimizer} AdamW schedulefree \citep{Defazio2024TheRL} with learning rate: $0.001$, $0.0001$, $0.00001$ \newline and batch-sizes $16$, $32$ and $128$
    \item \textbf{net-sizes}: large/medium/small
\end{itemize}

\textbf{GMM} For the conditional GMM-models, we do a grid-search over the following parameters:
\begin{itemize}
    \item \textbf{number of components {$K$}}: $4$, $32$, $50$, $80$, $100$ with full covariances
    \item \textbf{net-sizes}: large/medium/small
    \item \textbf{optimizer} Adam \citep{KingmaB14@adam} with learning rates:  $0.001$, $0.0001$, $0.00001$  and batch-sizes $16$, $32$ \newline ($128$ led to worse performance due to overfitting on initial-runs and was excluded)
\end{itemize}

\textbf{DSP} For the DSP models, we do a grid-search over the following parameters:
\begin{itemize}
    \item \textbf{optimizer} AdaMax \citep{KingmaB14@adam} with learning rates $0.01$, $0.001$, $0.0001$, $0.00001$ and batch-sizes $16$, $32$ \newline ($128$ led to worse performance due to overfitting on initial-runs and was excluded)
    \item \textbf{annealing starting-multiplier}: $0.1$, $1.0$
    \item \textbf{end-multiplier}: $5$
    \item \textbf{net-sizes}: large/medium/small
\end{itemize}
We use a tanh-scaling of the annealing multiplier with an alpha of $1e-4$ and train with the loss \ref{eq:dsp_loss}.\\

\FloatBarrier

\textbf{Results}\newline

\begin{table}[h!]
\centering
\input{tables/sdd_overview_replicated}
\caption{Results for the $p(\vY|\vX)$-case of the Stanford-Drone dataset for scenario 1. All Spline models have equal number of knots in $y_1$ and $y_2$. The average percent of probability mass covering invalid space ($\mathsf{Pr}(\neg \phi)$) over our test-set is given in percent. It is approximated by sampling $10^6$ times per datapoint $\vx$ in the test-set, computing constraint satisfaction, and then taking the average. After choosing the hyper-parameters, all runs were repeated $10$-times and we report mean and standard deviation.}
\label{tab:sdd_traj_big_table}
\end{table}

\begin{table}[h!]
    \centering
    \begin{adjustbox}{center}
\resizebox{1.0\linewidth}{!}{%
    \begin{tabular}{cp{2cm}p{2cm}p{2cm}p{2.2cm}p{2.2cm}p{2.2cm}}
Model & Sample 1 & Sample 2 & Sample 3 & Sample 4 & Sample 5 & Sample 6\\
\midrule
Spline 10/small &
\includegraphics[width=1.7cm]{imgs/sdd/appendix/26891/spline_10_small.png}
& \includegraphics[width=1.7cm]{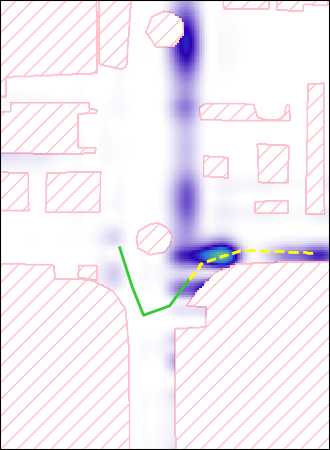}
& \includegraphics[width=1.7cm]{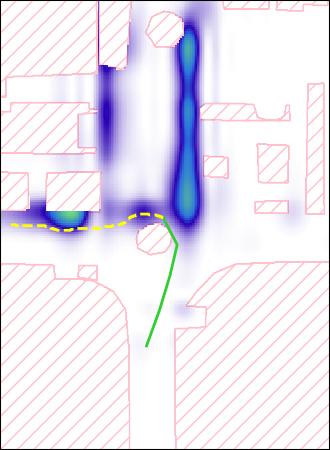}
& \includegraphics[width=1.7cm]{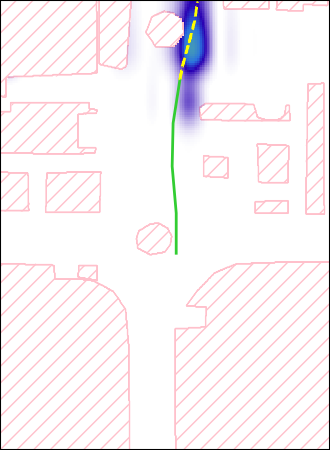}
& \includegraphics[width=1.7cm]{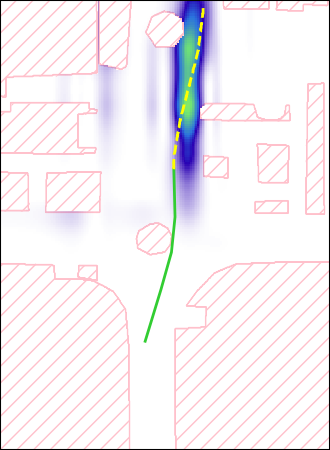}
& \includegraphics[width=1.7cm]{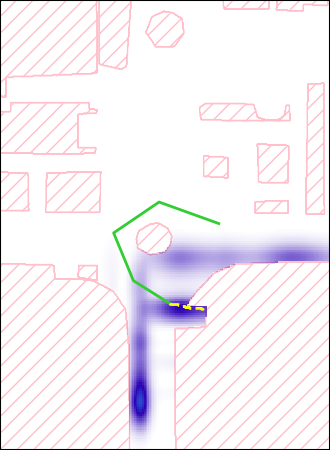}\\
Spline 14/small &
\includegraphics[width=1.7cm]{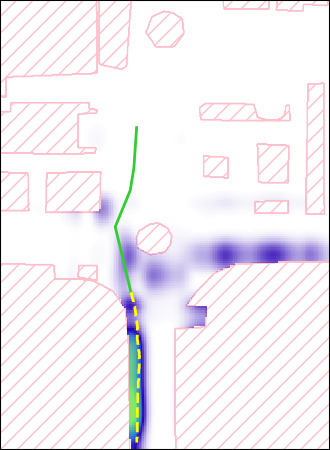}
& \includegraphics[width=1.7cm]{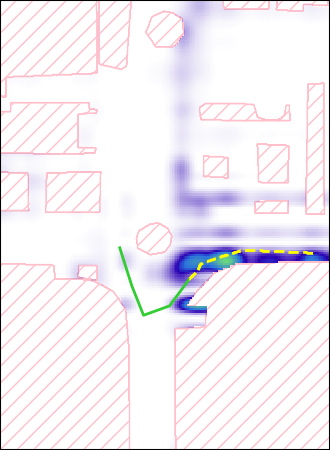}
& \includegraphics[width=1.7cm]{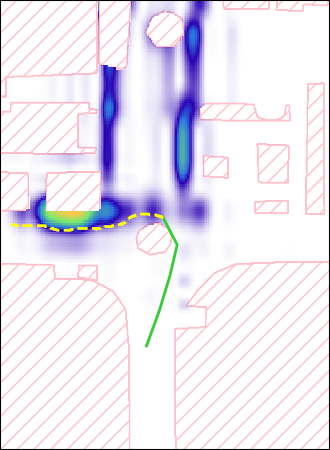}
& \includegraphics[width=1.7cm]{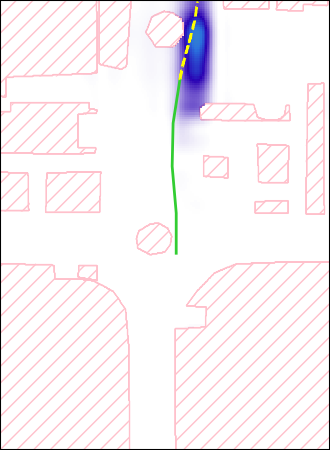}
& \includegraphics[width=1.7cm]{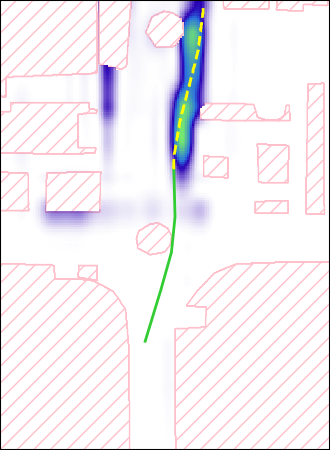}
& \includegraphics[width=1.7cm]{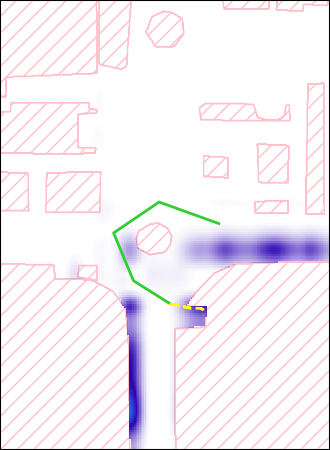}\\
Spline 14/large &
\includegraphics[width=1.7cm]{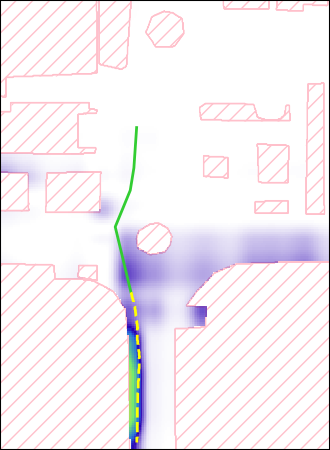}
& \includegraphics[width=1.7cm]{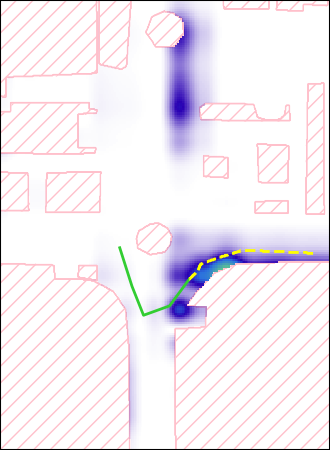}
& \includegraphics[width=1.7cm]{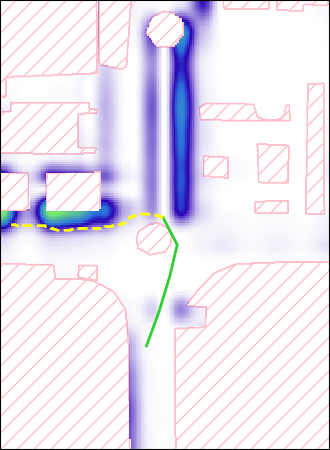}
& \includegraphics[width=1.7cm]{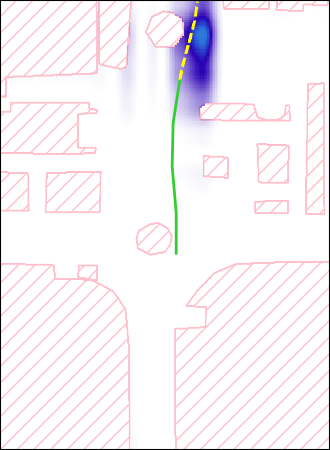}
& \includegraphics[width=1.7cm]{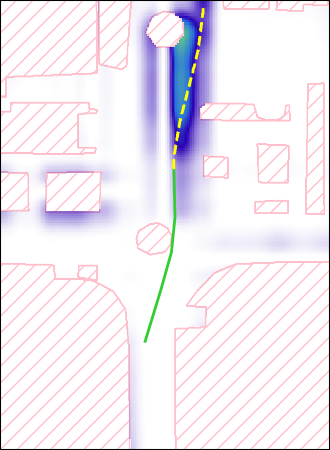}
& \includegraphics[width=1.7cm]{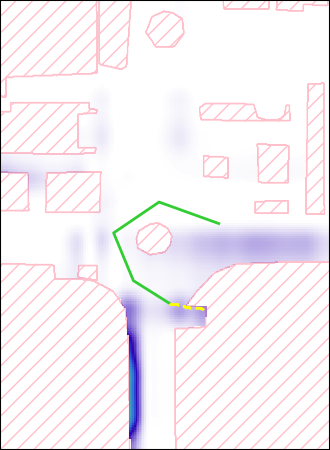}\\
Spline 14/medium &
\includegraphics[width=1.7cm]{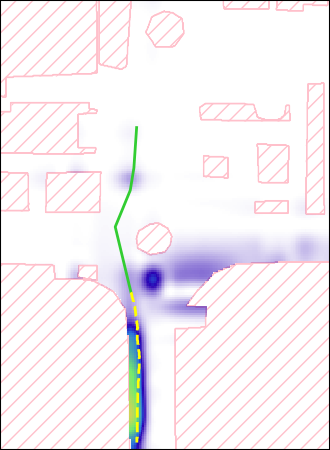}
& \includegraphics[width=1.7cm]{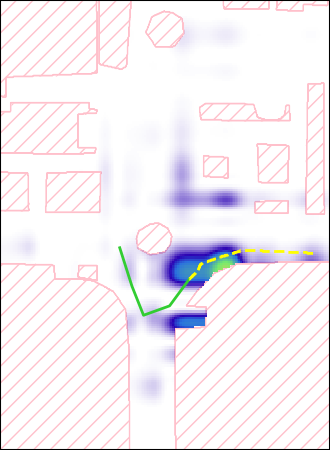}
& \includegraphics[width=1.7cm]{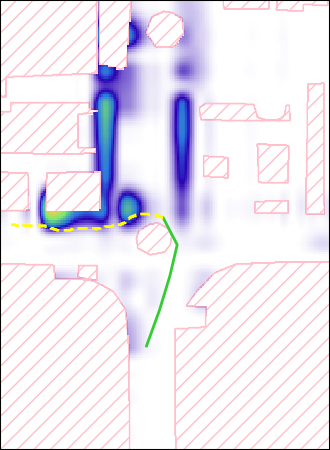}
& \includegraphics[width=1.7cm]{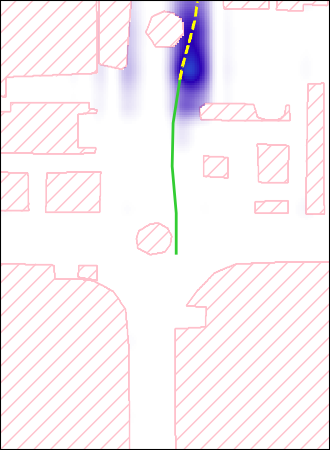}
& \includegraphics[width=1.7cm]{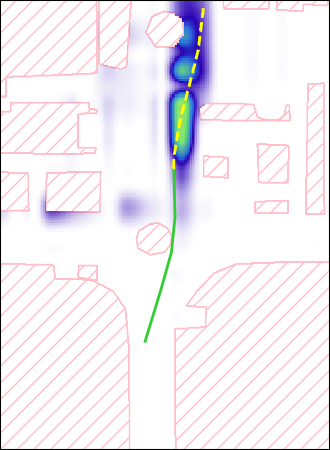}
& \includegraphics[width=1.7cm]{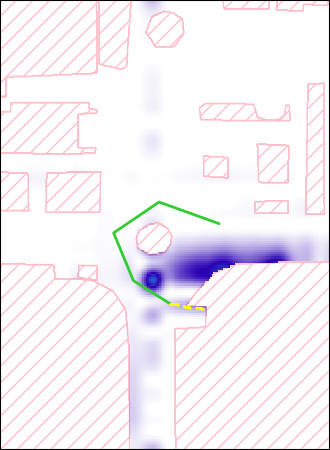}\\
GMM 50/medium &
\includegraphics[width=1.7cm]{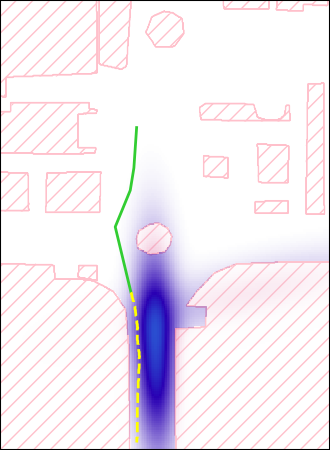}
& \includegraphics[width=1.7cm]{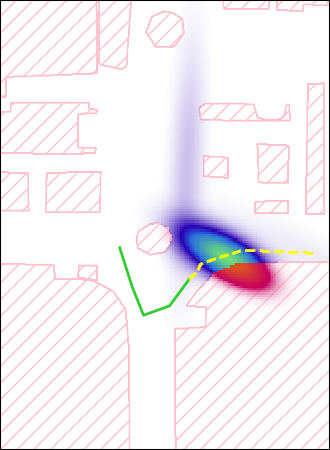}
& \includegraphics[width=1.7cm]{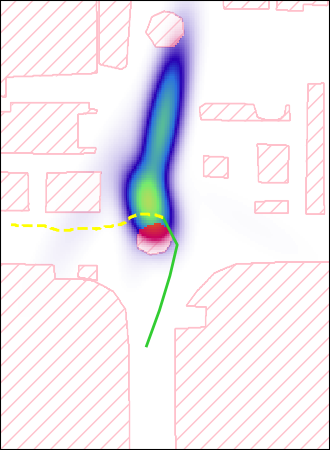}
& \includegraphics[width=1.7cm]{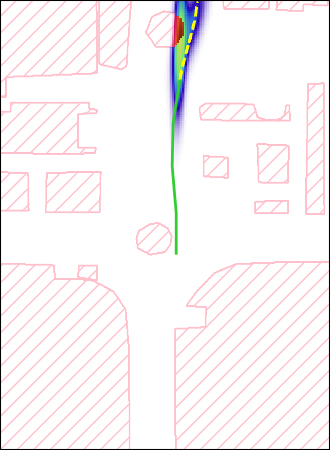}
& \includegraphics[width=1.7cm]{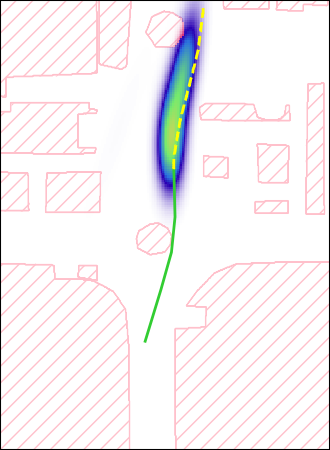}
& \includegraphics[width=1.7cm]{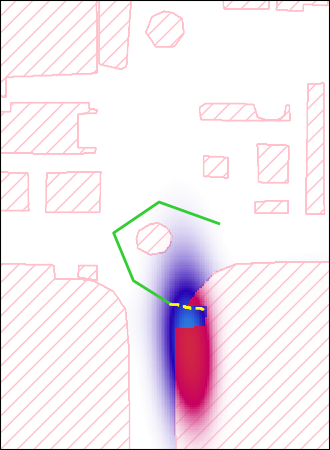}\\
GMM 80/medium &
\includegraphics[width=1.7cm]{imgs/sdd/appendix/26891/gmm_80_medium.png}
& \includegraphics[width=1.7cm]{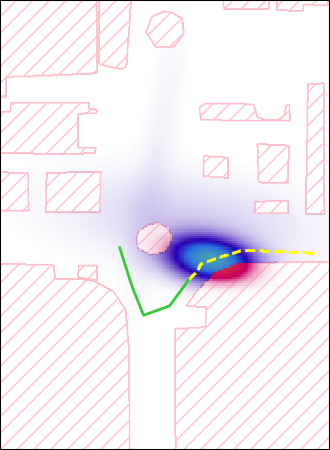}
& \includegraphics[width=1.7cm]{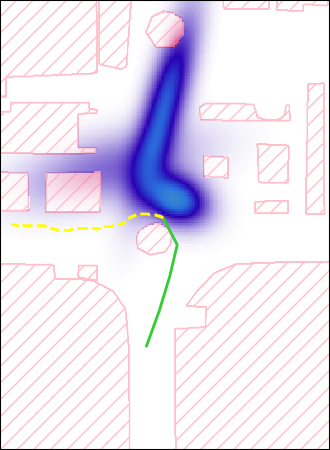}
& \includegraphics[width=1.7cm]{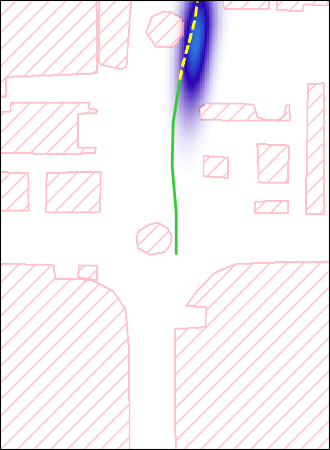}
& \includegraphics[width=1.7cm]{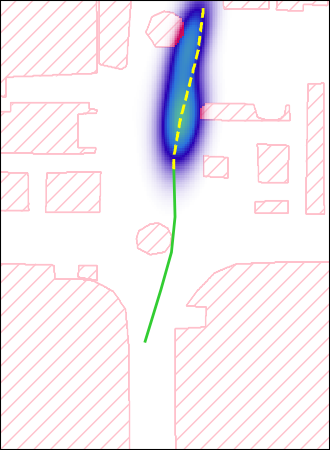}
& \includegraphics[width=1.7cm]{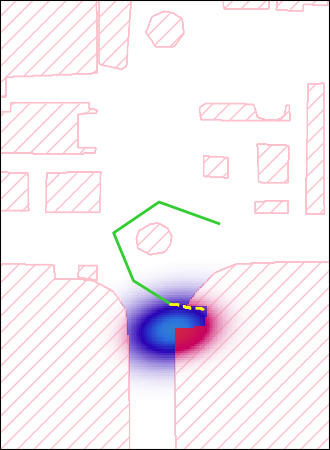}\\
GMM 50/small &
\includegraphics[width=1.7cm]{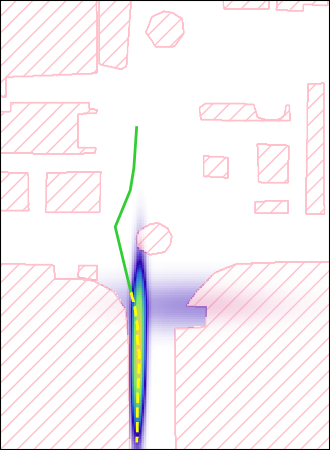}
& \includegraphics[width=1.7cm]{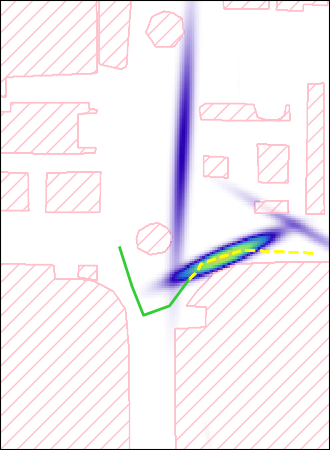}
& \includegraphics[width=1.7cm]{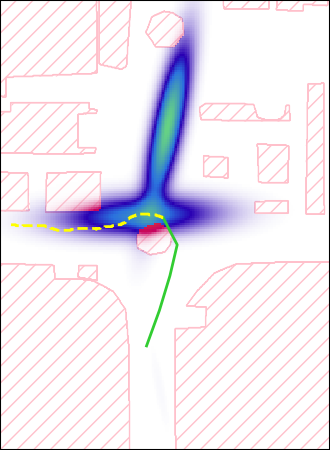}
& \includegraphics[width=1.7cm]{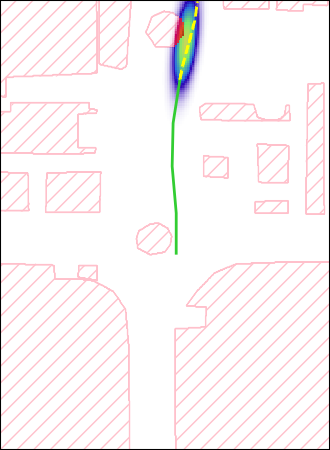}
& \includegraphics[width=1.7cm]{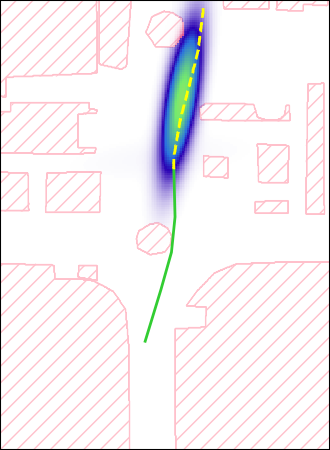}
& \includegraphics[width=1.7cm]{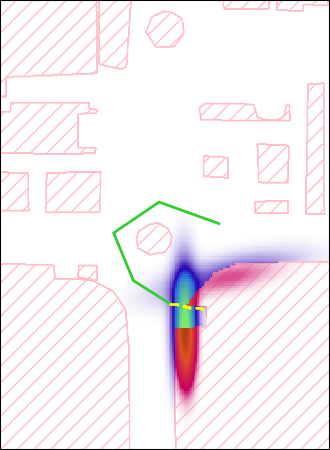}\\
GMM 100/small &
\includegraphics[width=1.7cm]{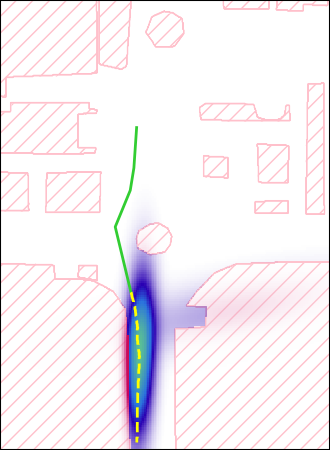}
& \includegraphics[width=1.7cm]{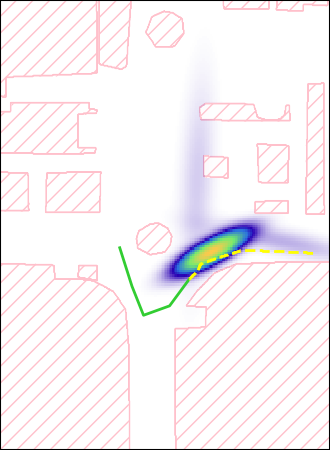}
& \includegraphics[width=1.7cm]{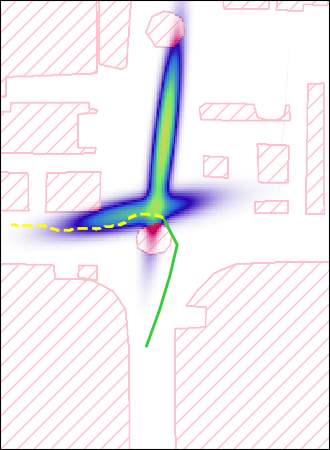}
& \includegraphics[width=1.7cm]{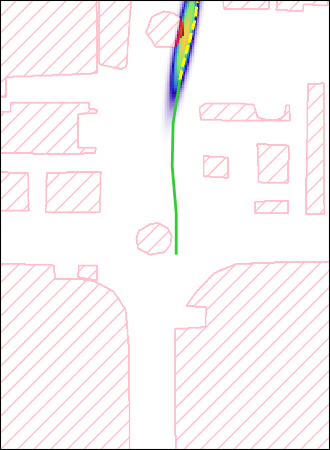}
& \includegraphics[width=1.7cm]{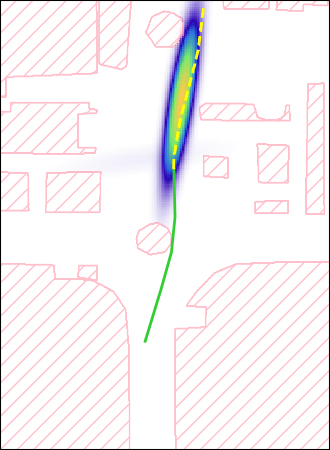}
& \includegraphics[width=1.7cm]{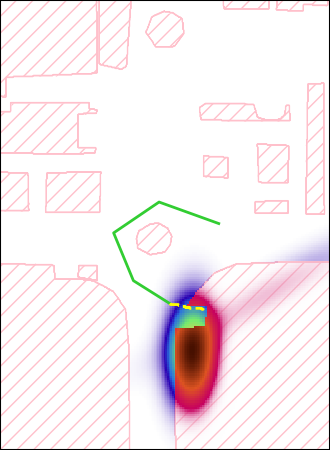}\\
DSP by loss, large &
\includegraphics[width=1.7cm]{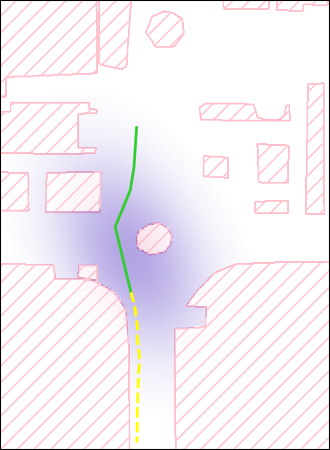}
& \includegraphics[width=1.7cm]{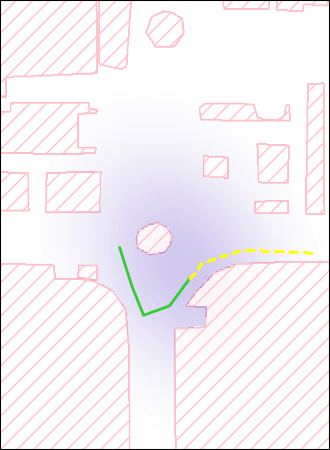}
& \includegraphics[width=1.7cm]{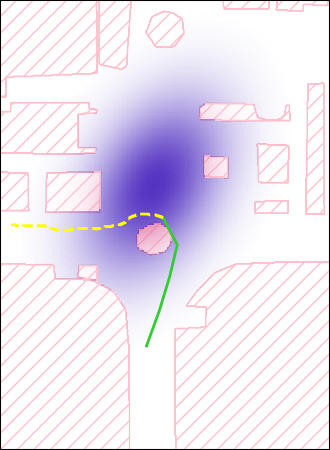}
& \includegraphics[width=1.7cm]{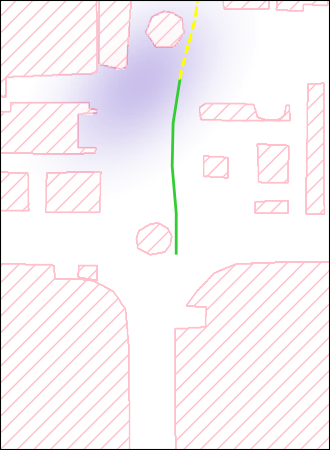}
& \includegraphics[width=1.7cm]{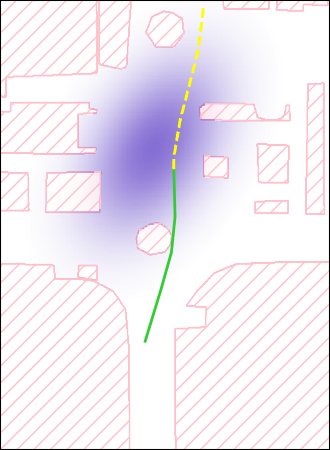}
& \includegraphics[width=1.7cm]{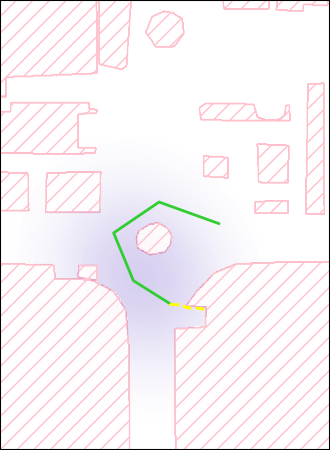}\\
DSP by log-like, large &
\includegraphics[width=1.7cm]{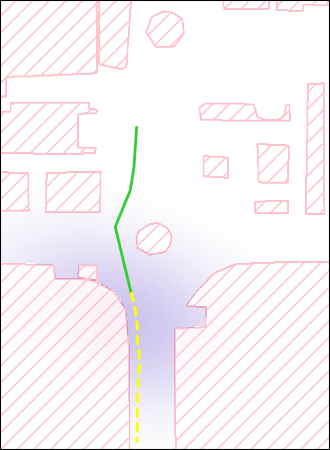}
& \includegraphics[width=1.7cm]{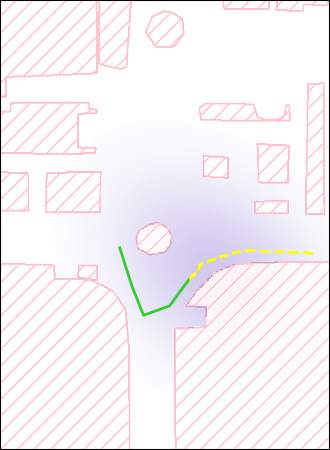}
& \includegraphics[width=1.7cm]{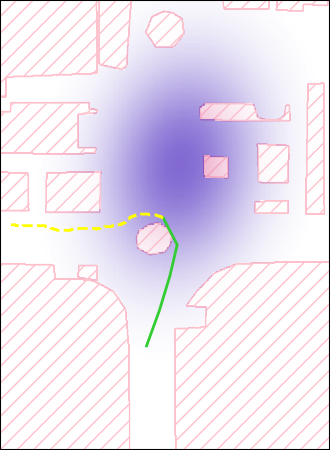}
& \includegraphics[width=1.7cm]{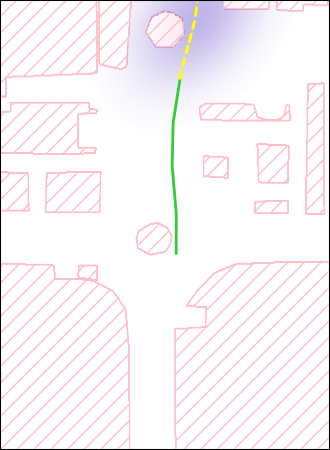}
& \includegraphics[width=1.7cm]{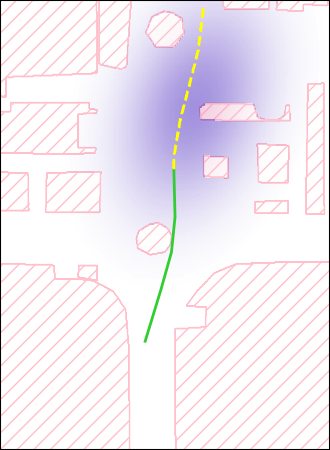}
& \includegraphics[width=1.7cm]{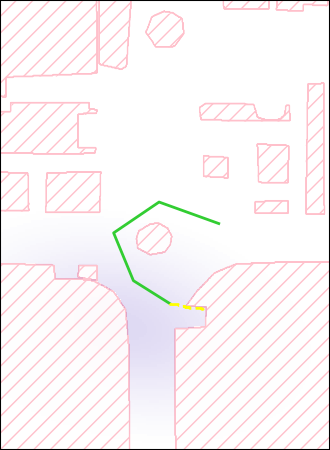}\\
\end{tabular}
}
\end{adjustbox}
    \caption{Densities for the predictive positions for the $P(\vY \mid \vX)$ case on the stanford drone dataset for scenario $1$. We compare the best $4$ spline-models against the best $4$ GMM models and the best DSP models both by log-likelihood and loss from \ref{tab:sdd_traj_big_table}. The colormap is normalized per sample.}
    \label{tab:sdd_conditional_predictions_scene_1}
\end{table}

\begin{table}[h!]
    \centering
    \begin{adjustbox}{center}
\resizebox{1.0\linewidth}{!}{%
    \begin{tabular}{cp{2cm}p{2cm}p{2cm}p{2.2cm}p{2.2cm}p{2.2cm}}
Model & Sample 1 & Sample 2 & Sample 3 & Sample 4 & Sample 5 & Sample 6\\
\midrule
by log-like, large network &
\includegraphics[width=1.7cm]{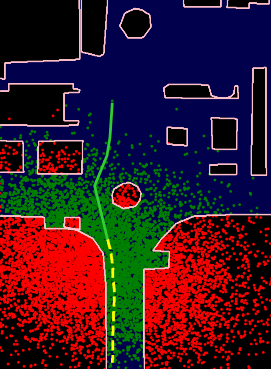}
& \includegraphics[width=1.7cm]{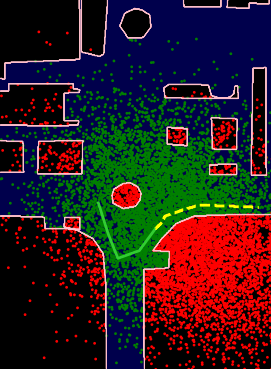}
& \includegraphics[width=1.7cm]{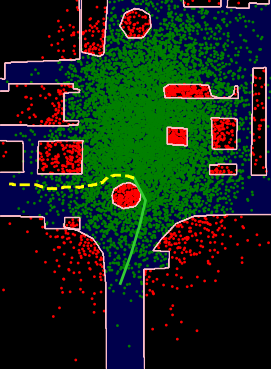}
& \includegraphics[width=1.7cm]{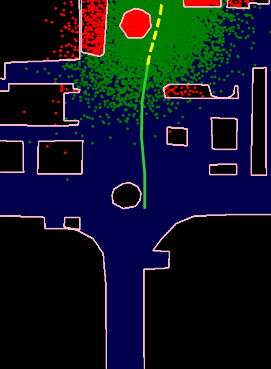}
& \includegraphics[width=1.7cm]{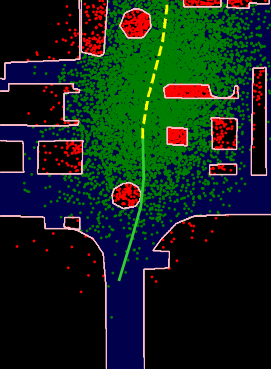}
& \includegraphics[width=1.7cm]{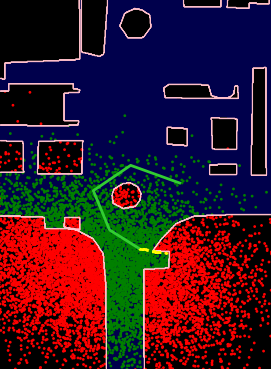}\\
by loss, large network&
\includegraphics[width=1.7cm]{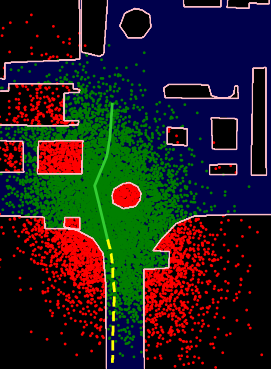}
& \includegraphics[width=1.7cm]{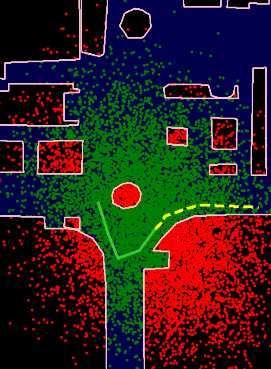}
& \includegraphics[width=1.7cm]{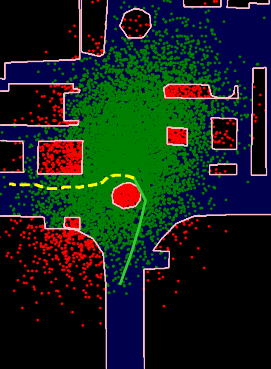}
& \includegraphics[width=1.7cm]{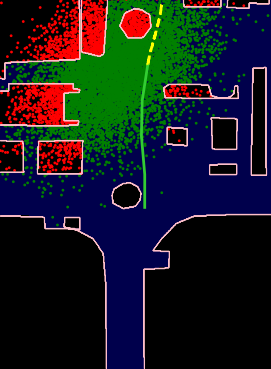}
& \includegraphics[width=1.7cm]{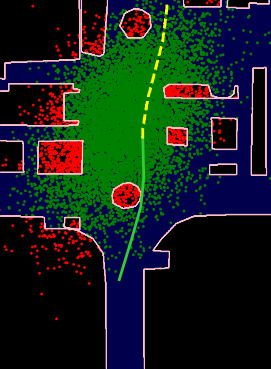}
& \includegraphics[width=1.7cm]{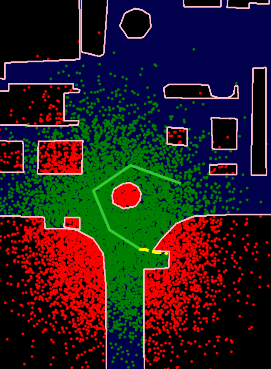}\\
\end{tabular}
}
\end{adjustbox}
    \caption{Samples from the Problog-Query representing the Constraints on the Stanford-Drone Dataset. We display the samples and the associated labels that we obtain from DeepSeaProbLog. We show the same samples as displayed in \ref{tab:sdd_conditional_predictions_scene_1}.}
    \label{tab:sdd_conditional_dspl_query_scene_1}
\end{table}

\FloatBarrier

\subsubsection{Scenario 2}
\label{app:further_sdd}

We provide further provide results on Image 2 from the Stanford Drone Dataset \citep{DBLP:conf/eccv/RobicquetSAS16}. This is a small scenario, with $119$ trajectories to start. For the conditional trajectory-prediction problem, we use the same setup is detailed in section \ref{app:sdd_conditional} and arrive at $21273$ train, $72710$ validation, and $62080$ test-datapoints.\\

\textbf{Model Details}\newline

We narrow down our search-space and pick the best-performing configurations from \ref{app:sdd_scenario1_model_details} to apply our grid-search on, but discard the large net-size as the dataset is smaller.

\textbf{\ours} For the \ours models, we do a grid-search over the following parameters:
\begin{itemize}
    \item \textbf{number of mixtures}: $8$ and $10$
    \item \textbf{number of knots}: $10$ and $14$ (equal over $y_1$ and $y_2)$
    \item \textbf{optimizer} AdamW schedulefree \citep{Defazio2024TheRL} with learning rate: $0.001$, $0.0001$ \newline and batch-sizes $16$, $32$ and $128$
    \item \textbf{net-sizes}: medium/small
\end{itemize}

\textbf{GMM} For the conditional GMM-models, we do a grid-search over the following parameters:
\begin{itemize}
    \item \textbf{number of components {$K$}}: $4$, $32$, $50$, $100$ with full covariances
    \item \textbf{net-sizes}: medium/small
    \item \textbf{optimizer} Adam \citep{KingmaB14@adam} with learning rates:  $0.001$, $0.0001$  and batch-sizes $16$, $32$, $128$
\end{itemize}

\textbf{DSP} For the DSP models, we do a grid-search over the following parameters:
\begin{itemize}
    \item \textbf{optimizer} AdaMax \citep{KingmaB14@adam} with learning rates  $0.001$, $0.0001$ and batch-sizes $16$, $32$, $128$
    \item \textbf{annealing starting-multiplier}: $0.1$, $1.0$
    \item \textbf{end-multiplier}: $5$
    \item \textbf{net-sizes}: medium/small
\end{itemize}
We use a tanh-scaling of the annealing multiplier with an alpha of $1e-4$ and train with the loss \ref{eq:dsp_loss}.\\

\textbf{Results}\newline

\begin{figure}
\hfill
\begin{minipage}{0.32\columnwidth}
    \centering
    Image
    \includegraphics[width=0.9\textwidth]{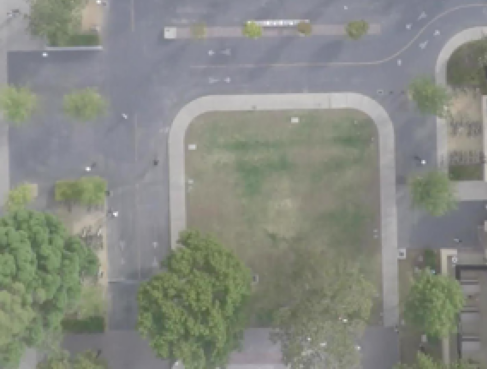} %
\end{minipage}
\hfill
\begin{minipage}{0.32\columnwidth}
    \centering
    Trajectories
    \includegraphics[width=0.9\textwidth, keepaspectratio]{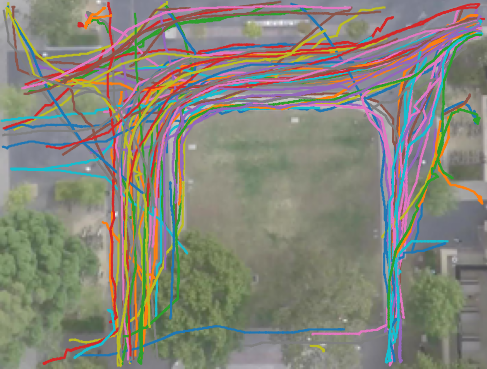}
\end{minipage}
\hfill
\begin{minipage}{0.32\columnwidth}
    \centering
    Constraints
    \includegraphics[width=0.9\textwidth, keepaspectratio]{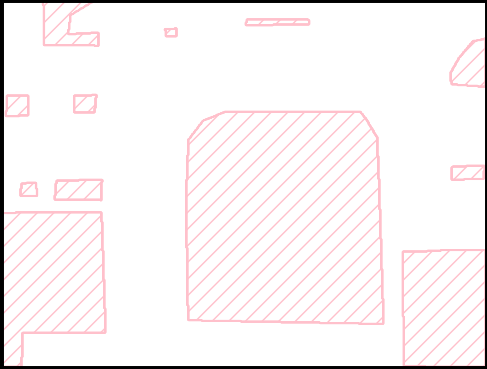}
\end{minipage}
\caption{\textbf{Our dataset combines challenging constraints with real-world data} on trajectories (middle) and aerial maps (left) taken from \citet{DBLP:conf/eccv/RobicquetSAS16}. We manually label the data, indicating invalid areas to move to. This is image 2 from the Stanford drone dataset.}
\label{fig:sdd_constraints_img_2}
\end{figure}

\begin{table}[h!]
\centering

\input{tables/sdd_overview_2_replicated}

\caption{Results for the $P(Y|X)$-case of the Stanford-Drone dataset for scenario $2$. All Spline models have equal number of knots in $y_1$ and $y_2$. The average percent of probability mass covering invalid space ($\mathsf{Pr}(\neg \phi)$) over our test-set is given in percent. It is approximated by sampling $10^6$ times per datapoint $\vx$ in the test-set, computing constraint satisfaction, and then taking the average. After choosing the hyper-parameters, all runs were repeated $10$-times and we report mean and standard deviation.}
\label{tab:sdd_traj_big_table_image_2}
\end{table}

\begin{table}[h!]
    \centering
    \begin{adjustbox}{center}
\resizebox{1.0\linewidth}{!}{%
    \begin{tabular}{cp{2cm}p{2cm}p{2cm}p{2.2cm}p{2.2cm}p{2.2cm}}
Model & Sample 1 & Sample 2 & Sample 3 & Sample 4 & Sample 5 & Sample 6\\
\midrule
\ours 14/medium &
\includegraphics[width=2.0cm]{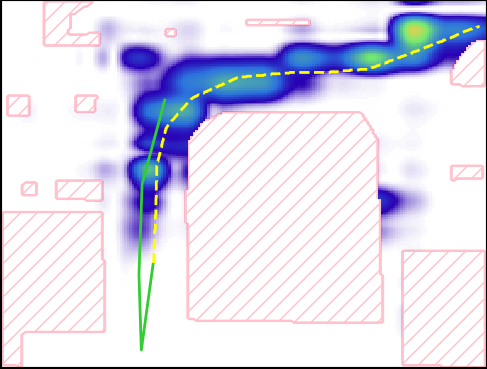}
& \includegraphics[width=2.0cm]{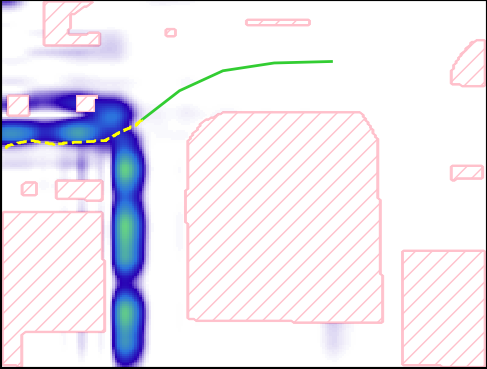}
& \includegraphics[width=2.0cm]{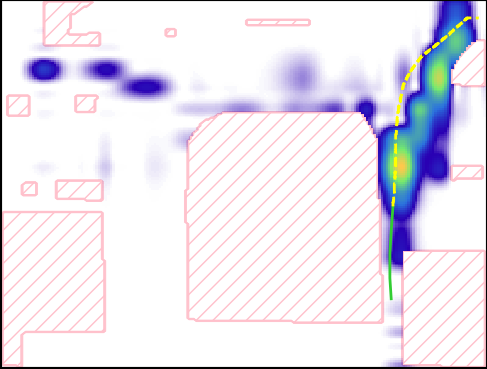}
& \includegraphics[width=2.0cm]{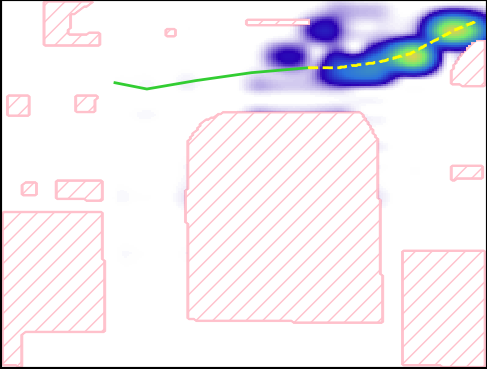}
& \includegraphics[width=2.0cm]{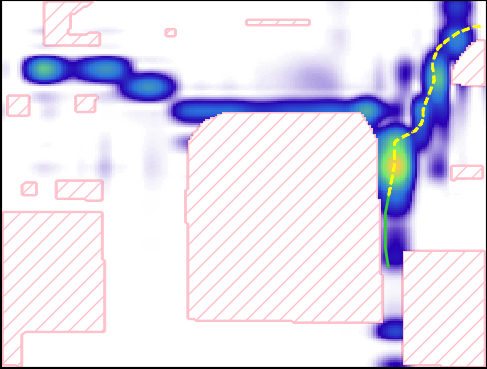}
& \includegraphics[width=2.0cm]{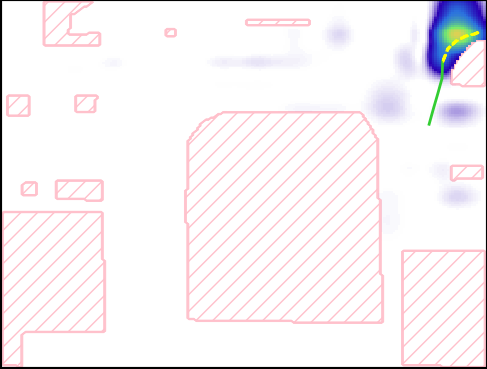}\\
\ours 10/small &
\includegraphics[width=2.0cm]{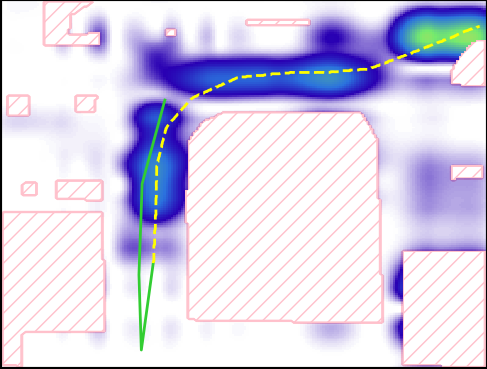}
& \includegraphics[width=2.0cm]{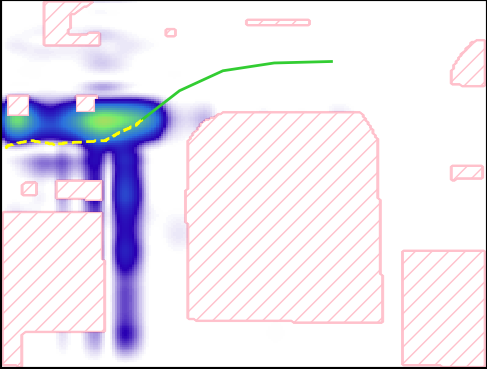}
& \includegraphics[width=2.0cm]{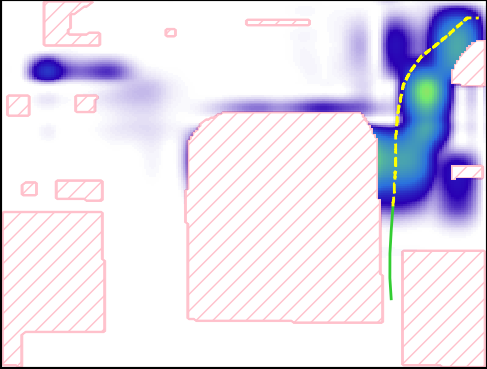}
& \includegraphics[width=2.0cm]{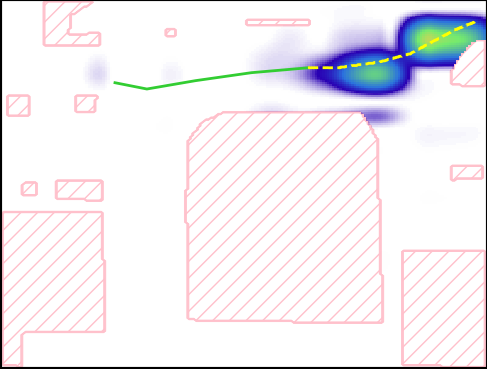}
& \includegraphics[width=2.0cm]{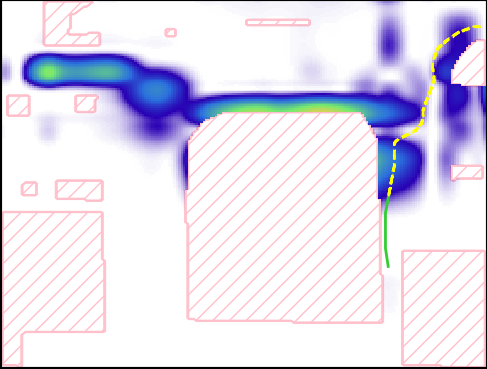}
& \includegraphics[width=2.0cm]{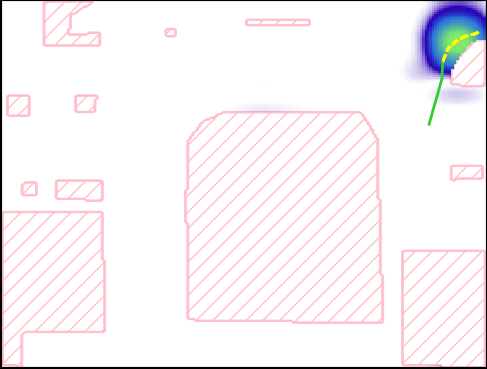}\\
\ours 14/small &
\includegraphics[width=2.0cm]{imgs/sdd/appendix/other_img_2/15546/spline_14_small.png}
& \includegraphics[width=2.0cm]{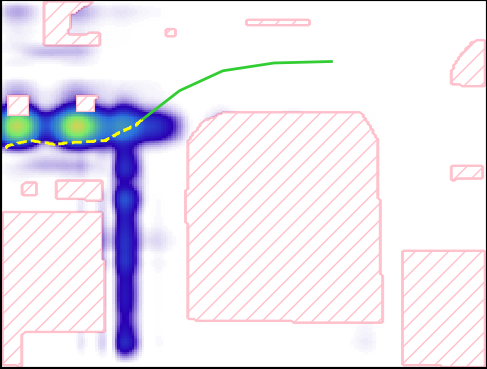}
& \includegraphics[width=2.0cm]{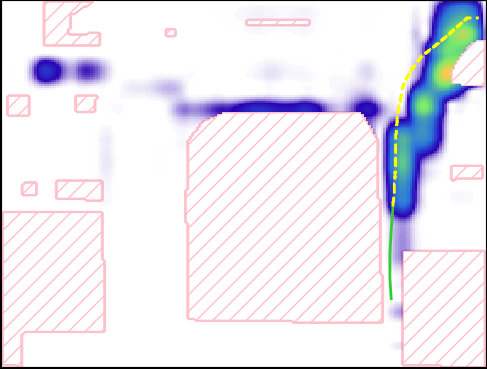}
& \includegraphics[width=2.0cm]{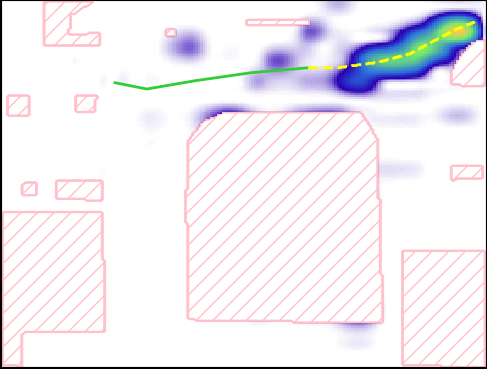}
& \includegraphics[width=2.0cm]{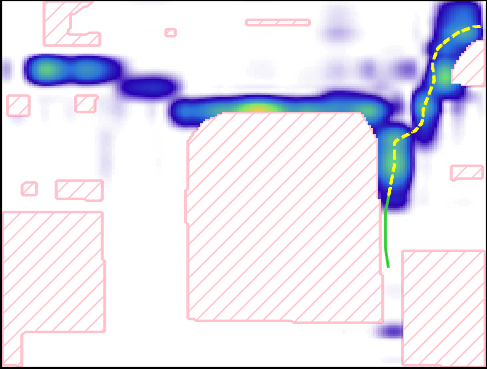}
& \includegraphics[width=2.0cm]{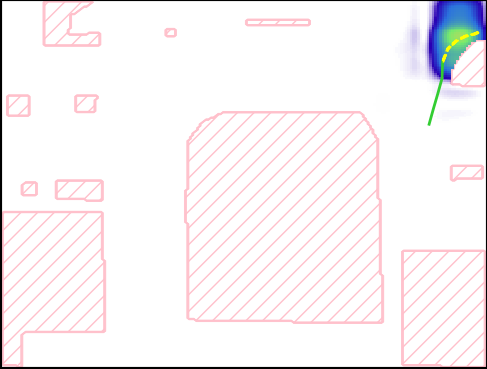}\\
\ours 10/medium &
\includegraphics[width=2.0cm]{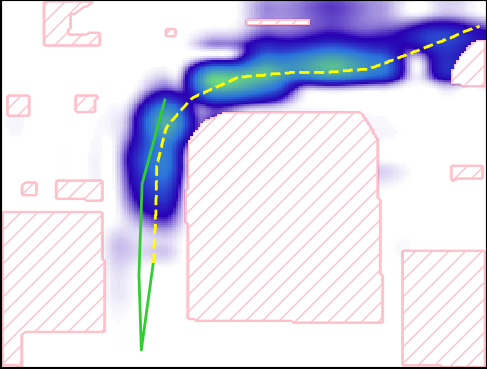}
& \includegraphics[width=2.0cm]{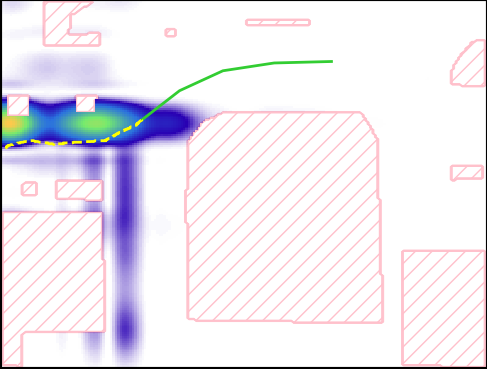}
& \includegraphics[width=2.0cm]{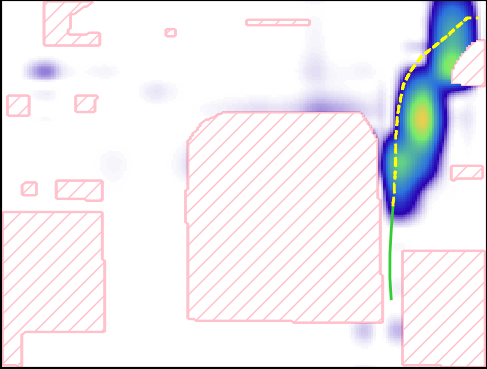}
& \includegraphics[width=2.0cm]{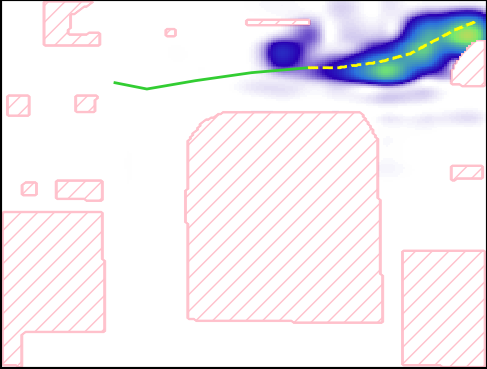}
& \includegraphics[width=2.0cm]{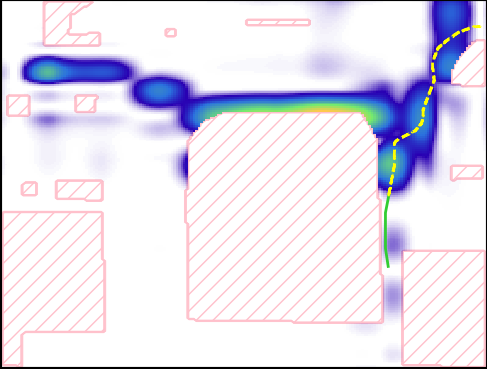}
& \includegraphics[width=2.0cm]{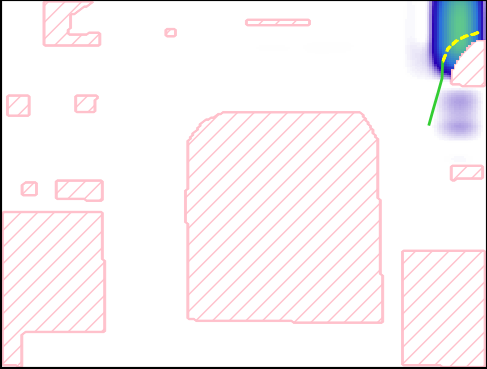}\\
GMM 32/small &
\includegraphics[width=2.0cm]{imgs/sdd/appendix/other_img_2/15546/gmm_32_small.png}
& \includegraphics[width=2.0cm]{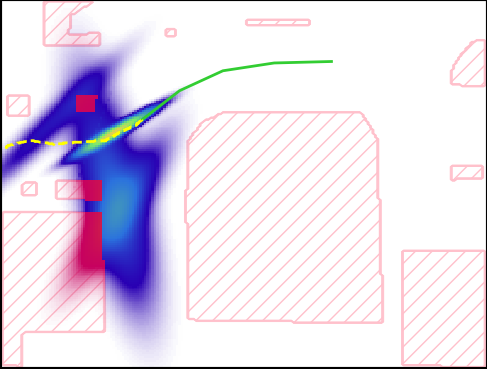}
& \includegraphics[width=2.0cm]{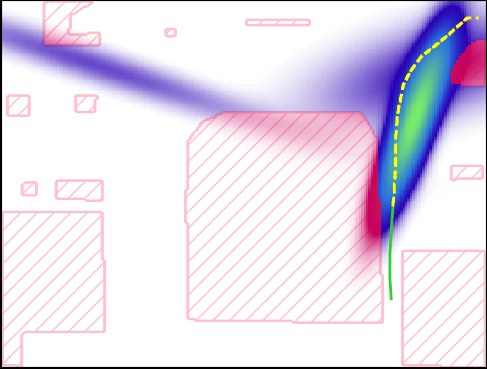}
& \includegraphics[width=2.0cm]{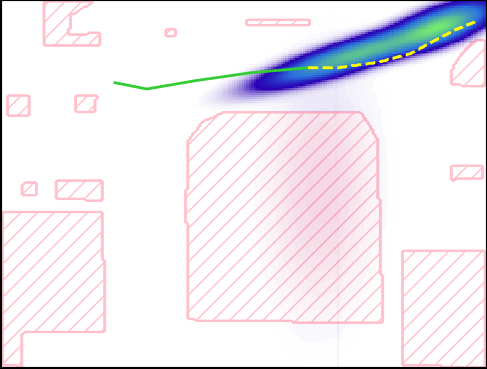}
& \includegraphics[width=2.0cm]{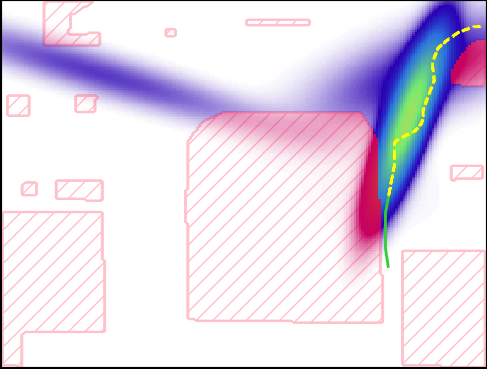}
& \includegraphics[width=2.0cm]{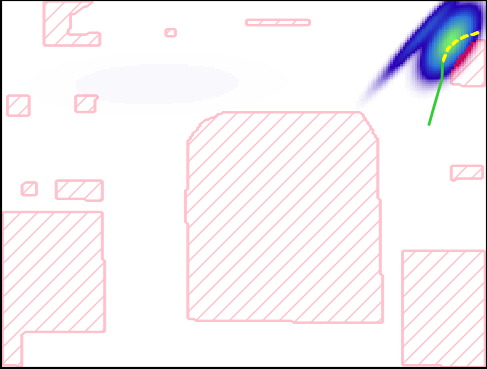}\\
GMM 4/small &
\includegraphics[width=2.0cm]{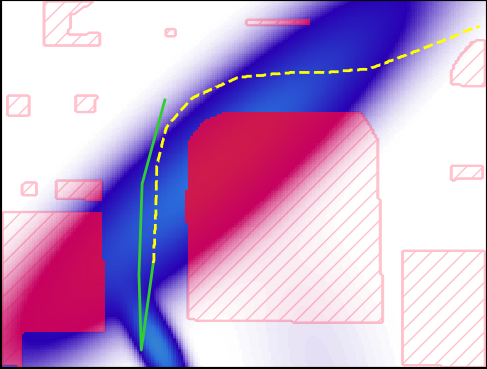}
& \includegraphics[width=2.0cm]{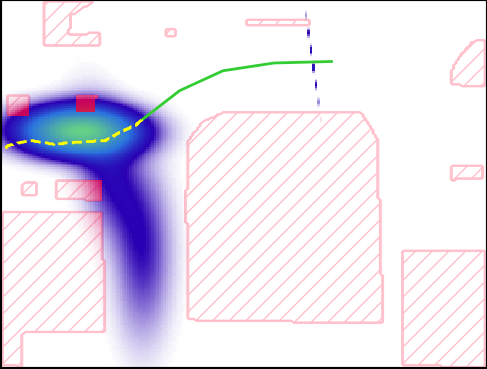}
& \includegraphics[width=2.0cm]{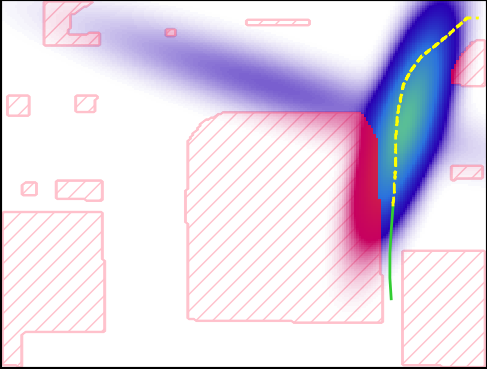}
& \includegraphics[width=2.0cm]{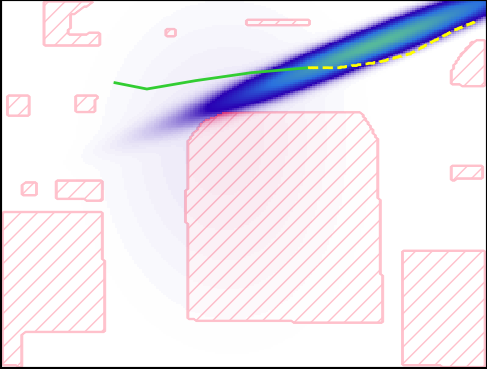}
& \includegraphics[width=2.0cm]{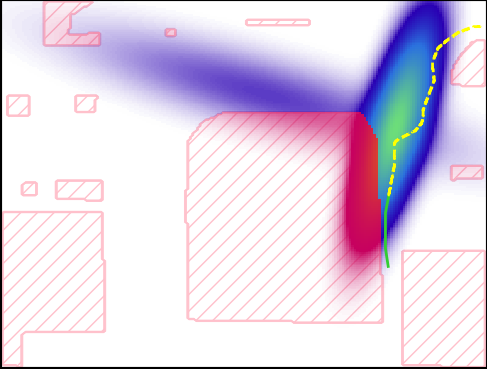}
& \includegraphics[width=2.0cm]{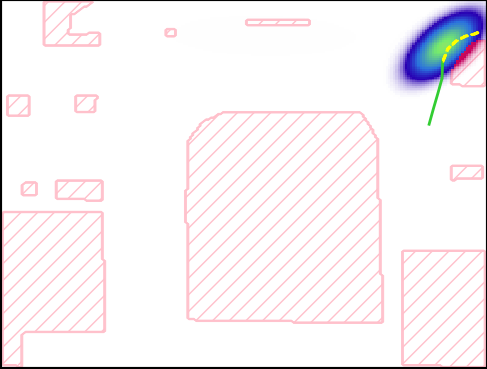}\\
GMM 50/small &
\includegraphics[width=2.0cm]{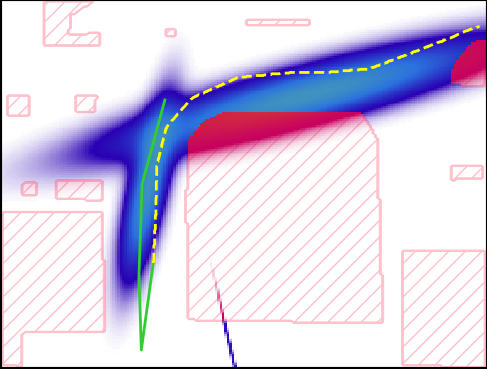}
& \includegraphics[width=2.0cm]{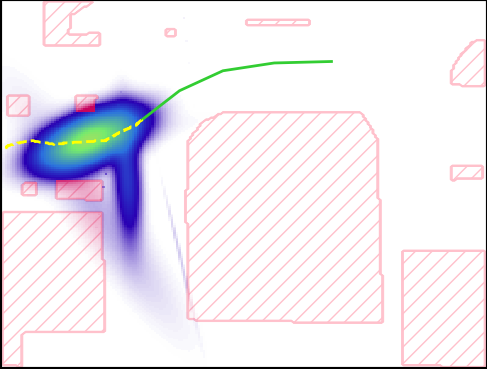}
& \includegraphics[width=2.0cm]{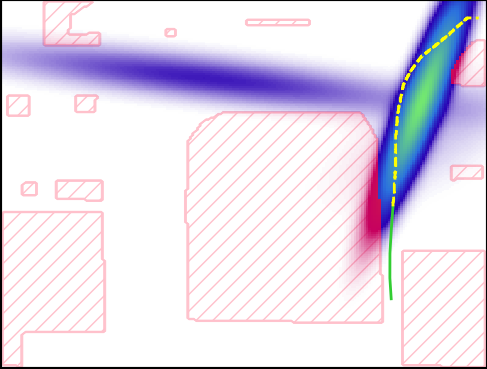}
& \includegraphics[width=2.0cm]{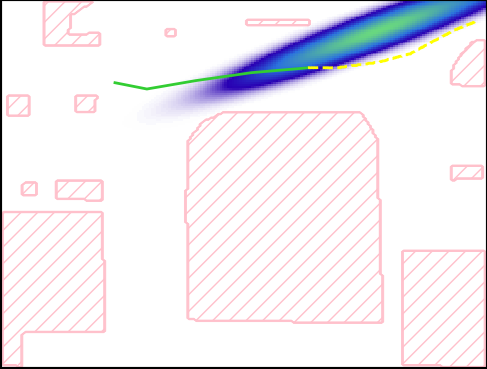}
& \includegraphics[width=2.0cm]{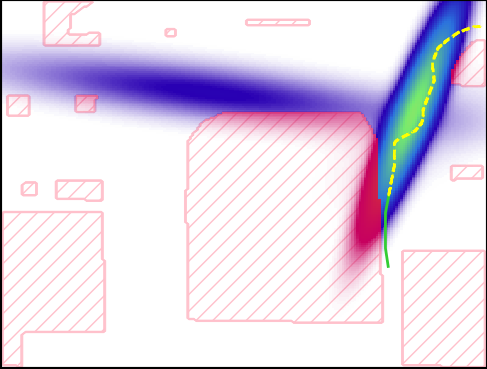}
& \includegraphics[width=2.0cm]{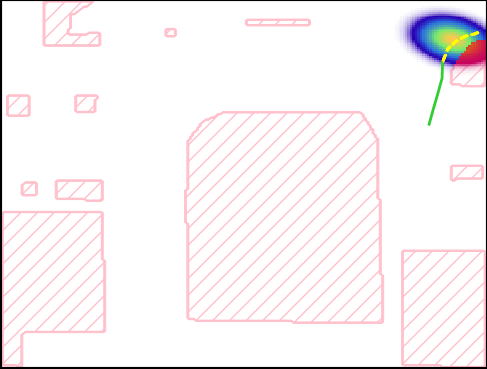}\\
GMM 100/small &
\includegraphics[width=2.0cm]{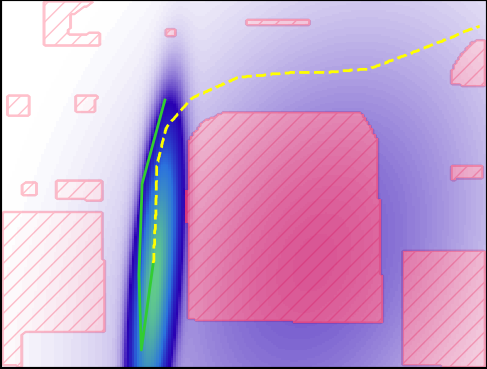}
& \includegraphics[width=2.0cm]{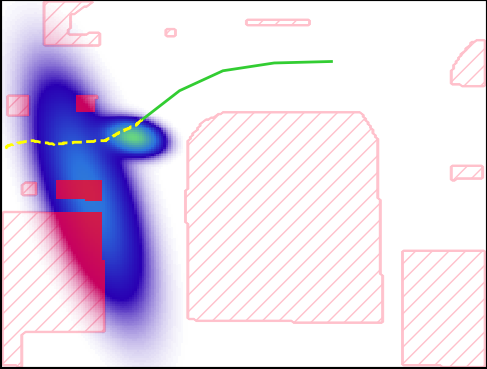}
& \includegraphics[width=2.0cm]{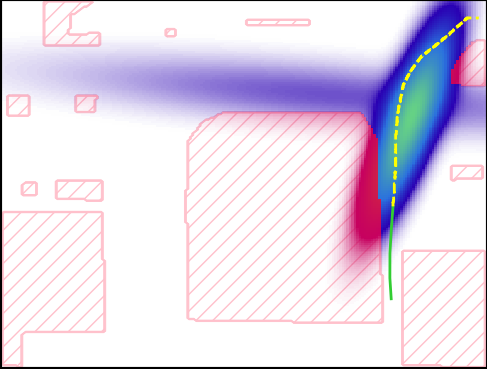}
& \includegraphics[width=2.0cm]{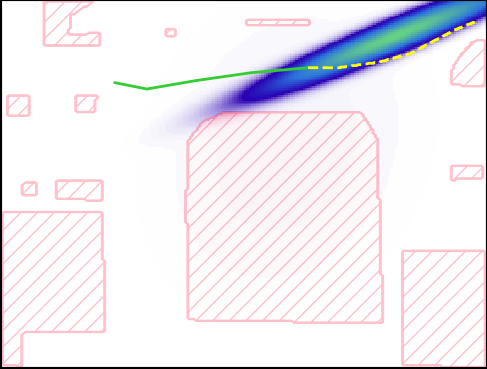}
& \includegraphics[width=2.0cm]{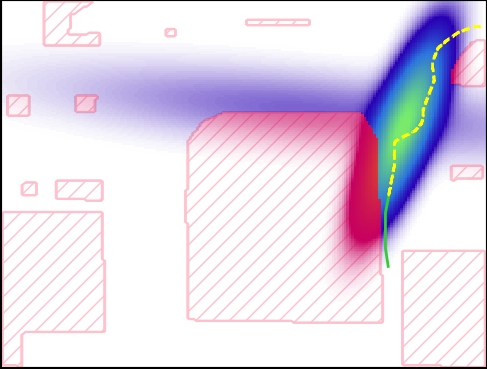}
& \includegraphics[width=2.0cm]{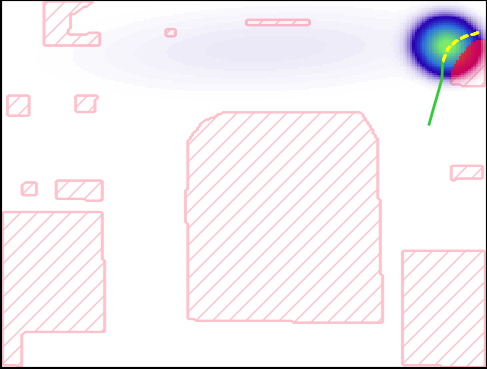}\\
DSP medium &
\includegraphics[width=2.0cm]{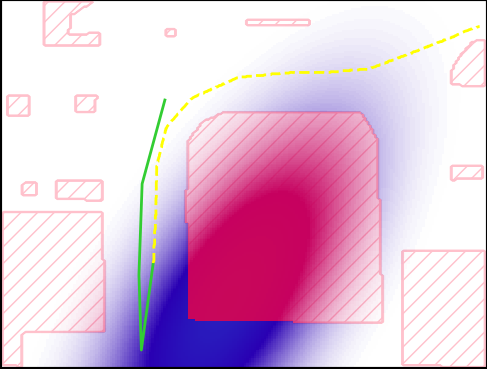}
& \includegraphics[width=2.0cm]{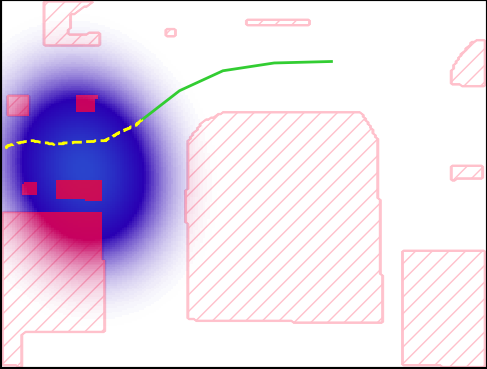}
& \includegraphics[width=2.0cm]{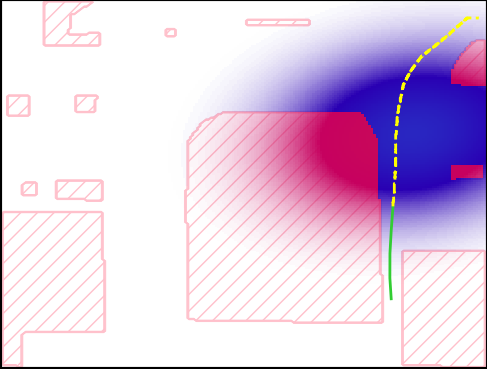}
& \includegraphics[width=2.0cm]{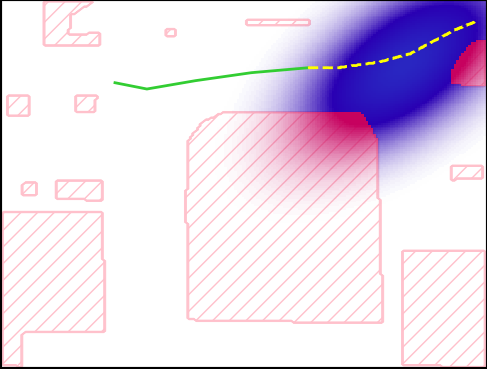}
& \includegraphics[width=2.0cm]{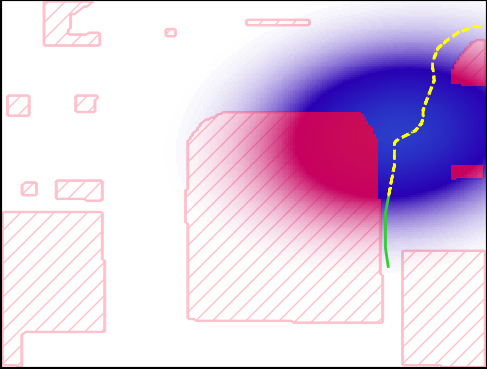}
& \includegraphics[width=2.0cm]{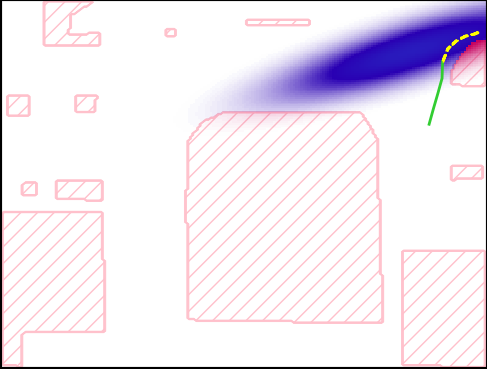}\\
DSP small &
\includegraphics[width=2.0cm]{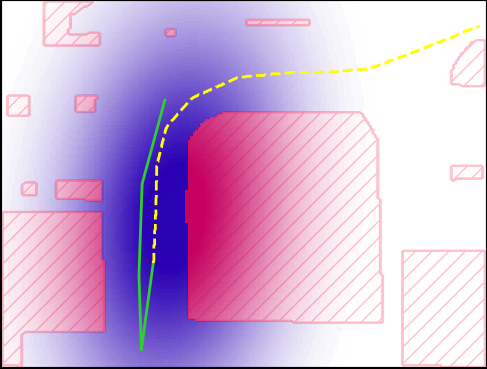}
& \includegraphics[width=2.0cm]{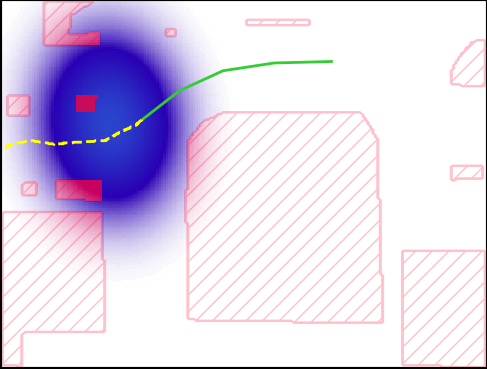}
& \includegraphics[width=2.0cm]{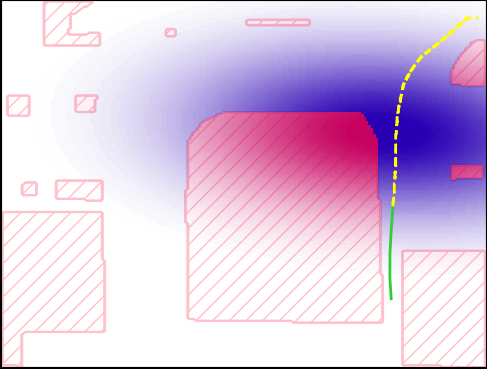}
& \includegraphics[width=2.0cm]{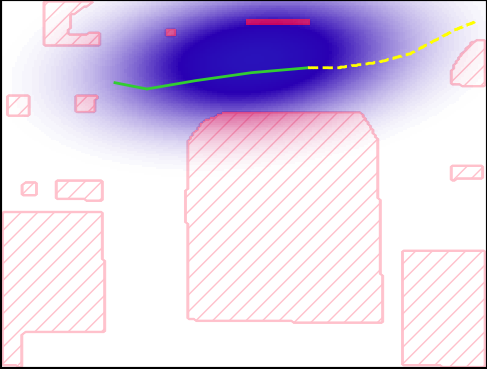}
& \includegraphics[width=2.0cm]{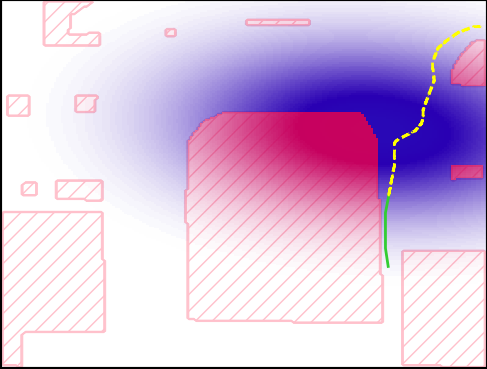}
& \includegraphics[width=2.0cm]{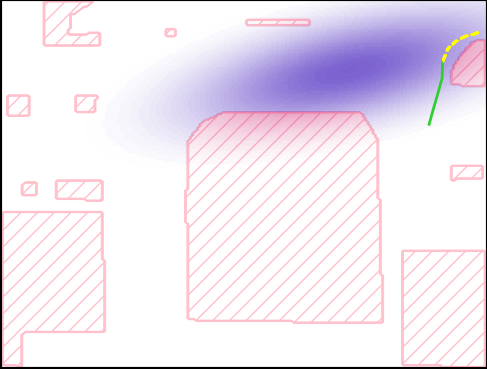}\\
DSP by LL med. &
\includegraphics[width=2.0cm]{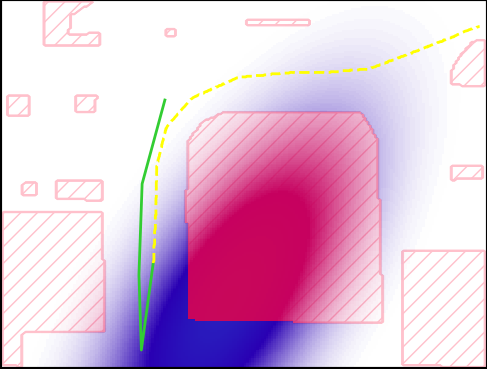}
& \includegraphics[width=2.0cm]{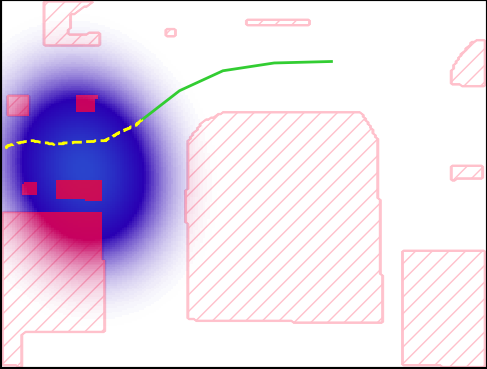}
& \includegraphics[width=2.0cm]{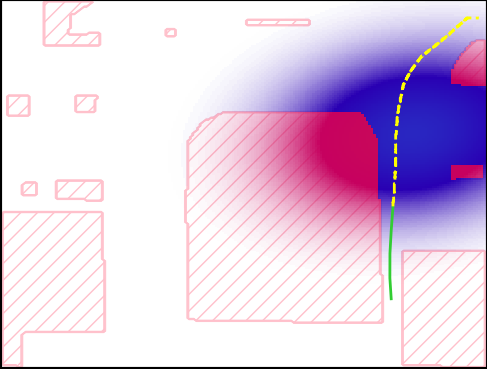}
& \includegraphics[width=2.0cm]{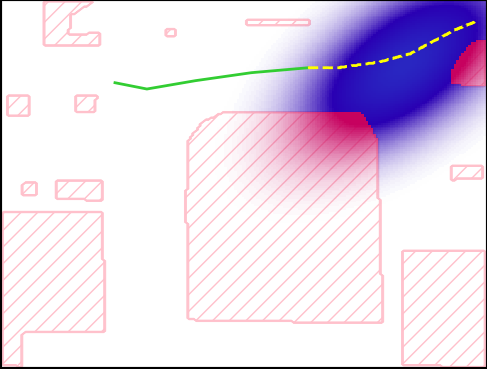}
& \includegraphics[width=2.0cm]{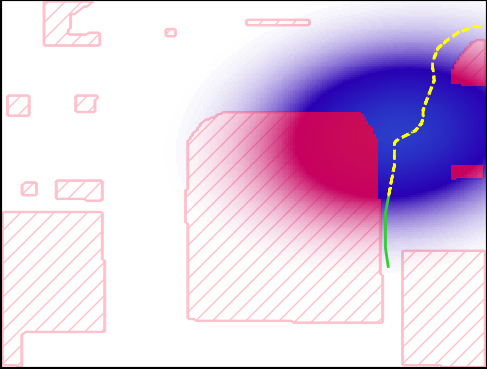}
& \includegraphics[width=2.0cm]{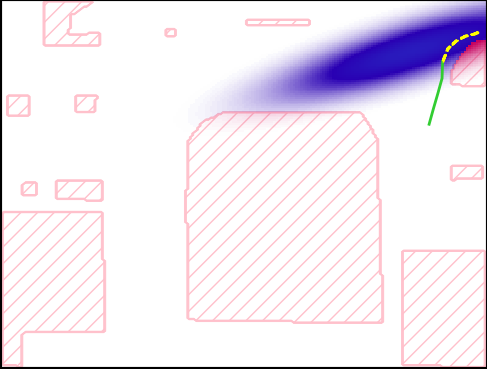}\\
DSP by LL small &
\includegraphics[width=2.0cm]{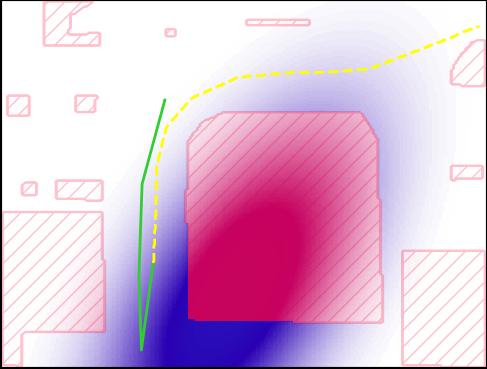}
& \includegraphics[width=2.0cm]{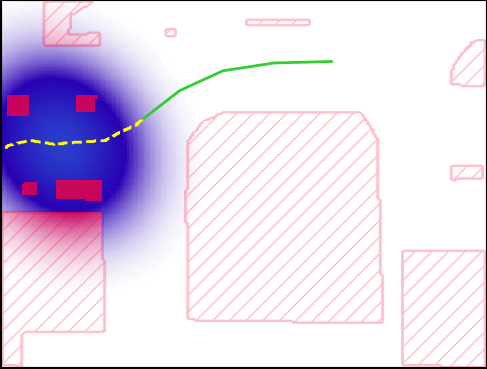}
& \includegraphics[width=2.0cm]{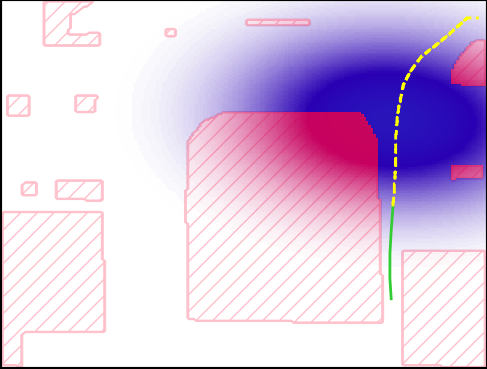}
& \includegraphics[width=2.0cm]{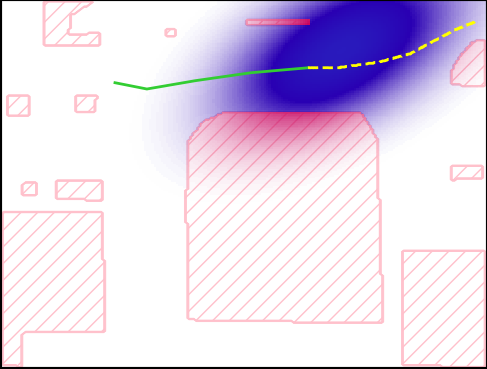}
& \includegraphics[width=2.0cm]{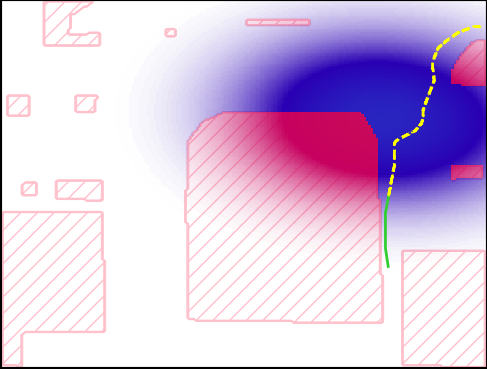}
& \includegraphics[width=2.0cm]{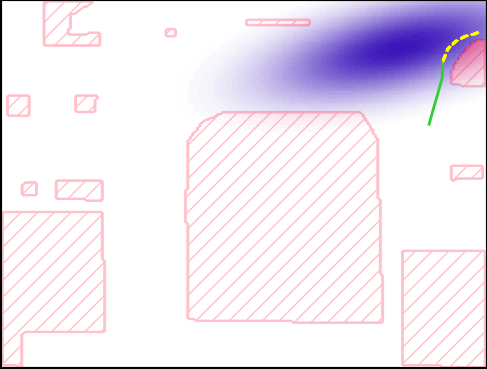}\\
\end{tabular}
}
\end{adjustbox}
    \caption{Densities for the predictive positions for the $P(\vY \mid \vX)$ case on the Stanford drone dataset on scenario 2. We compare the best $4$ spline models against the best $4$ GMM models and the best DSP models both by log-likelihood and loss from \ref{tab:sdd_traj_big_table_image_2}. Colormap is normalized per sample.}
    \label{tab:sdd_conditional_predictions_scene_2}
\end{table}

\FloatBarrier

\section{Sampling}
\label{app:sampling}

We implement sampling through autoregressive inverse-transform sampling using bisection search. In our sampling procedure we assume $\vY$ to be continuous, although it is straightforward to extend this approach to the hybrid case. Autoregressive inverse-transform sampling works through repeatedly applying the inverse-transform technique, conditioning on all the already sampled dimensions and marginalizing out all the subsequent dimensions.

In order to derive our algorithm, we will first focus on sampling the $i$-th dimension conditioned on our sampled previous dimensions, for which we first need to define the conditional CDF $F$:

\begin{align*}
    F(z) &= p(Y_i \leq z \mid \vY_{1:(i-1)}=\vy_{1:(i-1)}) \\
    &\propto \int \int_{-\infty}^z q(\vY_{i+1:n}=\vy_{i+1:n}, Y_i=y, \vY_{1:(i-1)}=\vy_{1:(i-1)}) \cdot \Ind{\vy \models \phi} dyd\vy_{i+1:n} \\
    &= \int \int_{-\infty}^z q'(\vY_{i+1:n}=\vy_{i+1:n}, Y_i=y) \cdot \Ind{(y, \vy_{i+1:n}) \models \phi'} dyd\vy_{i+1:n}\\
    &=\int q'(\vY_{i:n}=\vy_{i:n}) \cdot \Ind{\vy_{i:n} \models (\phi' \land (Y_i \leq z))} d\vy_{i:n}\\
    &= \hat{F}(z)
\end{align*}

with $q'$ denoting $q[\vY_{1:(i-1)} \mapsto \vy_{1:(i-1)}]$ and $\phi'$ denoting $\phi[\vY_{1:(i-1)} \mapsto \vy_{1:(i-1)}]$.

As we can now deduce $F(z) = \hat{F}(z) / \hat{F}(\infty)$, we have reduced our conditional CDF to our usual weighted model integral, which we can tackle with \gasp. As we need the inverse of $F$ for the inverse-transform sampling, we numerically invert the $F$ using bisection search. This leads us to our algorithm:

\begin{algorithm}
   \caption{$\mathsf{Sample}(q, \phi, \epsilon)$}
   \label{alg:sample}
   \textbf{Input} Polynomial $q$, Constraint $\phi$, Precision $\epsilon$ \\
   \textbf{Output} Sample $\vy$
   \begin{algorithmic}[1]
    \STATE $\vy' \gets []$
    \STATE $n \gets \mathsf{dim}(\mathsf{domain}(q))$
    \FOR{$i \gets 1$ \textbf{to} $n$}
        \STATE $q' \gets q[\vY_{1:(i-1)} \mapsto \vy']$
        \STATE $\phi' \gets \phi[\vY_{1:(i-1)} \mapsto \vy']$
        \STATE $f_{\infty} \gets \gasp(q, \phi')$
        \STATE $F(z) = \gasp(q, \phi' \land (Y_i \leq z)) / f_\infty$
        \STATE $lower, upper \gets \mathsf{GlobalBounds(i)}$
        \STATE $u \gets \mathsf{sample}(U[0,1])$
        \STATE $y_i \gets \mathsf{BisectionSearch}(F, u, lower, upper, \epsilon)$
        \STATE $\vy' \gets \mathsf{append}(\vy', y_i)$
    \ENDFOR
   \RETURN{$\vy'$}
   \end{algorithmic}
\end{algorithm}

\begin{algorithm}
    \caption{$\mathsf{BisectionSearch}(f, u, lower, upper, \epsilon)$}
    \label{alg:bisection_search}
    \textbf{Input} function $f: \mathbb{R} \rightarrow \mathbb{R}$, target $u$, bounds ($lower$, $upper$), precision $\epsilon$ \\
    \textbf{Output} Point $z$, $\epsilon$-close to the generalized left-inverse $F^{-1}(u)$
    \begin{algorithmic}[1]
    \WHILE{$upper - lower > \epsilon$}
        \STATE $m \gets \frac{upper + lower}{2}$
        \IF{$f(mid) \leq u$}
            \STATE $lower \gets mid$
        \ELSE
            \STATE $upper \gets mid$
        \ENDIF
    \ENDWHILE
    \RETURN $\frac{upper + lower}{2}$
    \end{algorithmic}
\end{algorithm}

In case we are dealing with splines, as we do in the Stanford-Drone Dataset, we already have an expression for the partition function, and therefore the integral over each bin at hand. We can therefore speed up the sampling by deciding on a bin first:

\begin{algorithm}
    \caption{$\mathsf{SamplePiecewise}(q_{i=1}^m, \phi, I_{i=1}^m, b_{i=1}^m, \epsilon)$}
    \label{alg:sample_piecewise}
    \textbf{Input} Polynomials $q$, Constraints $\phi$, Integrals per bin $I_{i=1}^m$, Bounds $b_{i=1}^m$, Precision $\epsilon$ \\
    \textbf{Output} Sample $\vy$
    \begin{algorithmic}[1]
    \STATE $j \gets \mathsf{sampleCategorical}(I_{i=1}^m/\mathsf{sum}(I_{i=1}^m))$
    \STATE $n \gets \mathsf{dim}(\mathsf{domain}(q_1))$
    \STATE $\phi' \gets \phi \land (\bigwedge_{i=1}^{n}(Y_i \geq \mathsf{lower}(b_j)_i \land Y_i \leq \mathsf{upper}(b_j)_i)$
    \STATE $\vy \gets \mathsf{Sample}(q_j, \phi', \epsilon)$
    \RETURN{$\vy$}
    \end{algorithmic}
\end{algorithm}

We visualize our results in figure \ref{fig:app_sampling_sdd_2}. Here we show the drawing of $100$ samples applied to Sample 1 from \ref{tab:sdd_conditional_predictions_scene_2} for the model \ours 14/medium. It takes around a second to generate a sample, with $15$ \gasp evaluations per sample. We generate a sample up to an epsilon of $0.1$ (in pixel space), which is a sensible value considering that we know that the precision of our ground truth positions is limited by the resolution of our image.

\begin{figure}[ht]
    \centering
    \begin{minipage}{0.3\textwidth}
        \centering
        Density
        \includegraphics[width=\textwidth]{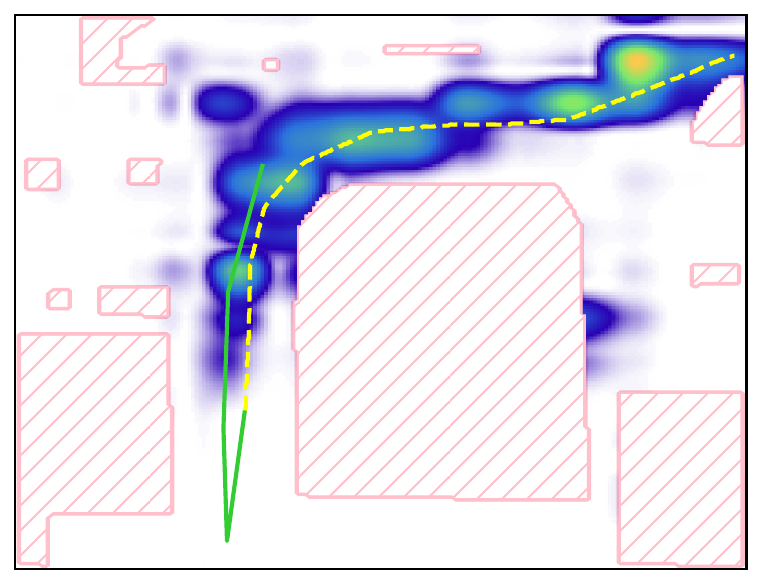}
    \end{minipage}
    \hfill
    \begin{minipage}{0.3\textwidth}
        \centering
        Integrals over Bins
        \includegraphics[width=\textwidth]{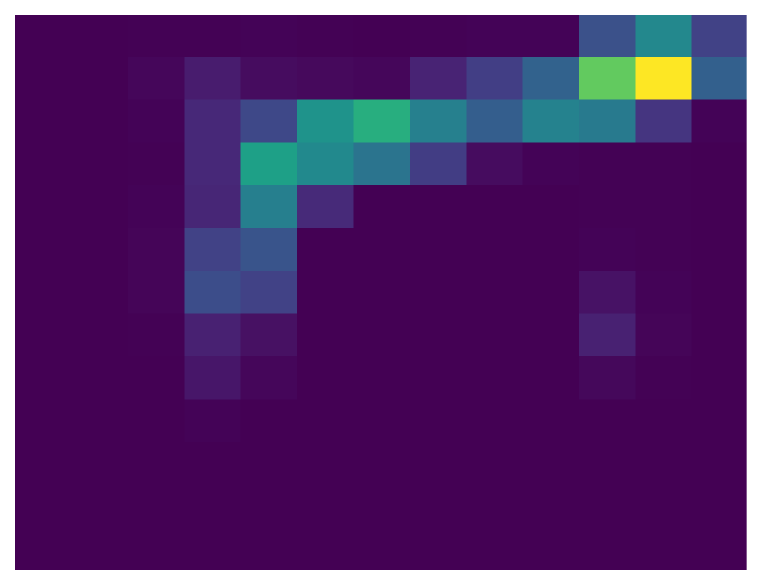}
    \end{minipage}
    \hfill
    \begin{minipage}{0.3\textwidth}
        \centering
        Samples
        \includegraphics[width=\textwidth]{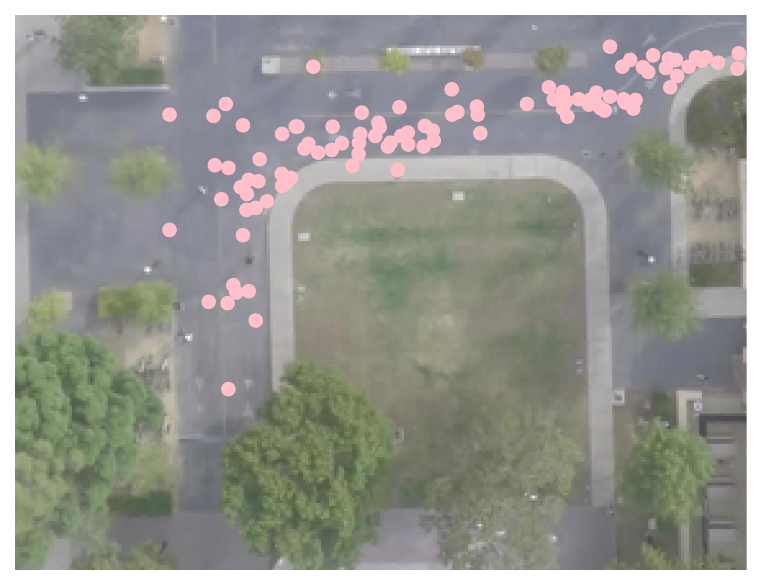}
    \end{minipage}
    \caption{The obtained samples from our sampling procedure applied to Sample 1 from \ref{tab:sdd_conditional_predictions_scene_2} for the model \ours 14/medium. On the left one can see the ground truth density of our spline-based \ours-model, followed by the integral over the bins. On the right, we show the $100$ samples drawn by applying algorithm \ref{alg:sample_piecewise}.}
    \label{fig:app_sampling_sdd_2}
\end{figure}

\end{document}

%% file: tables/benchmark_sympy.tex
\begin{tabular}{llllllll}
\toprule
$\text{total degree}$ & $0$ & $1$ & $4$ & $6$ & $8$ & $10$ & $12$ \\
$n$ &  &  &  &  &  &  &  \\
\midrule
3 & 00:00:00 & 00:00:00 & 00:00:00 & 00:00:02 & 00:00:05 & 00:00:17 & 00:00:40 \\
5 & 00:00:01 & 00:00:01 & 00:00:05 & 00:00:14 & 00:00:41 & 00:02:09 & 00:05:19 \\
7 & 00:00:04 & 00:00:06 & 00:00:14 & 00:00:42 & 00:02:04 & 00:07:07 & 00:17:55 \\
9 & 00:00:10 & 00:00:14 & 00:00:33 & 00:01:43 & 00:05:14 & 00:19:19 & 00:49:46 \\
11 & 00:00:19 & 00:00:25 & 00:01:01 & 00:03:14 & 00:10:02 & 00:37:02 & 01:36:09 \\
13 & 00:00:31 & 00:00:41 & 00:01:38 & 00:05:08 & 00:15:51 & 00:59:36 & 02:36:15 \\
15 & 00:00:53 & 00:01:09 & 00:02:42 & 00:08:21 & 00:25:55 & 01:37:29 & 04:19:17 \\
17 & 00:01:17 & 00:01:39 & 00:03:46 & 00:11:28 & 00:35:24 & 02:16:39 & 06:20:02 \\
\bottomrule
\end{tabular}

%% file: tables/benchmark_pa.tex
\begin{tabular}{llllllll}
\toprule
$\text{total degree}$ & $0$ & $1$ & $4$ & $6$ & $8$ & $10$ & $12$ \\
$n$ &  &  &  &  &  &  &  \\
\midrule
3 & 00:00:00 & 00:00:00 & 00:00:00 & 00:00:00 & 00:00:01 & 00:00:04 & 00:00:10 \\
5 & 00:00:00 & 00:00:00 & 00:00:00 & 00:00:00 & 00:00:01 & 00:00:06 & 00:00:15 \\
7 & 00:00:00 & 00:00:00 & 00:00:00 & 00:00:00 & 00:00:01 & 00:00:06 & 00:00:16 \\
7 & 00:00:00 & 00:00:00 & 00:00:00 & 00:00:00 & 00:00:01 & 00:00:06 & 00:00:16 \\
9 & 00:00:00 & 00:00:00 & 00:00:00 & 00:00:01 & 00:00:02 & 00:00:07 & 00:00:16 \\
11 & 00:00:01 & 00:00:01 & 00:00:01 & 00:00:01 & 00:00:02 & 00:00:07 & 00:00:16 \\
13 & 00:00:02 & 00:00:02 & 00:00:02 & 00:00:02 & 00:00:03 & 00:00:08 & 00:00:17 \\
15 & 00:00:02 & 00:00:02 & 00:00:02 & 00:00:03 & 00:00:04 & 00:00:09 & 00:00:18 \\
17 & 00:00:03 & 00:00:03 & 00:00:03 & 00:00:04 & 00:00:05 & 00:00:10 & 00:00:19 \\
\bottomrule
\end{tabular}

%% file: tables/nstar_overview_all_models_replicated.tex
\begin{tabular}{rrrrrrrrrr}
\toprule
& \multicolumn{3}{c}{NN + \ours } & \multicolumn{4}{c}{NN + GMM} & \multicolumn{2}{c}{DeepSeaProblog}\\
\cmidrule(lr){2-4} \cmidrule(lr){5-8} \cmidrule(lr){9-10}
$N$ & deg $10$ & deg $14$ & deg $18$ & $K{=}1$ & $K{=}4$ & $K{=}8$ & $K{=}32$ & by LL & by Loss\\
 \midrule
3 & -4.749 {\small{$\pm$}0.169}	 &\bfseries -4.674 {\small{$\pm$}0.009}	 & -4.840 {\small{$\pm$}0.251}	 & -5.016 {\small{$\pm$}0.001}	 & -4.792 {\small{$\pm$}0.003}	 & -4.740 {\small{$\pm$}0.003}	 & -4.723 {\small{$\pm$}0.006}	 & -5.027 {\small{$\pm$}0.012}	 & -42.821 {\small{$\pm$}41.989} \\
7 & -4.529 {\small{$\pm$}0.008}	 & \bfseries -4.527 {\small{$\pm$}0.009}	 & -4.741 {\small{$\pm$}0.287}	 & -5.009 {\small{$\pm$}0.000}	 & -4.850 {\small{$\pm$}0.014}	 & -4.708 {\small{$\pm$}0.007}	 & -4.612 {\small{$\pm$}0.005}	 & -5.019 {\small{$\pm$}0.010}	 & -206.411 {\small{$\pm$}132.409} \\
11 & \bfseries -4.570 {\small{$\pm$}0.223}	 & -4.584 {\small{$\pm$}0.171}	 & -4.591 {\small{$\pm$}0.180}	 & -4.988 {\small{$\pm$}0.000}	 & -4.922 {\small{$\pm$}0.015}	 & -4.791 {\small{$\pm$}0.018}	 & -4.620 {\small{$\pm$}0.008}	 & -83.042 {\small{$\pm$}58.990} & -43.151 {\small{$\pm$}42.199} \\
19 & -4.506 {\small{$\pm$}0.034}	 & \bfseries -4.492 {\small{$\pm$}0.004}	 & 4.616 {\small{$\pm$}0.228}	 & -4.995 {\small{$\pm$}0.000}	 & -4.981 {\small{$\pm$}0.003}	 & -4.925 {\small{$\pm$}0.010}	 & -4.652 {\small{$\pm$}0.003}	 & -5.001 {\small{$\pm$}0.001}	 & -155.139 {\small{$\pm$}101.533} \\
\bottomrule
\end{tabular}

%% file: tables/sdd_marginal_big_replicated.tex
\begin{tabular}{llrrr}
\toprule
type & detail & num. params & $\mathsf{Pr}(\neg \phi)$ & log-like \\
\midrule
\ours (Spline) & 8 knots, 10 mixtures & 330 & \bfseries 0.000 & -3.013 {\small{$\pm$}0.002}	 \\
\ours (Spline) & 8 knots, 8 mixtures & 264 & \bfseries 0.000 & 3.023 {\small{$\pm$}0.003}	 \\
\ours (Spline) & 8 knots, 4 mixtures & 132 & \bfseries 0.000 & -3.067 {\small{$\pm$}0.009}	 \\
\ours (Spline) & 10 knots, 10 mixtures & 410 & \bfseries 0.000 & -2.984 {\small{$\pm$}0.002}	 \\
\ours (Spline) & 10 knots, 8 mixtures & 328 & \bfseries 0.000 & -2.995 {\small{$\pm$}0.005}	 \\
\ours (Spline) & 10 knots, 4 mixtures & 164 & \bfseries 0.000 & -3.045 {\small{$\pm$}0.008}		 \\
\ours (Spline) & 12 knots, 10 mixtures & 490 & \bfseries 0.000 & -2.971 {\small{$\pm$}0.003}	 \\
\ours (Spline) & 12 knots, 8 mixtures & 392 & \bfseries 0.000 & -2.979 {\small{$\pm$}0.003}		 \\
\ours (Spline) & 12 knots, 4 mixtures & 196 & \bfseries 0.000 & -3.025 {\small{$\pm$}0.010}	 \\
\ours (Spline) & 14 knots, 10 mixtures & 570 & \bfseries 0.000 & -2.950 {\small{$\pm$}0.002}		 \\
\ours (Spline) & 14 knots, 8 mixtures & 456 & \bfseries 0.000 & -2.961 {\small{$\pm$}0.002}	 \\
\ours (Spline) & 14 knots, 4 mixtures & 228 & \bfseries 0.000 & -3.009 {\small{$\pm$}0.008}	 \\
\ours (Spline) & 16 knots, 10 mixtures & 650 & \bfseries 0.000 & -2.937 {\small{$\pm$}0.003}		 \\
\ours (Spline) & 16 knots, 8 mixtures & 520 & \bfseries 0.000 & -2.948 {\small{$\pm$}0.003}	 \\
\ours (Spline) & 16 knots, 4 mixtures & 260 & \bfseries 0.000 & -2.998 {\small{$\pm$}0.010}	 \\
\midrule
GMM & $K{=}5$ & 30 & $\approx$ 12.475 {\small{$\pm$}0.761}	 & -3.359 {\small{$\pm$}0.034}		 \\
GMM & $K{=}10$ & 60 & $\approx$ 8.887 {\small{$\pm$}0.224}	 & -3.223 {\small{$\pm$}0.008}	 \\
GMM & $K{=}20$ & 120 & $\approx$ 4.229 {\small{$\pm$}0.142}	 & -3.081 {\small{$\pm$}0.012}		 \\
GMM & $K{=}50$ & 300 & $\approx$ 2.375 {\small{$\pm$}0.112}	 & -2.983 {\small{$\pm$}0.004}		 \\
GMM & $K{=}100$ & 600 & $\approx$ 1.190 {\small{$\pm$}0.052}	 & \bfseries -2.917 {\small{$\pm$}0.005}	 \\
\midrule
Flow & 1 transformation ($t$), 128x2 hidden & 22830 & $\approx$ 5.643 {\small{$\pm$}0.487}	 & -3.098 {\small{$\pm$}0.013}		 \\
Flow & 1 transformation ($t$), 64x2 hidden & 7342 & $\approx$ 5.245 {\small{$\pm$}0.516}	 & -3.089 {\small{$\pm$}0.017}	 \\
Flow & 1 transformation ($t$), 32x2 hidden & 2670 & $\approx$ 5.651 {\small{$\pm$}0.596}	 & -3.109 {\small{$\pm$}0.016}		 \\
Flow & 2 transformations ($t$), 128x2 hidden & 45660 & $\approx$ 2.616 {\small{$\pm$}0.221}	 & -2.986 {\small{$\pm$}0.008}		 \\
Flow & 2 transformations ($t$), 64x2 hidden & 14684 & $\approx$ 2.157 {\small{$\pm$}0.250}	 & -2.972 {\small{$\pm$}0.011}	 \\
Flow & 2 transformations ($t$), 32x2 hidden & 5340 & $\approx$ 2.468 {\small{$\pm$}0.612}	 & -2.979 {\small{$\pm$}0.016}	 \\
Flow & 5 transformations ($t$), 128x2 hidden & 114150 & $\approx$ 1.771 {\small{$\pm$}0.159}	 & -2.940 {\small{$\pm$}0.007}		 \\
Flow & 5 transformations ($t$), 64x2 hidden & 36710 & $\approx$ 1.930 {\small{$\pm$}0.130}	 & -2.949 {\small{$\pm$}0.007}		 \\
Flow & 5 transformations ($t$), 32x2 hidden & 13350 & $\approx$ 1.677 {\small{$\pm$}0.231}	 & -2.943 {\small{$\pm$}0.012}		 \\
Flow & 10 transformations ($t$), 128x2 hidden & 228300 & $\approx$ 1.502 {\small{$\pm$}0.109}	 & -2.919 {\small{$\pm$}0.007}	 \\
Flow & 10 transformations ($t$), 64x2 hidden & 73420 & $\approx$ 1.698 {\small{$\pm$}0.202}	 & -2.943 {\small{$\pm$}0.018}	 \\
Flow & 10 transformations ($t$), 32x2 hidden & 26700 & $\approx$ 1.826 {\small{$\pm$}0.252}	 & -2.938 {\small{$\pm$}0.007}		 \\
\bottomrule
\end{tabular}

%% file: tables/sdd_marginal_big_2_replicated.tex
\begin{tabular}{llrrr}
\toprule
type & detail & num. params & perc. invalid & log-like \\
\midrule
\ours (Spline) & 8 knots, 10 mixtures & 330 & \bfseries 0.000 & -3.376 {\small{$\pm$}0.007}	 \\
\ours (Spline) & 8 knots, 8 mixtures & 264 & \bfseries 0.000 & -3.376 {\small{$\pm$}0.007}		 \\
\ours (Spline) & 10 knots, 10 mixtures & 410 & \bfseries 0.000 & -3.348 {\small{$\pm$}0.005}		 \\
\ours (Spline) & 10 knots, 8 mixtures & 328 & \bfseries 0.000 & -3.362 {\small{$\pm$}0.006}		 \\
\ours (Spline) & 12 knots, 10 mixtures & 490 & \bfseries 0.000 & -3.329 {\small{$\pm$}0.003}		 \\
\ours (Spline) & 12 knots, 8 mixtures & 392 & \bfseries 0.000 & -3.343 {\small{$\pm$}0.004}		 \\
\ours (Spline) & 14 knots, 10 mixtures & 570 & \bfseries 0.000 & -3.313 {\small{$\pm$}0.003}		 \\
\ours (Spline) & 14 knots, 8 mixtures & 456 & \bfseries 0.000 & -3.333 {\small{$\pm$}0.007}		 \\
\ours (Spline) & 16 knots, 10 mixtures & 650 & \bfseries 0.000 & -3.301 {\small{$\pm$}0.004}		 \\
\ours (Spline) & 16 knots, 8 mixtures & 520 & \bfseries 0.000 & -3.322 {\small{$\pm$}0.004}		 \\
\midrule
GMM & $K{=}5$ & 35 & $\approx$ 12.220 {\small{$\pm$}0.053}	 & -3.701 {\small{$\pm$}0.002}	 \\
GMM & $K{=}10$ & 70 & $\approx$ 6.589 {\small{$\pm$}0.283}	
 & -3.564 {\small{$\pm$}0.012}	 \\
GMM & $K{=}20$ & 140 & $\approx$3.223 {\small{$\pm$}0.542}	 & -3.449 {\small{$\pm$}0.019}	 \\
GMM & $K{=}50$ & 350 & $\approx$ 1.289 {\small{$\pm$}0.159}	 & -3.351 {\small{$\pm$}0.014}	 \\
GMM & $K{=}100$ & 700 & $\approx$ 0.656 {\small{$\pm$}0.042}	 & \bfseries -3.259 {\small{$\pm$}0.005}	 \\
\midrule
Flow & 1 transformation ($t$), 32x2 hidden & 2670 &  $\approx$ 2.274 {\small{$\pm$}0.214}	 & -3.431 {\small{$\pm$}0.012}	 \\
Flow & 2 transformations ($t$), 32x2 hidden & 5340 &  $\approx$ 1.210 {\small{$\pm$}0.285}	 & -3.332 {\small{$\pm$}0.020}	 \\
Flow & 5 transformations ($t$), 32x2 hidden & 13350 &  $\approx$ 0.710 {\small{$\pm$}0.126}	 & -3.266 {\small{$\pm$}0.015}	 \\
Flow & 10 transformations ($t$), 32x2 hidden & 26700 &  $\approx$ 0.867 {\small{$\pm$}0.126}	 & -3.273 {\small{$\pm$}0.012}	 \\
\bottomrule
\end{tabular}

%% file: tables/sdd_overview_replicated.tex
\begin{tabular}{rllrrrr}
\toprule
type & size dist. & net size %
& log-like & num. params dist. &  $\mathsf{Pr}(\neg \phi)$ \\
\hline
\ours (Spline) & 14 knots, 10 mixtures & medium %
& -2.209 {\small{$\pm$}0.136}	 & 570 & \bfseries 0.000 \\
\ours (Spline) & 10 knots, 8 mixtures & medium %
& -2.942 {\small{$\pm$}1.204}	 & 328 & \bfseries 0.000 \\
\ours (Spline) & 14 knots, 8 mixtures & small %
&  \bfseries  -2.086 {\small{$\pm$}0.144}	 & 456 & \bfseries 0.000 \\
\ours (Spline) & 10 knots, 8 mixtures & small %
& -2.272 {\small{$\pm$}0.130}	 & 328 & \bfseries 0.000 \\
\ours (Spline) & 14 knots, 10 mixtures & large %
& -2.174 {\small{$\pm$}0.120}	 & 570 & \bfseries 0.000 \\
\ours (Spline) & 10 knots, 8 mixtures & large %
& -2.263 {\small{$\pm$}0.168}	 & 328 & \bfseries 0.000 \\
\midrule
GMM & ${K}{=}100$ & medium %
& -3.323 {\small{$\pm$}0.738}	 &700 & $\approx$ 20.191 {\small{$\pm$}1.937}	 \\
GMM & ${K}{=}80$ & medium %
& -11.173 {\small{$\pm$}8.202}	  & 560 & $\approx$ 72.969 {\small{$\pm$}34.192}	 \\
GMM & ${K}{=}50$ & medium %
& -2.932 {\small{$\pm$}0.361}	  & 350 & $\approx$ 21.706 {\small{$\pm$}2.727}	 \\
GMM & ${K}{=}32$ & medium %
& -2.989 {\small{$\pm$}0.345}	  & 224 & $\approx$ 21.620 {\small{$\pm$}2.672}	 \\
GMM & ${K}{=}4$ & medium %
& -3.262 {\small{$\pm$}0.379}	  & 28 & $\approx$ 20.955 {\small{$\pm$}1.366}	 \\
GMM & ${K}{=}100$ & small %
& -2.838 {\small{$\pm$}0.485}	  & 700 & $\approx$ 20.321 {\small{$\pm$}1.672}	 \\
GMM & ${K}{=}80$ & small %
& -2.835 {\small{$\pm$}0.410}	  & 560 & $\approx$ 20.096 {\small{$\pm$}1.964}	 \\
GMM & ${K}{=}50$ & small %
& -2.644 {\small{$\pm$}0.258}	  & 350 & $\approx$ 20.989 {\small{$\pm$}1.774}	 \\
GMM & ${K}{=}32$ & small %
& -2.735 {\small{$\pm$}0.379}	  & 224 & $\approx$ 19.842 {\small{$\pm$}1.584}	 \\
GMM & ${K}{=}4$ & small %
& -3.235 {\small{$\pm$}0.970}	  & 28 & $\approx$ 21.859 {\small{$\pm$}1.600}	 \\
GMM & ${K}{=}50$ & large %
& -3.074 {\small{$\pm$}0.273}	  & 350 & $\approx$ 24.721 {\small{$\pm$}3.632}	 \\
GMM & ${K}{=}32$ & large %
& -6.383 {\small{$\pm$}6.179}	  & 224 & $\approx$ 36.660 {\small{$\pm$}33.943}	 \\
GMM & ${K}{=}100$ & large %
& -4.963 {\small{$\pm$}4.690}	  & 700 & $\approx$ 30.516 {\small{$\pm$}24.794}	 \\
GMM & ${K}{=}80$ & large %
& -16.420 {\small{$\pm$}9.413}	  & 560 & $\approx$ 83.409 {\small{$\pm$}31.042}	 \\
GMM & ${K}{=}4$ & large %
& -7.553 {\small{$\pm$}5.529}	  & 28 & $\approx$ 48.345 {\small{$\pm$}35.762}	 \\
\midrule
DSP (by loss) & 1 Gaussian & large %
& -8.532 {\small{$\pm$}11.325}	 & 6 & $\approx$ 29.035 {\small{$\pm$}3.074}	 \\
DSP (by loss) & 1 Gaussian & medium %
& -6.176 {\small{$\pm$}5.132}	 & 6 & $\approx$ 36.648 {\small{$\pm$}9.670}	 \\
DSP (by loss) & 1 Gaussian & small %
& -3.876 {\small{$\pm$}0.466}	 & 6 & $\approx$ 49.046 {\small{$\pm$}16.395}	 \\
DSP (by log-like) & 1 Gaussian & large %
& -8.520 {\small{$\pm$}11.330}	 & 6 & $\approx$ 34.963 {\small{$\pm$}9.933}	 \\
DSP (by log-like) & 1 Gaussian & medium %
& -6.152 {\small{$\pm$}5.144} & 6 & $\approx$ 33.322 {\small{$\pm$}5.761}	 \\
DSP (by log-like) & 1 Gaussian & small %
& -3.861 {\small{$\pm$}0.480} & 6 & $\approx$ 58.209 {\small{$\pm$}15.949}	 \\
\bottomrule
\end{tabular}

%% file: tables/sdd_overview_2_replicated.tex
\begin{tabular}{llrrrr}
\toprule
type & size dist. & net size & log-like & num. params dist. &  $\mathsf{Pr}(\neg \phi)$ \\
\midrule
NN + \ours  & 14 knots, 10 mixtures & medium %
& -2.237 {\small{$\pm$}0.268}	 & 570 & \bfseries 0.000 \\
NN + \ours  & 10 knots, 8 mixtures & medium %
& -2.302 {\small{$\pm$}0.175}	 & 328 & \bfseries 0.000 \\
NN + \ours  & 14 knots, 8 mixtures & small %
& -2.232 {\small{$\pm$}0.275}	 & 456 & \bfseries 0.000 \\
NN + \ours  & 10 knots, 10 mixtures & small %
& \bfseries -2.091 {\small{$\pm$}0.086}	 & 410 & \bfseries 0.000 \\
\midrule
NN + GMM  & $K$=100 & medium & -2.816 {\small{$\pm$}0.205}	 & 700 & $\approx$ 15.435 {\small{$\pm$}4.099}	 \\
NN + GMM  & $K$=50 & medium & -3.129 {\small{$\pm$}0.572}	 & 350 & $\approx$ 14.905 {\small{$\pm$}3.603}	 \\
NN + GMM  & $K$=32 & medium & -2.714 {\small{$\pm$}0.228}	 & 224 & $\approx$ 16.399 {\small{$\pm$}3.852}	 \\
NN + GMM  & $K$=4 & medium & -2.780 {\small{$\pm$}0.316}	 & 28 & $\approx$ 17.271 {\small{$\pm$}4.785}	 \\
NN + GMM  & $K$=100 & small & -2.423 {\small{$\pm$}0.193}	 & 700 & $\approx$ 14.684 {\small{$\pm$}2.973}	 \\
NN + GMM  & $K$=50 & small & -2.396 {\small{$\pm$}0.231}	 & 350 & $\approx$ 15.610 {\small{$\pm$}3.715}	 \\
NN + GMM  & $K$=32 & small & -2.674 {\small{$\pm$}0.310}	 & 224 & $\approx$ 15.390 {\small{$\pm$}4.273}	 \\
NN + GMM  & $K$=4 & small & -2.650 {\small{$\pm$}0.281}	 & 28 & $\approx$ 19.093 {\small{$\pm$}6.341}	 \\
\midrule
DSP by loss & 1 Gaussian & medium & -3.611 {\small{$\pm$}0.287}	 & 6 & $\approx$ 35.986 {\small{$\pm$}6.946}	 \\
DSP by loss & 1 Gaussian & small & -30.006 {\small{$\pm$}76.260}	 & 6 & $\approx$ 52.929 {\small{$\pm$}17.107}	 \\
DSP by log-like & 1 Gaussian & medium & -3.611 {\small{$\pm$}0.287}	 & 6 & $\approx$ 36.148 {\small{$\pm$}1.641}	 \\
DSP by log-like & 1 Gaussian & small & -29.967 {\small{$\pm$}76.274}	 & 6 & $\approx$ 36.433 {\small{$\pm$}2.976}	 \\
\bottomrule
\end{tabular}